\documentclass[10pt,journal,compsoc]{IEEEtran}
\ifCLASSOPTIONcompsoc
\usepackage[nocompress]{cite}
\else
\usepackage{cite}
\fi
\usepackage{algorithm}
\usepackage{algorithmic}
\usepackage{setspace}
\usepackage{amsmath}
\usepackage{amsfonts}
\usepackage{CJK}
\usepackage{indentfirst}
\usepackage{bm}
\usepackage{graphicx}
\makeatletter
\let\MYcaption\@makecaption
\makeatother

\usepackage[font=footnotesize]{subcaption}

\makeatletter
\let\@makecaption\MYcaption
\makeatother
\usepackage[hidelinks,colorlinks=true,linkcolor=blue,citecolor=blue]{hyperref}
\usepackage{booktabs}
\usepackage{multirow}
\usepackage{amssymb}
\usepackage{color}

\usepackage{float}
\usepackage{multicol}
\usepackage{amsthm}
\usepackage{makecell}

\usepackage{cite}
\theoremstyle{plain}
\newtheorem{theorem}{Theorem}[section]

\theoremstyle{definition}

\theoremstyle{remark}

\usepackage{amssymb}
\usepackage{pifont}
\newcommand{\cmark}{\ding{51}}%
\newcommand{\xmark}{\ding{55}}%

\begin{document}
	%
	\title{Learning Continuous Wasserstein Barycenter Space for Generalized All-in-One Image Restoration}
	%
	%
	%
	%
	
	\author{Xiaole Tang,
		Xiaoyi He,
		Jiayi Xu,
		Xiang Gu,
		Jian Sun
		\IEEEcompsocitemizethanks{
			\IEEEcompsocthanksitem This work was supported in part by National Key R\&D Program under Grant 2021YFA1003000, in part by NSFC under Grant 125B2028, Grant 12125104, Grant 12426313, and Grant 12501709, and in part by the Fundamental Research Funds for the Central Universities, China under Grant xzy022025047. 
			\IEEEcompsocthanksitem The authors are with the School of Mathematics and Statistics, Xi'an Jiaotong University, Shaanxi, P.R. China. E-mail: \{tangxl, hexiaoyi, jiayixu\}@stu.xjtu.edu.cn,  \{xianggu, jiansun\}@xjtu.edu.cn. Code will be publicly available and updated at: \url{https://github.com/xl-tang3/BaryIR}.
		}
		\thanks{(Corresponding author: Jian Sun.)
	}}
	
	%
	%


\markboth{IEEE TRANSACTIONS ON PATTERN ANALYSIS AND MACHINE INTELLIGENCE}%
{Shell \MakeLowercase{\textit{Tang et al.}}: Bare Demo of IEEEtran.cls for Computer Society Journals}

\IEEEtitleabstractindextext{%
	\begin{abstract}
		Despite substantial advances in all-in-one image restoration for addressing diverse degradations within a unified model, existing methods remain vulnerable to out-of-distribution degradations, thereby limiting their generalization in real-world scenarios. To tackle the challenge, this work is motivated by the intuition that multisource degraded feature distributions are induced by different degradation-specific shifts from an underlying degradation-agnostic distribution, and recovering such a shared distribution is thus crucial for achieving generalization across degradations. With this insight, we propose BaryIR, a representation learning framework that aligns multisource degraded features in the Wasserstein barycenter (WB) space, which models a degradation-agnostic distribution by minimizing the average of Wasserstein distances to  multisource degraded distributions. We further introduce residual subspaces, whose embeddings are mutually contrasted while remaining orthogonal to the WB embeddings. Consequently, BaryIR explicitly decouples two orthogonal spaces: a WB space that encodes the degradation-agnostic invariant contents shared across degradations, and residual subspaces that adaptively preserve the degradation-specific knowledge. This disentanglement mitigates overfitting to in-distribution degradations and enables adaptive restoration grounded on the degradation-agnostic shared invariance. Extensive experiments demonstrate that BaryIR performs competitively against state-of-the-art all-in-one methods. Notably, BaryIR generalizes well to unseen degradations (\textit{e.g.,} types and levels) and shows remarkable robustness in learning generalized features, even when trained on limited degradation types and evaluated on real-world data with mixed degradations.
		
	\end{abstract}
	}
\maketitle
\section{Introduction}

\IEEEPARstart{I}{mage} restoration (IR) plays a fundamental role in low-level vision, aiming to recover the high-quality images given the degraded counterparts affected by various degradations (\textit{e.g.,} noise, blur, rain, haze, low light). Recent advances of deep neural networks (NNs) \cite{he2016deep, vaswani2017attention, dosovitskiy2021an, liu2021swin} have triggered remarkable successes in image restoration, in which most works \cite{lai2017deep, lai2018fast, zhang2018ffdnet, Zamir2021MPRNet, Zamir2021Restormer, liang2021swinir, chen2022simple, luo2023image, zhou2023fourmer} develop task-specific restoration networks to handle single known degradations. However, the task-specific nature of these approaches hinders their applicability in real-world scenarios such as autonomous navigation \cite{levinson2011towards, prakash2021multi} and surveillance systems \cite{liang2018deep}, where varied and unseen degradations frequently arise. Accordingly, a new paradigm has emerged, known as ``All-in-One'' Image Restoration (AIR) \cite{jiang2025survey, wu2025dswinir}, which seeks to address multiple forms of degradations within a single model.

To handle the AIR problem, most existing works \cite{li2022all, valanarasu2022transweather, zhang2023ingredient, potlapalli2023promptir, luocontrolling, tang2024residualconditioned, cui2025adair} train joint restoration models over multisource degraded–clean image pairs by incorporating degradation-specific guidance. A common strategy is to condition the unified restoration networks with degradation-specific cues, \textit{e.g.}, learnable prompts \cite{potlapalli2023promptir, luocontrolling, valanarasu2022transweather, liu24k, wu2025learning}, residual embeddings \cite{tang2024residualconditioned, tang2025degradation}, or frequency bands \cite{cui2025adair}, while other approaches \cite{conde2024instructir, zamfir2024efficient, ai2024lora} employ mixture-of-experts or adaptation modules,  in which degradation-specific parameters are selectively activated via routing mechanisms. Although these designs inject degradation-specific dynamics into restoration networks, they often struggle to capture degradation-agnostic features, which are crucial for modeling commonalities beyond training samples that leads to generalization. In contrast, another line of works \cite{zheng2024selective, zamfir2025complexity, zamfir2024efficient} attempt to learn degradation-agnostic features by reusing shared parameters, typically through a common branch or an agnostic expert across different degradations. However, such parameter-sharing strategy essentially still reduces to fitting multisource degraded–clean pairs with unified architectures, which inadequately capture the invariant geometry shared across multisource degraded images. Consequently, these methods at best learn features that appear agnostic across the degradations within the training data, rather than uncovering an intrinsic degradation-agnostic distribution, which tend to overfit to the training domain and thus are vulnerable to out-of-distribution (OOD) degradations.

\begin{figure}[!t]
	\centering
	\includegraphics[width=1\linewidth]{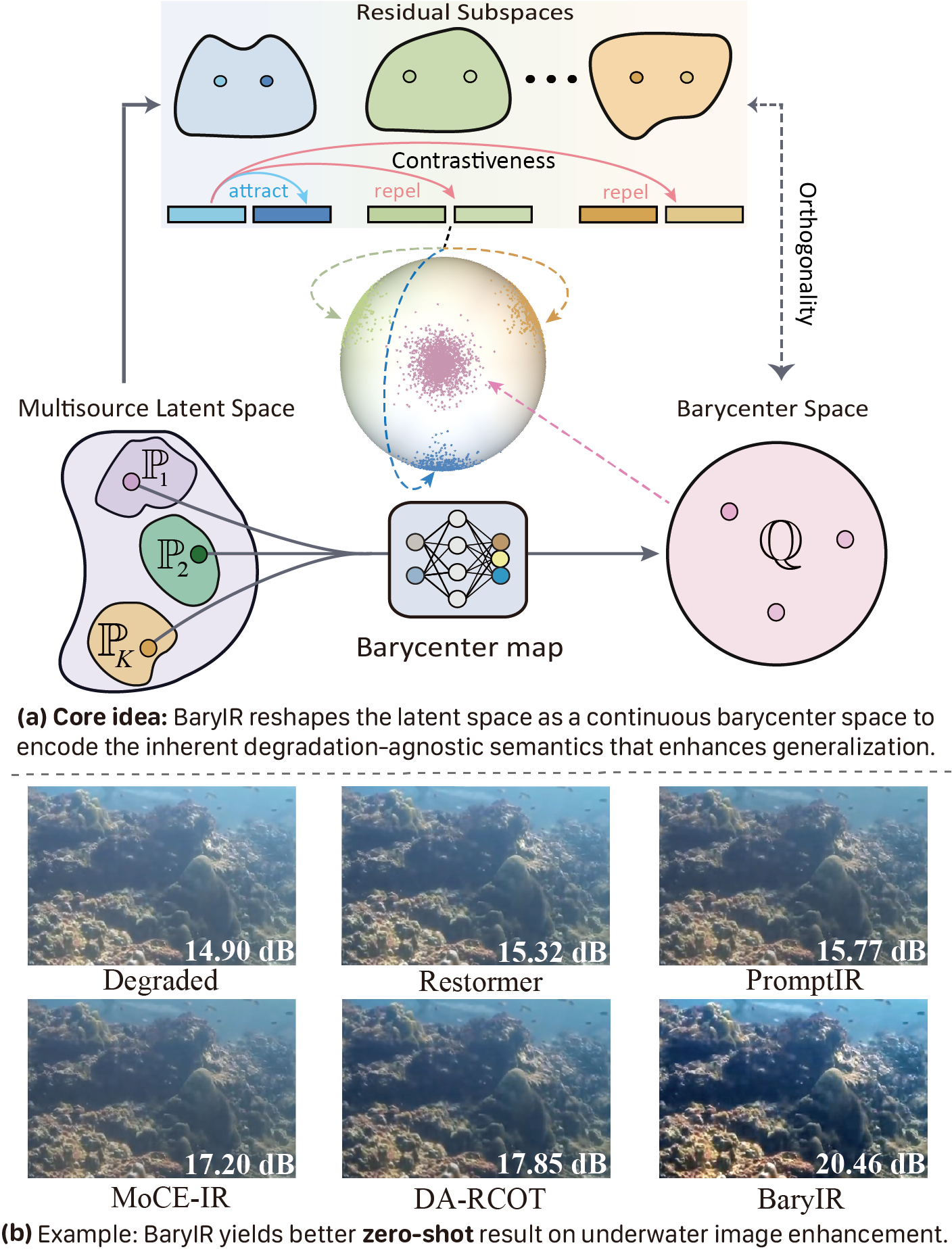}
	\caption{(a) BaryIR decouples the latent space of multisource degraded images into a barycenter space that captures the degradation-agnostic invariance, and residual subspaces that retain degradation-specific knowledge, enhancing generalization to unseen data. (b) A zero-shot example: BaryIR restores an  underwater image without being trained on such degradation. }
	\label{demo}
\end{figure}

To tackle the challenge, this work is motivated by the intuition that \textit{the multisource degraded feature distributions are induced by degradation-specific shifts from an underlying degradation-agnostic distribution}, which preserves the invariant structures across various forms of degraded images. Consequently, recovering this shared distribution is essential for ensuring generalization across degradations. With this insight, the core idea boils down to seeking a multisource joint embedding space that captures the degradation-agnostic invariant contents, even when training data covers only a limited set of degradation types, thereby providing a principled basis for generalization to unseen scenarios.

Specifically, we introduce BaryIR that learns a continuous barycenter map to transform the latent space of multisource degraded images into the Wasserstein barycenter (WB) space for generalized AIR. The WB space admits a distribution that minimizes the average of optimal transport (OT) distances to all types of degraded image embeddings while respecting the OT-grounded geometry. This property filters out degradation-specific factors within training domain and yields a degradation-agnostic embedding space that reduces the divergence  caused by different degradations.  Furthermore, we construct residual subspaces and apply contrastive loss over the residual embeddings (\textit{i.e.,} the gaps between multisource degraded image embeddings and WB embeddings) while enforcing their orthogonality to the WB embeddings (Fig. \ref{demo} (a)).  This design allows residual subspaces to adaptively preserve degradation-specific knowledge, while the WB space captures general degradation-agnostic invariance beyond training data, which alleviates overfitting and enables BaryIR to dynamically adjust its behavior for AIR (Fig. \ref{demo} (b)).

The key contributions are summarized as follows:
\begin{itemize}
	\item We present BaryIR, which explicitly constructs two orthogonal spaces for generalized AIR. The WB space encodes degradation-agnostic invariant contents, while the residual subspaces retain degradation-specific knowledge, which alleviates overfitting to in-domain degradations and enables adaptive restoration based on shared invariance.
	
	\item We advocate an max-min optimization algorithm to learn the NN-based barycenter map, yielding a continuous WB space that captures the fine-grained geometric structures of multisource data, which enhances BaryIR's ability in preserving visual patterns (\textit{e.g.,} colors and textures).
	
	\item We theoretically establish the error bounds for the NN-based barycenter map, providing approximation guarantees for the  recoverd barycenter distribution.
	
	\item Extensive experiments on both synthetic and real-world data show that BaryIR achieves state-of-the-art performance for AIR. Notably, BaryIR exhibits superior generalization to unseen degradations, as well as the robustness in learning generalized features with limited types of degradations.
\end{itemize}

\section{Related Work}
\subsection{All-in-One Image Restoration} 

Recently, AIR has emerged as a prominent low-level vision task that aims to address various degradations within a single restoration model. Pioneer methods typically utilize informative degradation embeddings \cite{li2022all, zamfir2024efficient, potlapalli2023promptir, luocontrolling, tang2024residualconditioned, chen2024learning, cui2025adair, chen2025unirestore} to guide the restoration. For instance, AirNet \cite{li2022all} trains an extra encoder using contrastive learning to extract degradation embeddings from degraded images.  PromptIR \cite{potlapalli2023promptir} and DA-CLIP \cite{luocontrolling} employ learnable visual prompts to encode the information of degradation type. DA-RCOT \cite{tang2025degradation} models AIR as an OT problem and leverages residual embeddings as adaptive conditions for degradation-aware restoration. Another line of works, \textit{e.g.,} InstructIR \cite{conde2024instructir}, DaAIR \cite{zamfir2024efficient}, Histoformer \cite{sun2024restoring},  route samples with different degradation patterns to specific experts or architectures for adaptive restoration. However, they are hardly generalizable due to the difficulty in capturing general commonality among degraded images. Differently, some intriguing works \cite{zhu2023learning, zamfir2024efficient, li2024foundir,zamfir2025complexity} attempt to model degradation-agnostic features by fitting multisource degraded–clean image pairs with a shared branch or expert. While effective on in-distribution data, they remain vulnerable to OOD degradations (\textit{e.g.,} unseen degradation types or levels) since they are seeking agnostic features within the training domain.   In contrast, BaryIR seeks degradation-agnostic invariance in the Wasserstein barycenter space, which models  the ``closest'' distribution to all multisource distributions and inherently alleviates overfitting to the training domain.

\subsection{Unified Representation Learning} Learning unified representations is a fundamental aspect of multisource representation learning. The majority of existing works aim to align diverse sources/modalities (\textit{e.g.,} text and images) within a shared latent space \cite{radford2021learning, sarkar2024xkd, andonian2022robust, xue2023ulip} or train a source-agnostic encoder to extract information across heterogeneous sources \cite{chen2020uniter, wang2022vlmixer}. Another line of works explores how to express the shared content from different domains with explicit unified representations, \textit{e.g.,} codebooks \cite{lu2023unifiedio, liu2022cross, ao2022speecht5} or prototypes \cite{yang2023prototypical, duan2022multi}. For example, Duan et al. \cite{duan2022multi} employ discrete OT to map the features extracted from different modalities to the prototypes. Despite their successes, these methods typically learn unified representations for only two sources, and scaling beyond two modalities or sources remains challenging. On the other hand, codebook-based approaches \cite{liu2022cross, duan2022multi} learn discrete spaces for unified representations, which hinders their ability to capture the fine-grained structures of multisource data. Differently, BaryIR learns a continuous barycenter space that is naturally scalable to an arbitrary source number by virtue of the OT barycenter formulation.

\section{Preliminaries}
\vspace{-0.1cm}
\textbf{Notation.} In this paper, we denote $\bar K=\{1,2,\dots,K\}$ for $K\in \mathbb N$. Given elements \( e_1, e_2, \ldots \) indexed by natural numbers, we denote the tuple \((e_1, e_2, \ldots, e_K)\) as \( e_{1:K} \). $\mathcal X\subset \mathbb R^d, \mathcal Y\subset \mathbb R^{d'},\mathcal X_k\subset\mathbb R^{d_k}$  are compact subsets of Euclidean space. $\mathcal C(\mathcal X)$ is the space of continuous functions on $\mathcal X$. The set of distributions on $\mathcal X$ is denoted by $\mathcal P(\mathcal X)$. For $\mathbb P \in \mathcal P(\mathcal X)$,  $\mathbb Q \in \mathcal P(\mathcal Y)$, the set of \textit{transport plans} is denoted as $\Pi(\mathbb P, \mathbb Q)$, \textit{i.e.,} probability distributions on $\mathcal X \times \mathcal Y$ with first and second marginals $\mathbb P$ and $\mathbb Q$. The pushforward of distribution $\mathbb P$ under some measurable map $T$ is denoted by $T_\#\mathbb P$. The Operator $\langle\cdot,\cdot\rangle$ denotes the cosine similarity that involves the normalization of features (on the unit sphere).
\begin{figure*}[!t]
	\centering
	\includegraphics[width=1\linewidth]{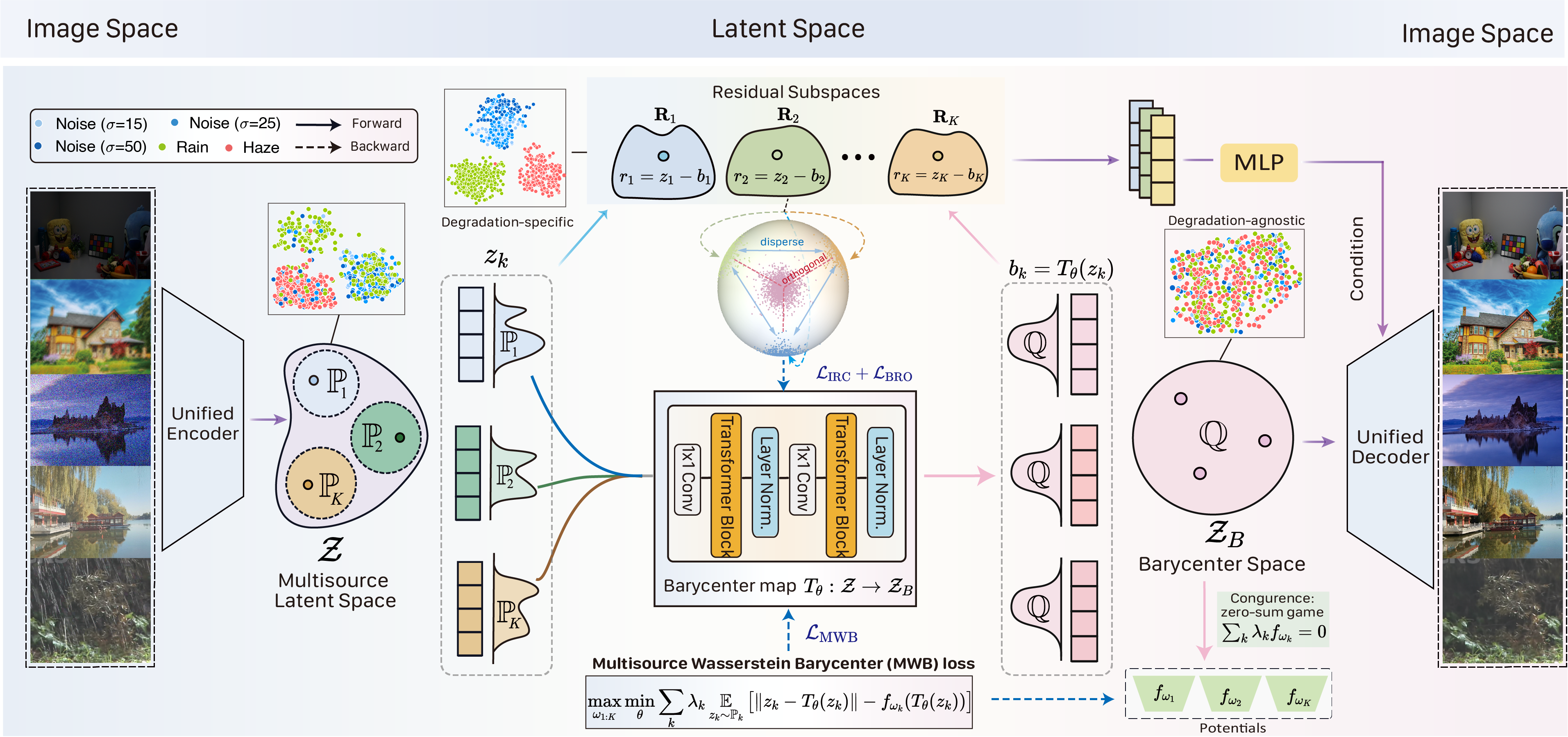}
	\caption{\textbf{Overview of BaryIR.} BaryIR learns a barycenter map to reshape the multisource latent space of degraded images into an inherent degradation-agnostic barycenter space beyond training domain.  Residual subspaces are constructed to retain the  degradation-specific knowledge, with the residual embeddings contrasted with each other while maintaining orthogonal to the WB embeddings. By integrating the WB and residual embeddings in the decoding layers, BaryIR captures the degradation-agnostic semantics while retaining adaptive degradation-specific knowledge, thereby enabling generalized restoration. }
	\label{pipeline}
	\vspace{-0.2cm}
\end{figure*}
\subsection{Optimal Transport}
\label{pre-ot}
Given two distributions $\mathbb P\in \mathcal P(\mathcal X)$ and $\mathbb Q\in\mathcal P(\mathcal Y)$ with a transport cost function $c: \mathcal X \times \mathcal Y \rightarrow \mathbb{R}_+$, the Kantorovich formulation \cite{kantorovich1942translocation} of the OT problem is defined as:
\begin{align}
	\text{OT}_c(\mathbb{P},\mathbb{Q})\triangleq\inf_{\pi\in\Pi(\mathbb P,\mathbb Q)}\mathbb E_{(x,y)\sim\pi}\left[c(x,y)\right].
	\label{Kon}
\end{align}
where $\pi\in\Pi(\mathbb P,\mathbb Q)$ is a transport plan. The choice of $c(x,y)=\|x-y\|$ yields  $W(\mathbb P,\mathbb Q)=\inf_{\pi\in\Pi(\mathbb P,\mathbb Q)}\int_{\mathcal X\times \mathcal Y}\|x-y\|d\pi(x,y)$, known as the 1-Wasserstein distance. The plan $\pi^*$ attaining the infimum is the \textit{OT plan}. Particularly, the choice of transport  cost $c(x,y)=\|x-y\|^2$ yields squared 2-Wasserstein distance $W^2_2(\mathbb P,\mathbb Q)$. The problem (\ref{Kon}) admits the dual form \cite{villani2009optimal}:
\begin{align}
	\mathrm{OT}_c(\mathbb{P},\mathbb{Q})=\sup_{f\in\mathcal{C}(\mathcal{Y})}\left\{\mathbb{E}_{x\sim\mathbb{P}}f^c(x)+\mathbb{E}_{y\sim\mathbb{Q}}f(y)\right\},
	\label{dual}
\end{align}
where $\displaystyle f^{c}(x)=\inf_{y\in \mathcal Y}\left[c(x,y)-f(y)\right]$ is the $c$-transform of the potential function $f\in\mathcal C(\mathcal Y)$.
\subsection{Wasserstein Barycenter}
Given distributions $\mathbb P_k\in\mathcal P(\mathcal X_k)$ for $k\in\bar K$ and transport costs $c_k:\mathcal X_k\times \mathcal Y\rightarrow \mathbb R_+$. For weights $\lambda_k>0$ with $\sum_{k=1}^K\lambda_k=1$, the classic OT barycenter problem seeks the distribution $\mathbb Q$ that attains the minimum of the weighted sum of OT problems with fixed first marginals $\mathbb P_{1:K}$:
\begin{align}
	\inf_{\mathbb Q\in\mathcal P(\mathcal Y)}\sum_{k=1}^{K}\lambda_k\text{OT}_{c_k}(\mathbb P_k, \mathbb Q). \label{classic-bary}
\end{align}
The choice of $c_k(x,y)=\|x-y\|$ yields the WB problem:
\begin{align}
	\inf_{\mathbb Q\in\mathcal P(\mathcal Y)}\sum_{k=1}^{K}\lambda_kW(\mathbb P_k, \mathbb Q). \label{WB}
\end{align}

In practice, given $N_k$ empirical samples $x_{1:N_k}^k\sim\mathbb P_k$ in a multisource space $\mathcal X=\cup_{k=1}^K\mathcal X_k$, the distributions $\mathbb P_{k}$ for $\mathcal X_k$ can be assessed using these empirical samples. Based on the Wasserstein barycenter problem (\ref{WB}), we can establish a map $ T:\mathcal X\rightarrow \mathcal Y$, which allows sampling points $T(x_k)$ from the approximate barycenter space with $x_k \sim \mathbb{P}_k$ as inputs. The setup leads to a continuous barycenter problem. Different from prior works \cite{li2020continuous, korotin2021continuous, chi2023variational} that model individual maps for each source, we seek the WB of multisource data by learning a unified NN-based barycenter map.

In the context of AIR, we apply the WB formulation to the multisource latent space of degraded image features $\mathcal Z$, where the features of $k$-th degradation type lie in a subspace $\mathcal Z_k$ and are distributed as $\mathbb P_k$. The WB encodes source-agnostic contents by modeling the distribution that minimizes the average of Wasserstein distances to distributions $\mathbb P_{1:K}$.
\section{Method}
Throughout the paper, we are motivated by the intuition that \textit{the multisource degraded feature distributions are induced by degradation-specific shifts from an underlying degradation-agnostic distribution} and thus recovering such shared distribution is essential for generalization across degradations. In this sense, we present BaryIR, a representation learning framework for generalized AIR. The key idea is to transform the multisource latent space  of degraded images into a barycenter space that captures the degradation-agnostic invariance across degradations. Building upon this invariant foundation, BaryIR further introduces residual subspaces that retain degradation-specific knowledge, enabling adaptive restoration across diverse degradations.

\textbf{Method overview.}  We first model the degradation-agnostic distribution for unified feature encoding with the Wasserstein barycenter (WB) formulation, in which a barycenter map is derived to transform the multisource latent space into the continuous barycenter space. We also establish the error bounds for the barycenter map under the dual OT framework (\S\ref{4.1}). Secondly, we construct the residual subspaces, where the residual embeddings are contrasted with each other while maintaining orthogonal to the WB embeddings (\S\ref{4.2}), which jointly contribute to the learning of WB space.   \S\ref{4.3} summarizes the restoration pipeline and optimization algorithm while \S\ref{4.4} presents the t-SNE visualization of barycenter and residual embeddings on unseen degradations. By integrating the WB and residual embeddings, BaryIR can capture the degradation-agnostic invariance while preserving degradation-specific knowledge for generalized AIR (Fig. \ref{pipeline}).

\subsection{Wasserstein Barycenter of Degraded Features}
\label{4.1}
Let $\mathbb{P}_k$ be the distribution of features $\bm z_k\in\mathcal Z_k\subset\mathcal Z$ for degradation type $k \in \bar K$, defined in the multisource latent space $\mathcal{Z}\subset\mathbb R^D$. The WB space is defined as $\mathcal Z_B:=\text{supp}(\mathbb Q)$ where $\mathbb Q$ denotes the WB distribution and  $\mathcal Z_B$ contains the barycenter features $\bm b$. Given the multisource degraded feature distributions $\mathbb{P}_{1:K}$, our goal is to establish the barycenter space $(\mathcal Z_B, \mathbb Q)$ and use it as the joint embedding space that encodes the degradation-agnostic semantics of multisource degraded images. Based on the WB formulation (\ref{WB}) over latent space, the WB  problem of degraded image features can be written as
\begin{align}
	\mathcal L_{\mathrm{MWB}}^*=\inf_{\mathbb Q\in\mathcal P(\mathcal Z_B)}\sum_{k=1}^{K}\lambda_kW(\mathbb P_k, \mathbb Q). \label{MWB}
\end{align} 

Given the challenge of directly solving the multisource Wasserstein barycenter problem (\ref{MWB}), we present its dual reformulation in Theorem \ref{the-bary}, which leads to the following sup-inf objective. This theorem enables us to compute the barycenters in a max-min optimization manner if the potentials $f_{1:K}\in \mathcal C(\mathcal Z_B)^K$ satisfy the \textit{congruence condition} $\sum_{k=1}^K\lambda_kf_k\equiv0$.
\begin{theorem}[\normalfont{Dual reformulation for multisource WB problem (\ref{MWB})}]The minimum objective value of the multisource WB problem (\ref{MWB}) $\mathcal L_{\mathrm{MWB}}^*$ can be expressed as \label{the-bary}	
	\begin{align}
		\mathcal L_{\mathrm{MWB}}^*&=\sup\limits_{\sum_k \lambda_k f_k = 0}\inf_{\substack{\\[0.1ex]\mathbb Q \in \mathcal P(\mathcal Z_B)}} \nonumber\\ &\sum_{k=1}^K\lambda_k\mathop{\mathbb E}_{\substack{\bm z_k\sim\mathbb P_k\\\bm b_k\sim\mathbb Q}}\big[\|\bm z_k-\bm b_k\|-f_k(\bm b_k)\big],
		\label{duality1}
	\end{align}
\end{theorem}
The proof can be found in the \textit{Appendix}. Here the supremum is taken over all the dual potentials $f_k:\mathcal Z_B\rightarrow\mathbb R$. We aim to learn the distribution $\mathbb Q$ by sampling the WB $\bm b_k=T(z_k)$ via a shared barycenter map $T:\mathcal Z\rightarrow\mathcal Z_B$. This can be achieved by replacing the optimization over $\bm b_k$ with an equivalent optimization (Rockafellar interchange theorem \cite{rockafellar1976integral}, Theorem 3A) over $T$, yielding the following objective:
\begin{align}
	\mathcal L_{\mathrm{MWB}}^*=&\sup\limits_{\sum_k \lambda_k f_k = 0}\inf_{\substack{\\[0.1ex]T:\mathcal Z\rightarrow\mathcal Z_B}}\bigg\{\mathcal L_{\mathrm{MWB}}(f_{1:K},T)\nonumber\\ \triangleq\sum_{k=1}^K&\lambda_k\mathop{\mathbb E}_{\bm z_k\sim\mathbb P_k}\big[\|\bm z_k-T(\bm z_k)\|-f_k(T(\bm z_k))\big]\bigg\},
	\label{duality2}
\end{align}
\textbf{On the parameterization.} In practice, we parameterize $T_\theta$ and $f_{1:K}$ with two groups of neural networks $T_\theta$ and $f_{\omega_{1:K}}$. 
The barycenter map $T_\theta$ incorporates two gating-based transformer blocks, which are composed of two main sub-modules: the Multi-Dconv Head Transposed Attention (MDTA) and the Gated-Dconv Feedforward Network (GDFN) \cite{Zamir2021Restormer}. The MDTA incorporates depth-wise convolutions to better capture local structural patterns in images, while the GDFN employs a gating strategy that filters out uninformative responses, ensuring that the most relevant features are propagated. 
For the family of potentials $f_{\omega_{1:K}}$, we parameterize $f_{\omega_k}$ as $g_{\omega_k}-\sum_{i=1}^K\lambda_ig_{\omega_i}$ with MLPs $g_{\omega_k}:\mathbb R^D\rightarrow\mathbb R$ to guarantee congruent constraint $\sum_{k=1}^K\lambda_kf_k$, which is a common trick used in \cite{li2020continuous,kolesov2024estimating,kolesov2024energy}.

\quad\\
\noindent\textbf{Multisource Wasserstein Barycenter (MWB) loss.} With this parameterization, we obtain the following MWB loss, which can be optimized in a max-min adversarial manner to  compute the barycenter map $T_\theta$ for approximating the WB space.
\begin{align}
	\max_{\omega_{1:K}}\min_{\theta}&\bigg\{\mathcal L_{\mathrm{MWB}}(\omega_{1:K},\theta)\triangleq\nonumber\\\sum_{k=1}^K &\lambda_k\mathop{\mathbb E}_{\bm z_k\sim\mathbb P_k}\big[\|\bm z_k-T_\theta(\bm z_k)\|-f_{\omega_k}(T_\theta(\bm z_k))\big]\bigg\}.
	\label{Lbary}
\end{align}
To tackle this optimization problem, we train the networks $T_\theta$ and $f_{\omega_{1:K}}$ in an alternating manner: maximizing \textit{w.r.t.} $\omega_{1:K}$ while minimizing \textit{w.r.t.} $\theta$ under the MWB loss. At each step, the expectations are estimated using sampled mini-batches. 

\quad\\
\noindent\textbf{Error bounds.} Let $\widehat T$ denote the barycenter map that approximately solves (\ref{Lbary}). A natural question is how close $\widehat T$ is to the true barycenter map $T^*$, which transports each $\mathbb P_k$ to the barycenter distribution $\mathbb Q^*$. Theorem \ref{error} establishes an error bound for the estimated barycenter map, showing that for the pair $(\widehat f_{1:K}, \widehat T)$ solving the optimization problem (\ref{Lbary}), the recovered map $\widehat T$ remains close to the true barycenter map $T^*$. For convenience, we set simplified notation as follows:
\begin{align}&\mathcal F(f_{1:K},T):=\mathcal L_{\mathrm{MWB}}(f_{1:K},T), \label{func1}\\
	\mathcal L(f_{1:K}):=&\inf_{T:\mathcal Z\rightarrow\mathcal Z_B}\mathcal F(f_{1:K},T)~~\text{and}~~\mathcal L^*:=\mathcal L_{\mathrm{MWB}}^*. \label{func2} \end{align}
\begin{theorem}[\normalfont{Error analysis via duality gaps for the recovered maps}] Let $C_k$ be any transport costs (not only Euclidean included). 
	Assume that the maps $\bm b_k\to C_k(\bm z_k,\bm b_k)-\widehat f_k(\bm b_k)$ are $\beta$-strongly convex for $\bm z_k\in\mathcal Z_k$, $k\in\bar K$. Consider the duality gaps for an approximate solution $(\widehat f_{1:K},\widehat T)$ of {\normalfont(\ref{duality2})}:
	\begin{align}
		\mathcal E_1(\widehat{f}_{1:K}, \widehat{T}) &\triangleq \mathcal F(\widehat{f}_{1:K}, \widehat T) - \mathcal L(\widehat{f}_{1:K}); \label{eq:deltagap1} \\
		\mathcal E_2(\widehat{f}_{1:K}) &\triangleq \mathcal L^* - \mathcal L(\widehat{f}_{1:K}), \label{eq:deltagap2}
	\end{align}
	which are the errors of solving the inner $\inf$ and outer $\sup$ problems in (\ref{duality2}). Then the following inequality holds: \label{error}
	\begin{align}
		\sum_{k=1}^{K}\lambda_{k}W_{2}^{2}\left(\widehat T_{\#}\mathbb P_k,T^{*}_\#\mathbb P_k\right)\leq\frac{4}{\beta}(\mathcal E_1+\mathcal E_2).\nonumber
	\end{align}
	Here $W^2_2(\mathbb P,\mathbb Q)$ is the squared 2-Wasserstein distance as denoted in Preliminary (Sec. \ref{pre-ot}).
\end{theorem}
\subsection{Disentangled WB and Residual Feature Space Learning}
\label{4.2}
We construct the residual subspace $\mathbf R_k$ for the $k$-th type degradation, by defining its elements as residual embeddings $\bm r_k=\bm z_k-\bm b_k$, in which $\bm r_k\in\mathbf R_k,k=1,\ldots,K$. According to its definition, the residual embeddings naturally retains the  information discarded by WB embeddings and capture adaptive degradation-specific knowledge across degraded images.

To learn disentangled WB and residual feature spaces, we further introduce two regularization terms: 1) Inter-residual contrastive loss, promoting similarity within the same residual subspace and dissimilarity across different subspaces, and 2) Barycenter-residual orthogonal loss, which enforces orthogonality between the WB and residual embeddings.  The first term augments the degradation-specific semantics in the residual subspaces while the second term encourages the disentanglement of the degradation-agnostic contents in WB space from the degradation-specific knowledge in residual subspaces.

\quad\\
\noindent\textbf{Inter-residual contrastive (IRC) loss.}  
The IRC loss encourages the learning of residual embeddings with separated semantics while preserving maximal information across degradations. Formally, for any residual embedding $\bm r_k\in\mathbf R_k$ in a batch $B$, we treat the embeddings from the same subspace  within this batch as positive samples $r_{k}^+$. The negative samples $\bm r_k^{-}$ are embeddings from other residual subspaces $\mathbf R_i$, where $i\not=k$. By letting $\bm r_k$ attract positive samples and repel the negative ones, the IRC loss is formulated as

\begin{small}
	\begin{align}
		\mathcal L_{\mathrm{IRC}}\triangleq-\sum_{\bm r_k\in B}\log\frac{\displaystyle\sum_{\bm r_{k}^+\in B}\exp(\langle \bm r_k,\bm r_{k}^+\rangle/\tau)}{\displaystyle\sum_{\bm r_{k}^+\in B}\exp(\frac{\langle \bm r_k,\bm r_{k}^+\rangle}{\tau})+\sum_{\bm r_{k}^-\in B}\exp(\frac{\langle \bm r_k,\bm r_{k}^-\rangle}{\tau})},
	\end{align}
\end{small}

\noindent where $\langle \cdot,\cdot\rangle$ is the cosine similarity and $\tau$ is the temperature hyper-parameter.  During training, the residual embeddings are sampled as $\bm r_k=\bm z_k-T_\theta(\bm z_k)$, and the IRC loss $\mathcal L_{\mathrm{IRC}}$ contributes to the optimization of the barycenter map $T_\theta$.

\quad\\
\noindent\textbf{Barycenter-residual orthogonal (BRO) loss.} The BRO loss encourages the orthogonality between the WB space and all residual subspaces, which is defined as
\begin{align}
	\mathcal L_{\mathrm{BRO}}\triangleq\sum_{\bm b_k\in B}\sum_{\bm r_j\in B}\langle \bm b_k, \bm r_j \rangle^2,
\end{align}
The BRO loss penalizes the inner product between WB and all residual embeddings, encouraging their orthogonality to ensure disentangled representation learning.

\subsection{Restoration Pipeline and Optimization Algorithm}
\label{4.3}
As presented in Fig. \ref{pipeline}, we approximate the WB space from the multisource latent space by adversarially training the barycenter map $T_\theta$ against potentials $f_{\omega_{1:K}}$. Then the residual subspaces are constructed to retain degradation-specific knowledge, in which the residual embeddings are contrasted with each other while maintaining orthogonal to the WB embeddings. This process reduces to optimizing the barycenter map $T_\theta$ with the training objective $\mathcal L_{Bary}$, which combines the proposed losses $\mathcal L_{\mathrm{MWB}}$, $\mathcal L_{\mathrm{IRC}}$, and $\mathcal L_{\mathrm{BRO}}$:
\begin{align}
	\mathcal L_{Bary}(\omega_{1:K},\theta) =\mathcal L_{\mathrm{MWB}}(\omega_{1:K},\theta)  +\alpha(\mathcal L_{\mathrm{IRC}}+\mathcal L_{\mathrm{BRO}})(\theta).
	\label{baryloss}
\end{align}
This procedure is  summarized as Algorithm \ref{algo}, which solves the barycenter map $T_\theta$ to obtain the WB and residual embeddings. Subsequently, BaryIR can restore clean images from any degraded input by integrating these two types of embeddings  in the decoding layers (Fig. \ref{pipeline}). The restoration is supervised by an $L_1$ loss that minimizes the difference between the restored image and the ground truth. This loss minimization is performed simultaneously with the minimization of $\mathcal L_{Bary}$ through a single backward propagation, as $T_\theta$ is incorporated as a sub-module within the restoration network.

\begin{algorithm}[!t]
	\label{Algo}
	\caption{Barycenter map solver for computing WB.}  
	\textbf{Input}:  Multisource distribution $\mathbb P_k$ accessible by encoding degraded samples; NN-based barycenter map: $T_\theta$; NN-based potentials: $f_{\omega_{1:K}}$; the inner iteration number $n_T$.\\
	\textbf{Output}: NN $T_{\theta^*}$ approximating the barycenter map between $\mathbb P_k$ and WB $\mathbb Q^*$.
	\begin{algorithmic}[1]
		\label{algo}
		\WHILE{$\theta$ has not converged}
		\STATE Sample batches $B$ with $\bm z_k\sim\mathbb P_k,k\in\bar K$;
		\STATE $\mathcal L^f_{Bary}\leftarrow\frac{1}{|B|}\sum_{\bm z_k\in B}\lambda_kf_{\omega_k}(T_\theta(\bm z_k))$; 
		\STATE	Update $\omega_{1:K}$ by using $\frac{\partial \mathcal L^f_{Bary}}{\partial \omega_{1:K}}$;
		\FOR{$t=0,\cdots,n_T$}
		\STATE Sample batches $B$ with $\bm z_k\sim\mathbb P_k,k\in\bar K$;
		\STATE $\mathcal L^T_{Bary}\leftarrow\frac{1}{|B|}\bigg\{\sum_{\bm z_k\in B}\lambda_k\big[\|\bm z_k-T_\theta(\bm z_k)\|-f_{\omega_k}(T_\theta(\bm z_k))\big]+\alpha(\mathcal L_{\mathrm{IRC}}+\mathcal L_{\mathrm{BRO}})(\theta)\bigg\}$;
		\STATE	Update $T_\theta$ by using $\frac{\partial \mathcal L^T_{Bary}}{\partial\theta}$;
		\ENDFOR
		\ENDWHILE
	\end{algorithmic} 
\end{algorithm}

\subsection{t-SNE Visualization of WB and Residual Embeddings}
\label{4.4}
\begin{figure*}[!t]
	\centering
	\includegraphics[width=1\linewidth]{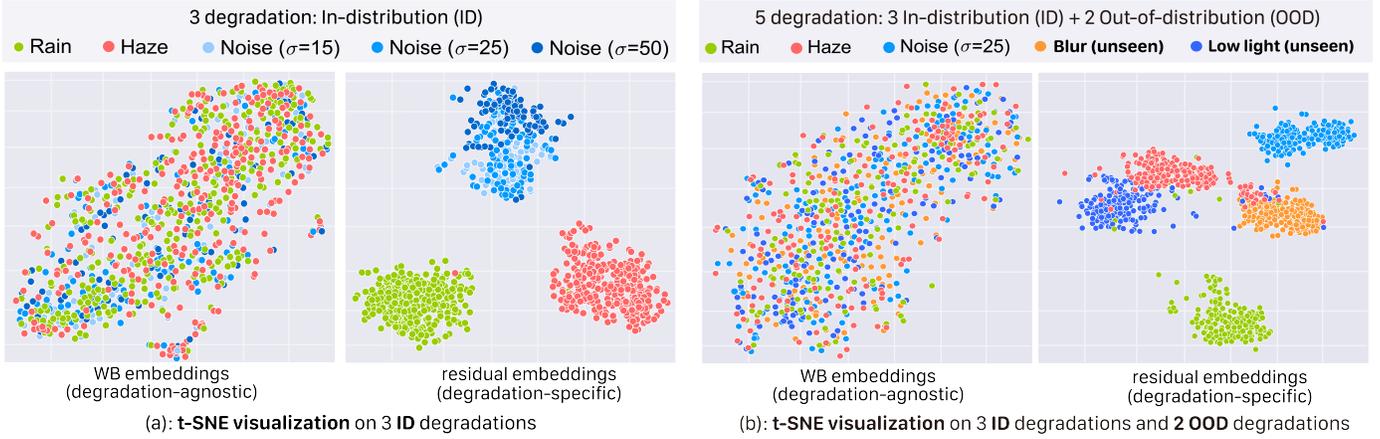}
	\caption{t-SNE visualization of WB and residual embeddings with BaryIR trained with three degradations (\textit{i.e.,} rain, haze and noise). The WB embeddings exhibit degradation-agnostic distribution while the residual embeddings are clearly separated according to the specific degradations. Notably, the WB and residual embeddings remain robust on unseen degradations (\textit{i.e.,} blur and low light) for capturing degradation-agnostic and degradation-specific semantics. }
	\label{WBvis}
\end{figure*}
To understand the semantic nature of the WB and residual embeddings in BaryIR, we employ t-SNE for visualizing the embeddings. The model is trained with three types of degradation: rain, haze, and noise, using 300 images for each degradation type. We also test the embeddings on the five-degradation setting with unseen degradation types such as blur and low light. The WB and residual embeddings are extracted for each image and visualized using t-SNE. Embeddings from different degradation types are color-coded for clarity.

As shown in Fig. \ref{WBvis}, the t-SNE visualization reveals that the WB embeddings exhibit a degradation-agnostic distribution, meaning they remain clustered regardless of the specific degradation type. In contrast, the residual embeddings are clearly separated according to specific degradations (rain, haze, and noise), reflecting their ability to capture degradation-specific semantics. Notably, when tested on unseen degradations such as blur and low light, the WB embeddings still show a degradation-agnostic distribution, while the residual embeddings remain distinct and separated according to the specific type of degradation. This demonstrates that BaryIR effectively captures both degradation-agnostic invariance and adaptive degradation-specific knowledge across degradations, making it robust to OOD degradations beyond training samples.
\section{Experiments}
\label{exp}
We evaluate BaryIR for all-in-one restoration on benchmark datasets, which cover comparisons with state-of-the-art methods on both in-distribution data across five degradations and OOD data. The OOD evaluations include synthetic-to-real generalization and generalization to OOD degradation types and levels.   We also evaluate its generalization performance on synthetic and real-world scenarios with mixed degradations.  We use PSNR/SSIM for measuring pixel-wise similarity, LPIPS \cite{zhang2018unreasonable}/FID \cite{heusel2017gans} for measuring perceptual deviation, and two  non-reference indexes NIQE \cite{mittal2012making} and PIQE  \cite{venkatanath2015blind} to assess real-world multiple-degradation images. The best and second-best are \textbf{highlighted} and \underline{underlined} respectively. 

\quad\\
\noindent\textbf{Implementation details.} We train our models using the RMSProp optimizer with a learning rate of $1\times10^{-4}$ for the restoration network which includes barycenter network $T_\theta$ as its sub-module, and $2\times10^{-4}$ for the potentials $f_{\omega_k}$. Besides the training of the barycenter map, we adopt an end-to-end pairwise training using $L_1$ loss for the overall framework. The learning rate is decayed by a factor of 10 after specific epochs.  The temperature hyper-parameter $\tau$ is empirically set as 0.07. The inner iteration number $n_T$ is set to be 1. The barycenter weights $\lambda_k$ follow the proportion of the number of training samples for each source, and the trade-off parameter in (\ref{baryloss}) is set as $\alpha=0.05$. The restoration backbone network (containing the encoder and decoder) is implemented with Restormer \cite{Zamir2021Restormer}. During training, we crop 128$\times$128 patches as inputs. All the experiments are conducted on Pytorch 2.1.0 with an NVIDIA 4090 GPU. The FID scores are computed using 256$\times$256 center-cropped patches. The residual embeddings are upscaled via MLPs as the conditions at different scales for the decoder. $f_{\omega_{1:K}}$ consists of $K$ parallel MLPs (where $K$ is the degradation number of training samples). Each branch of $f_{\omega_{1:K}}$ employs independent parameters $\omega_k,k\in\bar K$.

\quad\\
\noindent\textbf{Training datasets.} We train the BaryIR on benchmark datasets covering both synthetic and real-world data. For denoising, we merge  BSD400 \cite{arbelaez2010contour} and WED \cite{ma2016waterloo} datasets, adding Gaussian noise with levels $\sigma\in\{15,25,50\}$. Testing is conducted on the BSD68 \cite{martin2001database} datasets. We use the Rain100L \cite{yang2017deep} for deraining, and  SOTS \cite{li2018benchmarking} for dehazing. The deblurring and low-light enhancement tasks leverage real-world datasets GoPro \cite{nah2017deep} and LOL-v1 \cite{wei2018deep}, respectively. For the All-in-One configuration, we merge these datasets into a mixed one with three or five degradation types for training a unified model.
\subsection{All-in-One Restoration Results}
\label{5.1}
\begin{table*}[!htbp]
	\centering
	
	\caption{The All-in-One comparison of our BaryIR with the state-of-the-art methods on \textbf{three} degradations. }
	\label{air3}
	\vspace{-0.2cm}
	\setlength{\tabcolsep}{7pt}
	\renewcommand{\arraystretch}{1.0}
	\resizebox{\textwidth}{!}{
		\begin{tabular}{llcccccccccccc}
			\toprule
			\multirow{4}{*}{Method} & \multirow{4}{*}{Venue}&
			\multicolumn{2}{c}{\textit{Dehazing}} & \multicolumn{2}{c}{\textit{Deraining}} & \multicolumn{6}{c}{\textit{Denoising}} 
			& \multicolumn{2}{c}{\multirow{2}{*}{Average}}  \\  \cmidrule(lr){3-4} \cmidrule(lr){5-6} \cmidrule(lr){7-12}
			&& \multicolumn{2}{c}{SOTS} & \multicolumn{2}{c}{Rain100L} & \multicolumn{2}{c}{BSD68\textsubscript{$\sigma$=15}} & \multicolumn{2}{c}{BSD68\textsubscript{$\sigma$=25}} & \multicolumn{2}{c}{BSD68\textsubscript{$\sigma$=50}}  &  & \\  \cmidrule(lr){3-4} \cmidrule(lr){5-6} \cmidrule(lr){7-12} \cmidrule(lr){13-14}
			&& PSNR & SSIM & PSNR & SSIM & PSNR & SSIM & PSNR & SSIM & PSNR & SSIM & PSNR & SSIM \\
			\midrule

			Restormer \cite{Zamir2021Restormer}  &CVPR'22  & 29.92 & 0.970 &  35.64 & 0.971 &  33.81 & 0.932 &  31.00 & 0.880 & 27.85 & 0.792 & 31.62 & 0.909 \\
			PromptIR \cite{potlapalli2023promptir}&NeurIPS'23& 30.58 & 0.974 & 36.37 & 0.972& 33.97 & 0.933& 31.29 & 0.888 & 28.06 & 0.798 & 32.05 & 0.913  \\
			\midrule
			DA-CLIP \cite{luocontrolling}&ICLR'24 & 30.12 & 0.972& 35.92 & 0.972& 33.86 & 0.925& 31.06 & 0.865 &  27.55 & 0.778 & 31.70 & 0.901  \\
			DiffUIR \cite{zheng2024selective}&CVPR'24  &30.18 & 0.973& 36.78 & 0.973& 33.94 & 0.932& 31.26 & 0.887 & 28.04 & 0.797 & 32.04 & 0.912 \\
			InstructIR \cite{conde2024instructir}&ECCV'24 &30.22 & 0.959& 37.98 & 0.978& {34.15} & 0.933 & {31.52} & {0.890}& \underline{28.30} & \underline{0.804} & 32.43 & 0.913 \\
			AdaIR \cite{cui2025adair}&ICLR'25 &31.06 & \underline{0.980} & {38.64} & 0.983& 34.12 & \underline{0.935} & 31.45 & \underline{0.892} & 28.19 & {0.802} & 32.69 & {0.918} \\
			MoCE-IR \cite{zamfir2025complexity}&CVPR'25 &{31.34} & {0.979} & 38.57 & {0.984}& 34.11 & 0.932 & 31.45 & 0.888& 28.18 & 0.800 & {32.73} & 0.917 \\
			DA-RCOT \cite{tang2025degradation}&TPAMI'25 &31.26 & 0.977 & 38.36 & 0.983& 33.98 & {0.934} & 31.33 & {0.890} & 28.10 & 0.801 & 32.60 & 0.917 \\
			\midrule
			BaryIR$_{\text{Restormer}}$ & Ours&\underline{31.40} & \underline{0.980} & \underline{39.02} & \underline{0.985}& \underline{34.16} & \underline{0.935}& \underline{31.54} & \underline{0.892} & {28.25} & {0.802} & \underline{32.86} & \underline{0.919}  \\
			BaryIR$_{\text{PromptIR}}$ & Ours&\textbf{31.95} & \textbf{0.988} & \textbf{39.36} & \textbf{0.988}& \textbf{34.30} & \textbf{0.936}& \textbf{31.61} & \textbf{0.893} & \textbf{28.45} & \textbf{0.806} & \textbf{33.13} & \textbf{0.923}  \\
			\bottomrule
	\end{tabular}}
	
	\vspace{5pt}
	
	\caption{The All-in-One comparison of our BaryIR with the state-of-the-art methods on \textbf{five} degradations.}
	\label{air5}
	\vspace{-0.2cm}
	\setlength{\tabcolsep}{7pt}
	\renewcommand{\arraystretch}{1.0}
	\resizebox{\textwidth}{!}{
		\begin{tabular}{llcccccccccccc}
			\toprule
			\multirow{4}{*}{Method} & \multirow{4}{*}{Venue}
			& \multicolumn{2}{c}{\textit{Dehazing}} & \multicolumn{2}{c}{\textit{Deraining}} & 
			\multicolumn{2}{c}{\textit{Denoising}} & \multicolumn{2}{c}{\textit{Deblurring}} & \multicolumn{2}{c}{\textit{Low-light}}
			& \multicolumn{2}{c}{\multirow{2}{*}{Average}}  \\  \cmidrule(lr){3-4} \cmidrule(lr){5-6} \cmidrule(lr){7-8} \cmidrule(lr){9-10} \cmidrule(lr){11-12}
			&& \multicolumn{2}{c}{SOTS} & \multicolumn{2}{c}{Rain100L} &  \multicolumn{2}{c}{BSD68\textsubscript{$\sigma$=25}} & \multicolumn{2}{c}{GoPro} & \multicolumn{2}{c}{LOL-v1} && \\  \cmidrule(lr){3-4} \cmidrule(lr){5-6} \cmidrule(lr){7-8} \cmidrule(lr){9-10} \cmidrule(lr){11-12} \cmidrule(lr){13-14} 
			&& PSNR & SSIM & PSNR & SSIM & PSNR & SSIM & PSNR & SSIM & PSNR & SSIM & PSNR & SSIM\\
			\midrule
			
			Restormer \cite{Zamir2021Restormer} &CVPR'22  & 24.09 & 0.927 &  34.81 & 0.971  &  30.78 & 0.876 &  27.22 & 0.829 &   20.41 & 0.806 & 27.46 & 0.881 \\
			PromptIR \cite{potlapalli2023promptir} &NeurIPS'23 &30.41 & 0.972 & 36.17 & 0.970 & 31.20 & 0.885 & 27.93 & 0.851 & 22.89 & 0.829 &29.72 & 0.901 \\
			\midrule
			DA-CLIP \cite{luocontrolling} &ICLR'24 &29.78 & 0.968 &35.65 & 0.962 &30.93 & 0.885 &27.31 & 0.838 &21.66 & 0.828 &29.07 & 0.896 \\
			DiffUIR \cite{zheng2024selective} &CVPR'24  &29.47 & 0.965 &35.98 & 0.968 &31.02 & 0.885 &27.50 & 0.845 &22.32 & 0.826 &29.25 & 0.898 \\
			InstructIR \cite{conde2024instructir} &ECCV'24  &27.10 & 0.956 &36.84 & 0.973 &31.40 & {0.890} &29.40 & 0.886 &23.00 & 0.836 &29.55 & 0.908 \\
			AdaIR \cite{cui2025adair} &ICLR'25  &30.54 & {0.978} &38.02 & {0.981} &{31.35} & 0.889 &28.12 & 0.858 &23.00 & 0.845 &30.20 & 0.910 \\
			MoCE-IR \cite{zamfir2025complexity} &CVPR'25 &30.48 & 0.974 &{38.04} & \underline{0.982} &31.34 & 0.887 &\textbf{30.05} & \textbf{0.899} &23.00 & {0.852} &30.25 & \underline{0.919} \\
			DA-RCOT \cite{tang2025degradation} &TPAMI'25 &{30.96} & 0.975 &37.87 & 0.980 &31.23 & 0.888 &28.68 & 0.872 &{23.25} & 0.836 &{30.40} & 0.911 \\
			\midrule
			BaryIR$_{\text{Restormer}}$ & Ours & \underline{31.20} & \underline{0.979} &\underline{38.10} & \underline{0.982} & \underline{31.43} & \underline{0.891} &{29.51} & {0.889} & \underline{23.37} & \underline{0.854} &\underline{30.72} & \underline{0.919} \\
			BaryIR$_{\text{PromptIR}}$ & Ours&\textbf{31.68} & \textbf{0.980} & \textbf{38.36} & \textbf{0.984}& \textbf{31.49} & \textbf{0.894}& \underline{29.84} & \underline{0.895} & \textbf{23.88} & \textbf{0.862} & \textbf{31.05} & \textbf{0.923}  \\
			\bottomrule
			
	\end{tabular}}
	
	\vspace{10pt}
	
	\setlength\tabcolsep{1pt}
	\renewcommand{\arraystretch}{0.7} 
	\centering
	\begin{tabular}{cccccc}
		\includegraphics[width=0.16\linewidth]{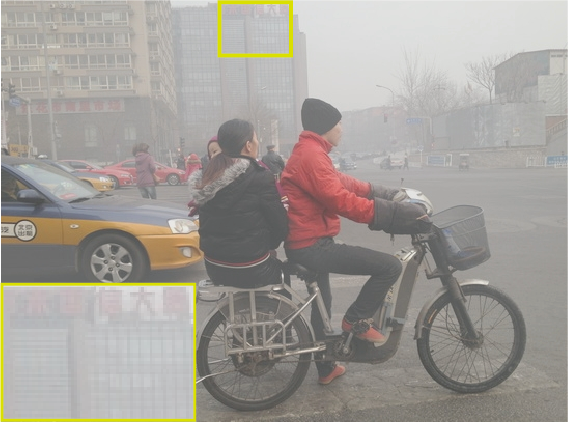}&
		\includegraphics[width=0.16\linewidth]{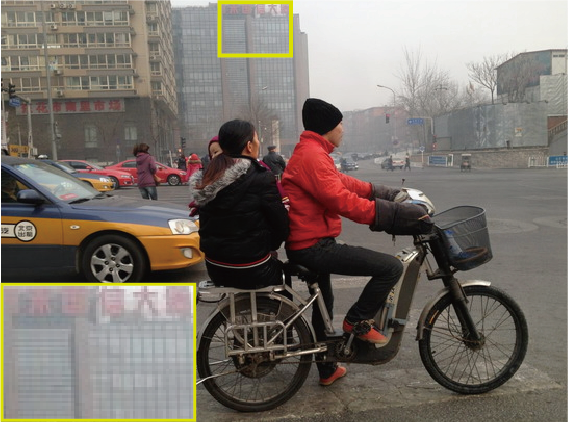}&
		\includegraphics[width=0.16\linewidth]{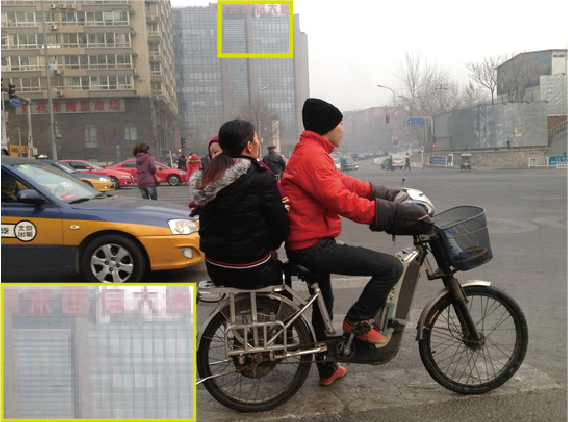}&
		\includegraphics[width=0.16\linewidth]{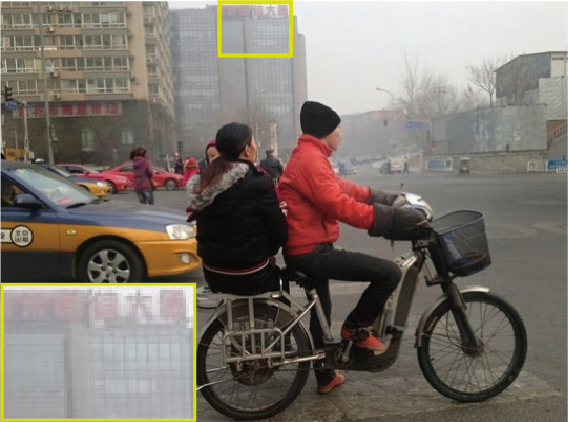}&
		\includegraphics[width=0.16\linewidth]{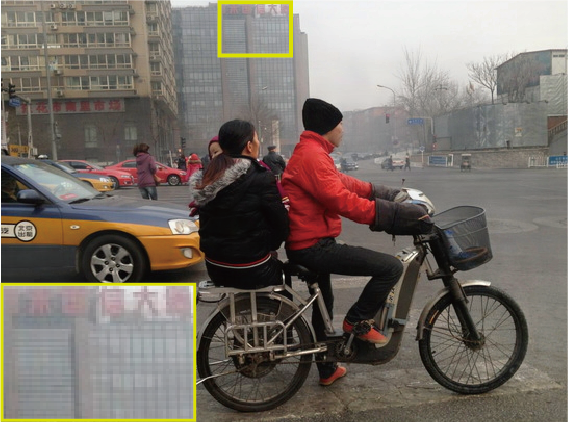}&
		\includegraphics[width=0.16\linewidth]{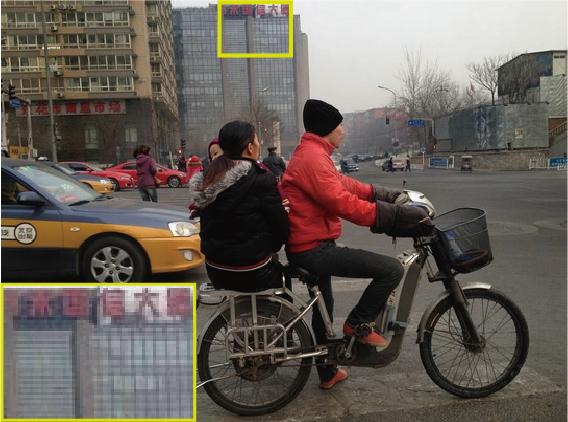}\\
		\includegraphics[width=0.16\linewidth]{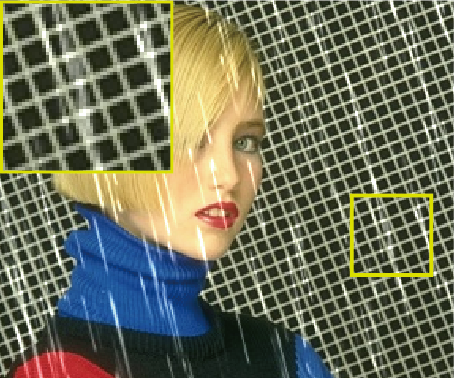}&
		\includegraphics[width=0.16\linewidth]{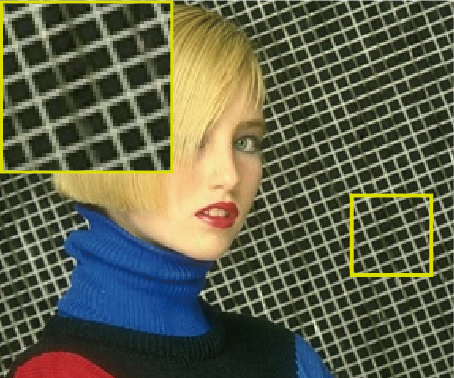}&
		\includegraphics[width=0.16\linewidth]{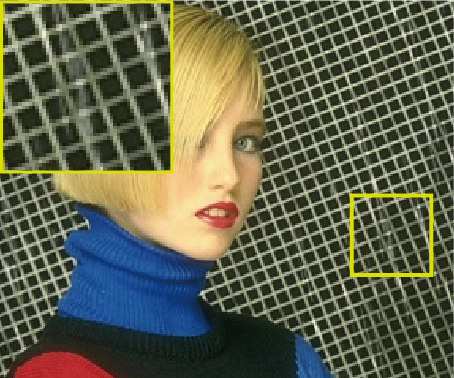}&
		\includegraphics[width=0.16\linewidth]{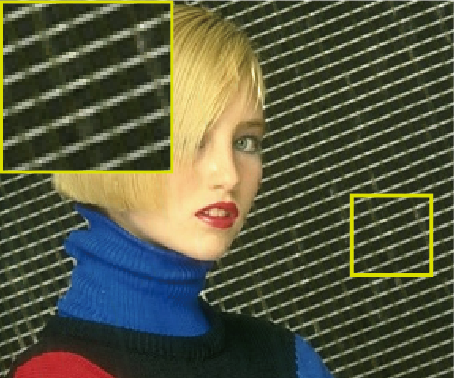}&
		\includegraphics[width=0.16\linewidth]{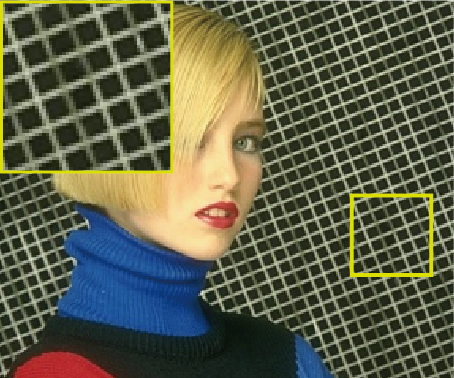}&
		\includegraphics[width=0.16\linewidth]{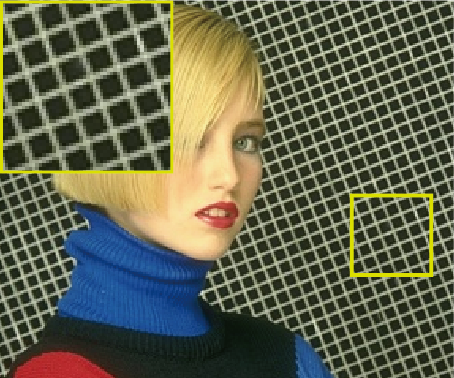}\\
		\includegraphics[width=0.16\linewidth]{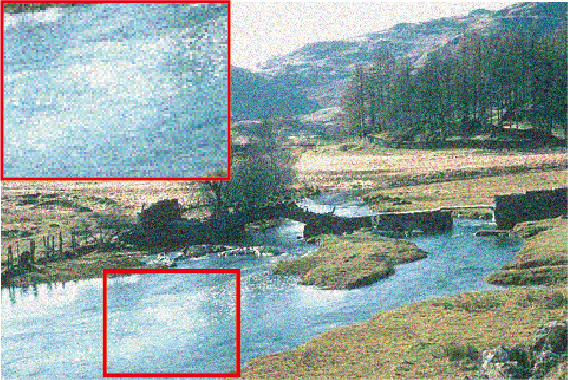}&
		\includegraphics[width=0.16\linewidth]{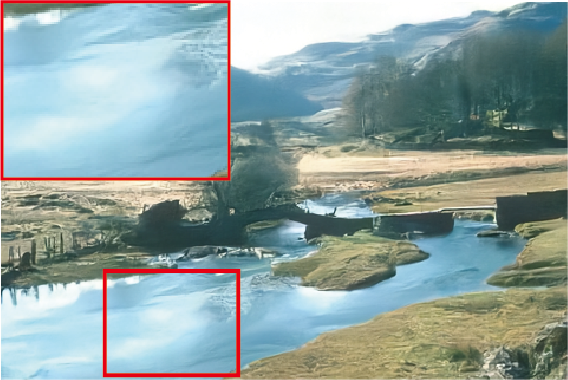}&
		\includegraphics[width=0.16\linewidth]{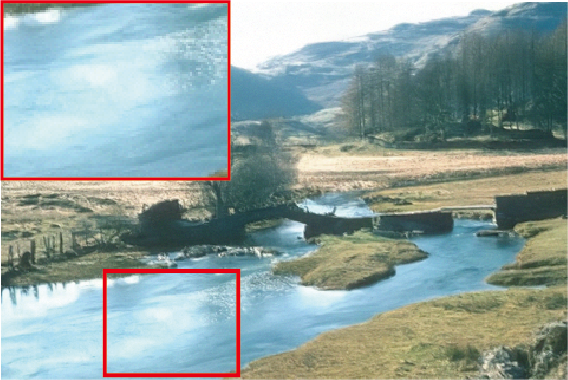}&
		\includegraphics[width=0.16\linewidth]{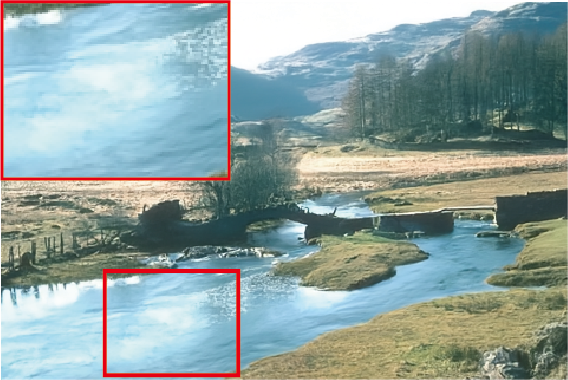}&
		\includegraphics[width=0.16\linewidth]{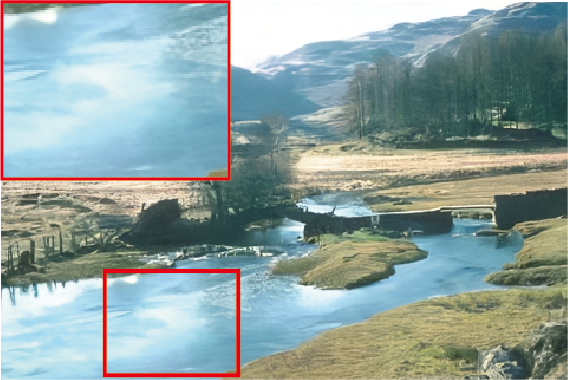}&
		\includegraphics[width=0.16\linewidth]{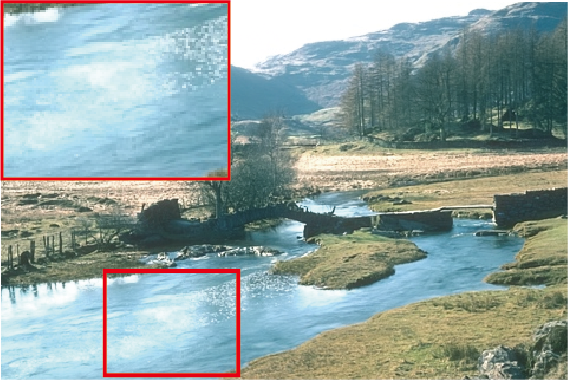}\\
		\includegraphics[width=0.16\linewidth]{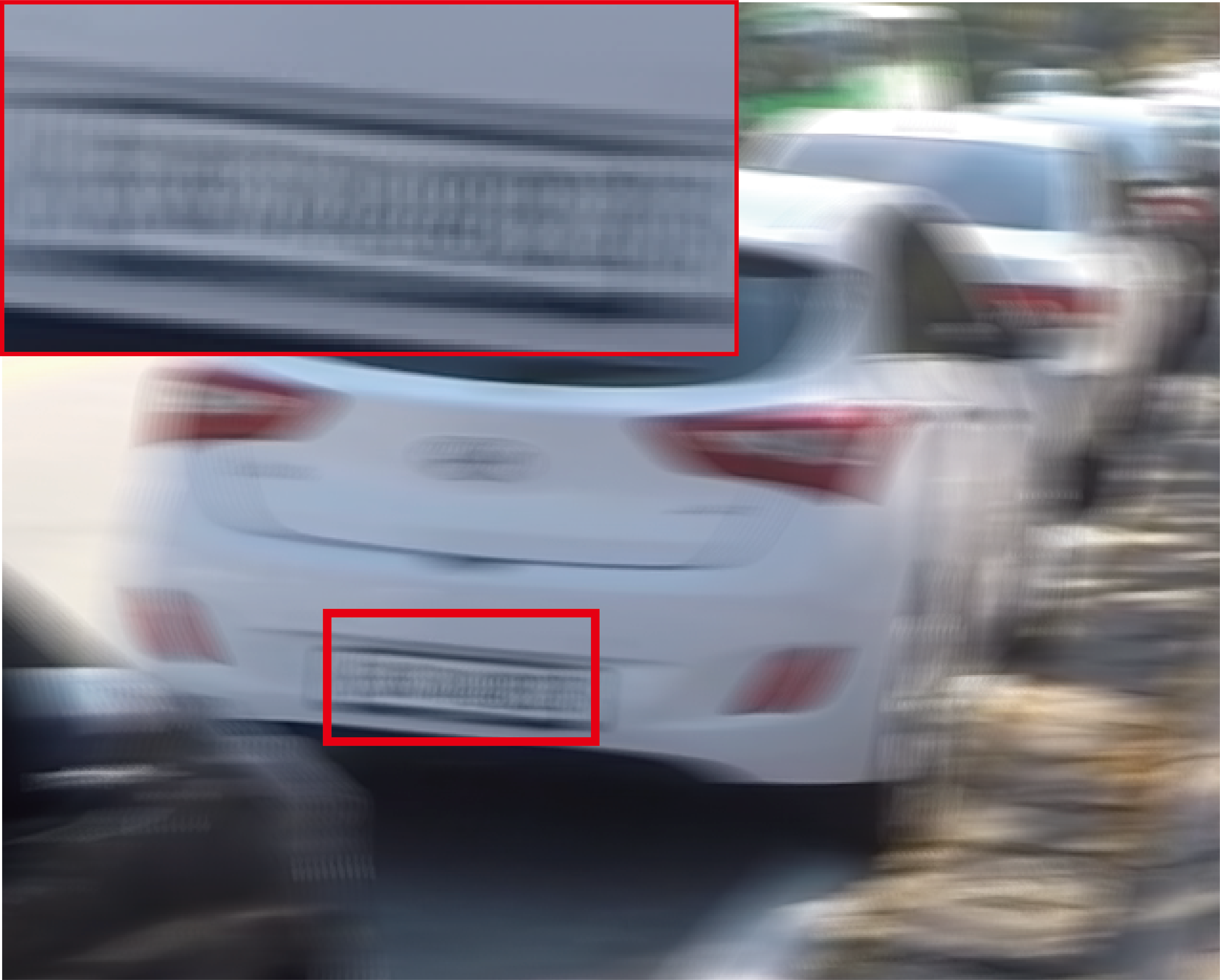}&
		\includegraphics[width=0.16\linewidth]{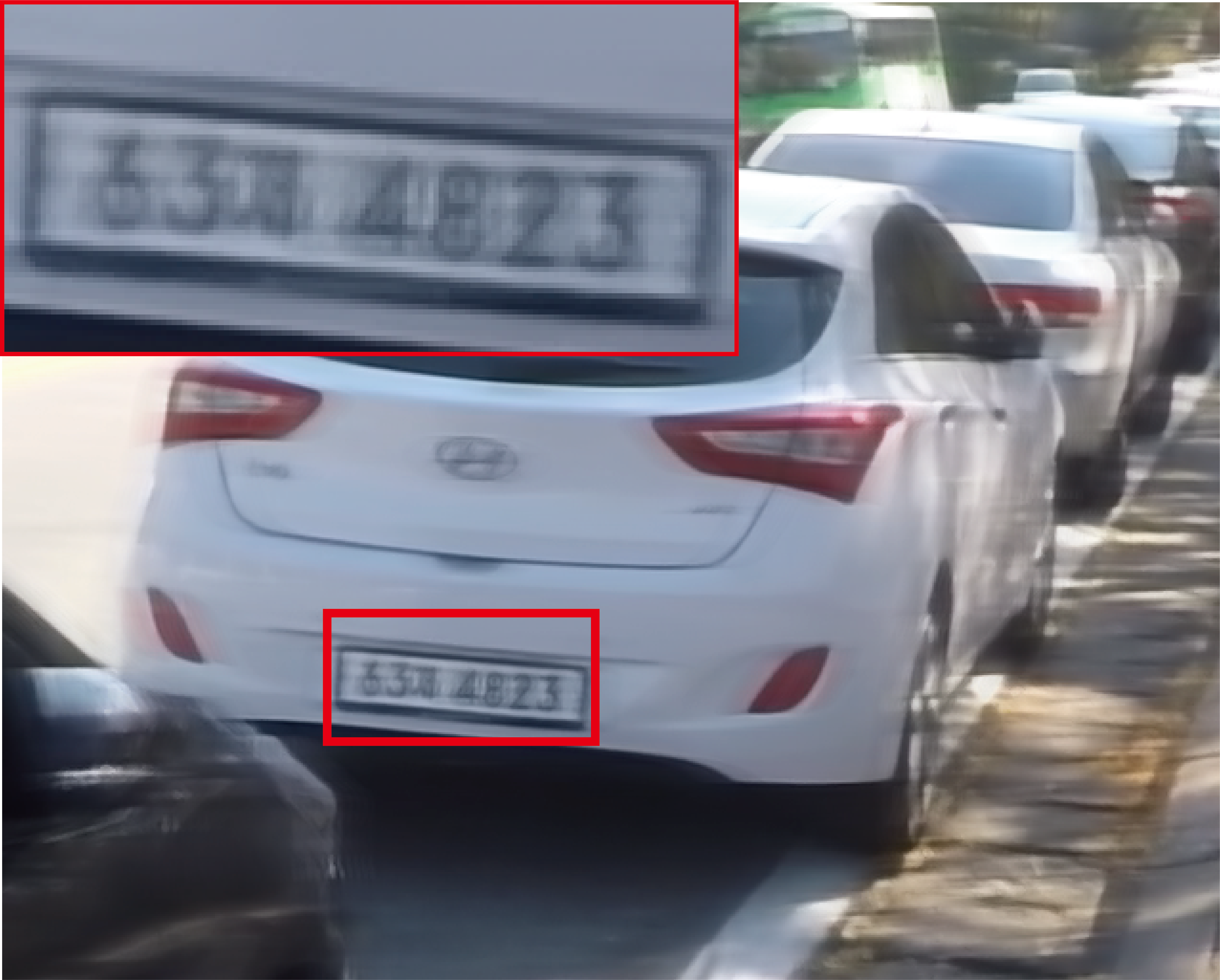}&
		\includegraphics[width=0.16\linewidth]{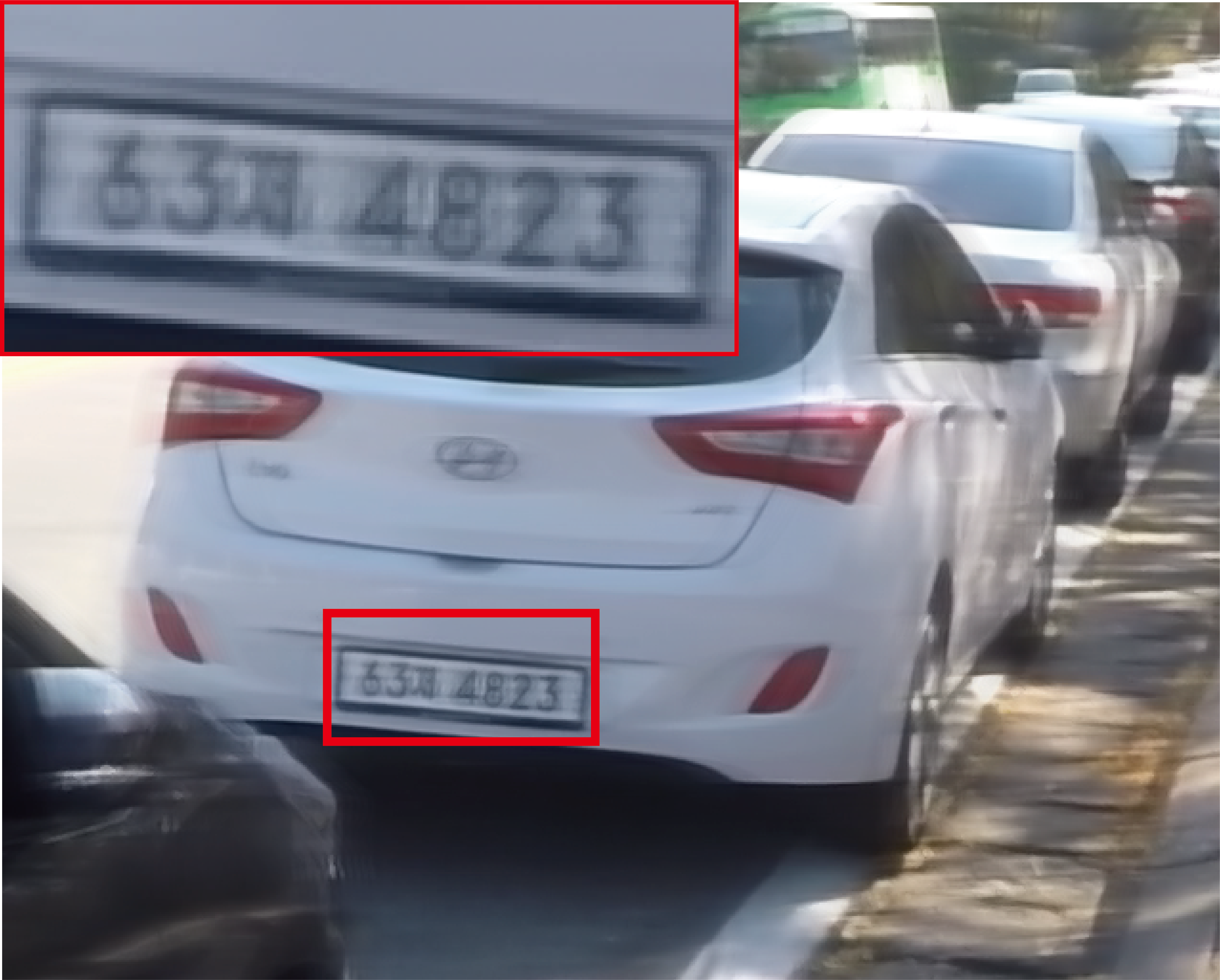}&
		\includegraphics[width=0.16\linewidth]{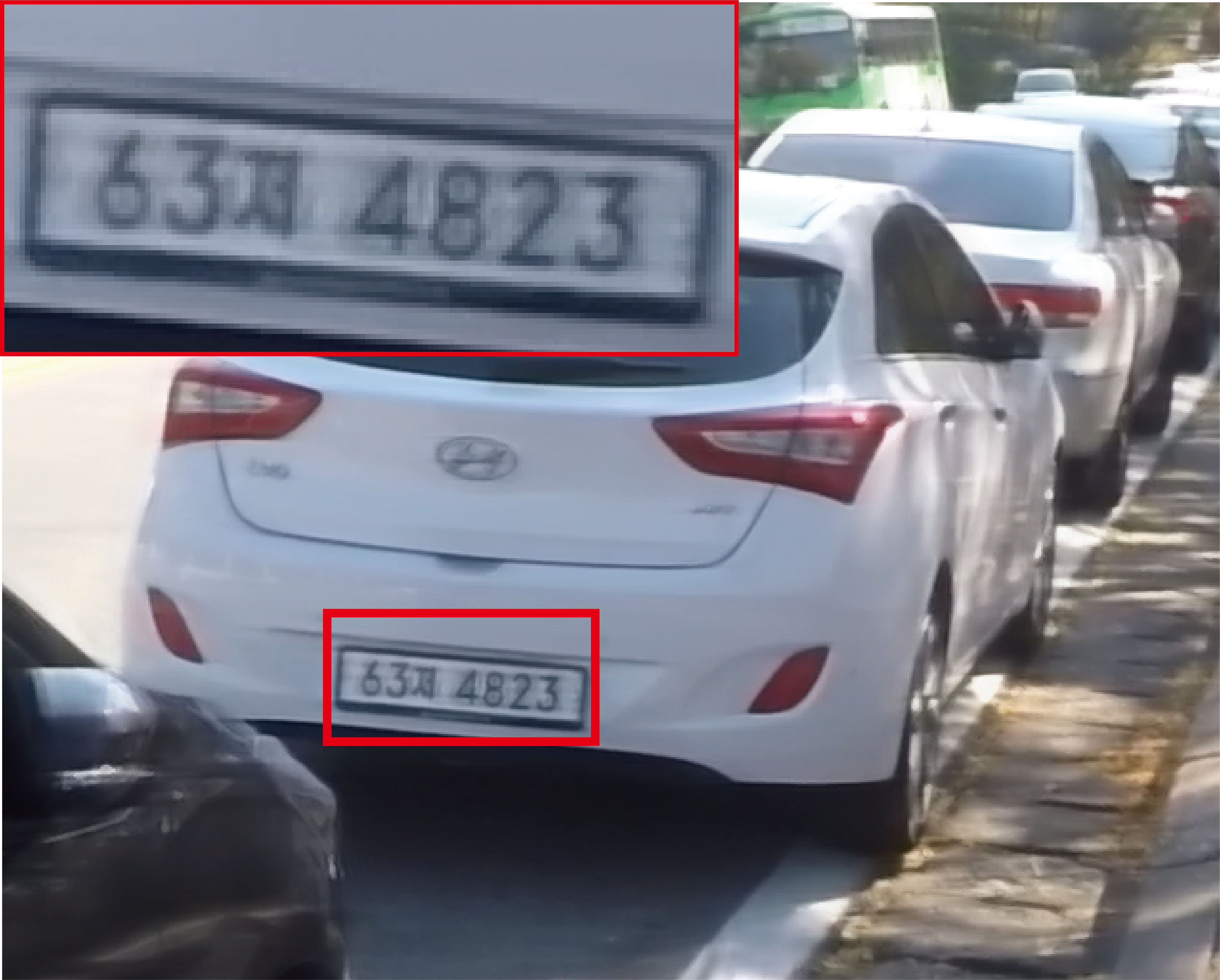}&
		\includegraphics[width=0.16\linewidth]{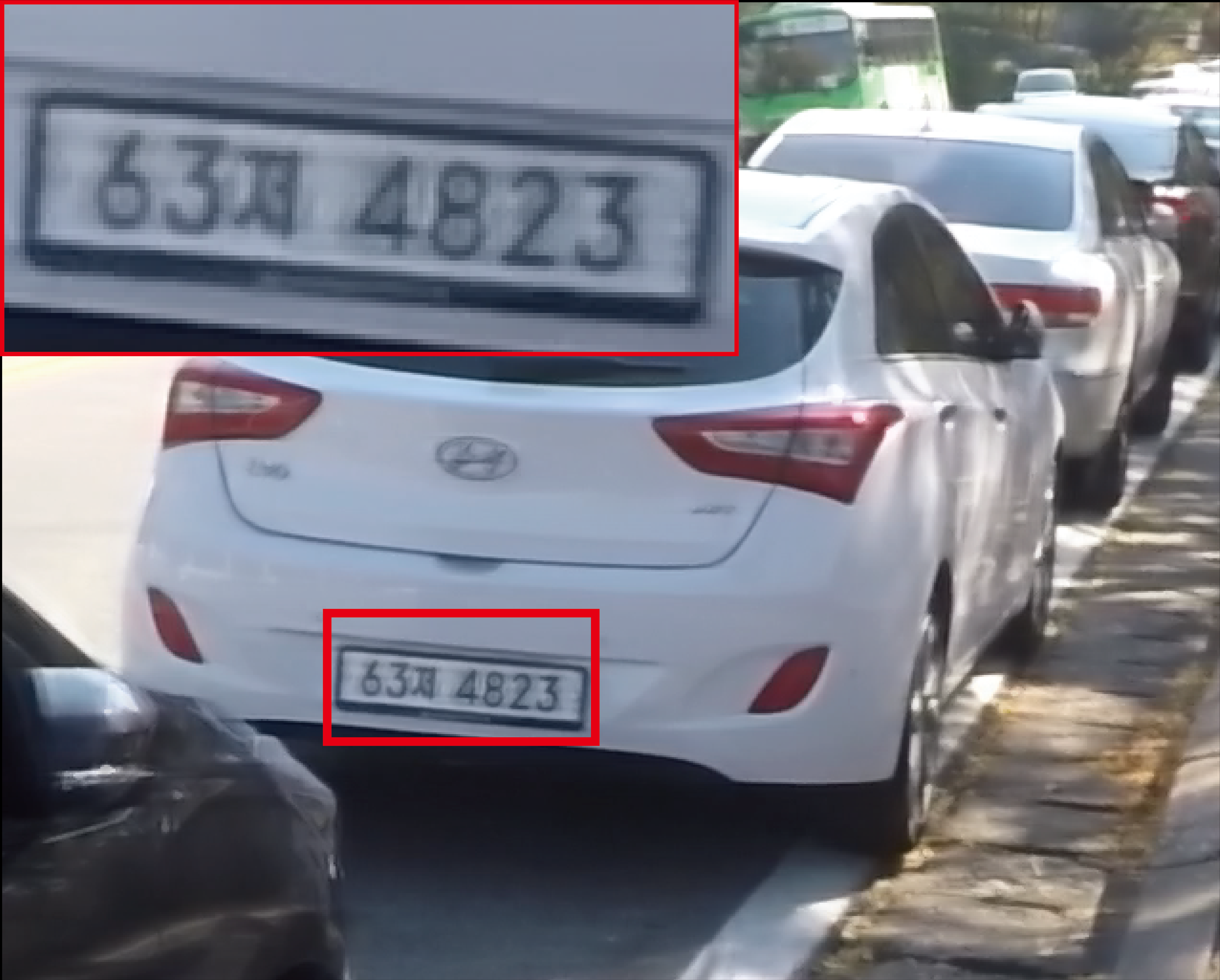}&
		\includegraphics[width=0.16\linewidth]{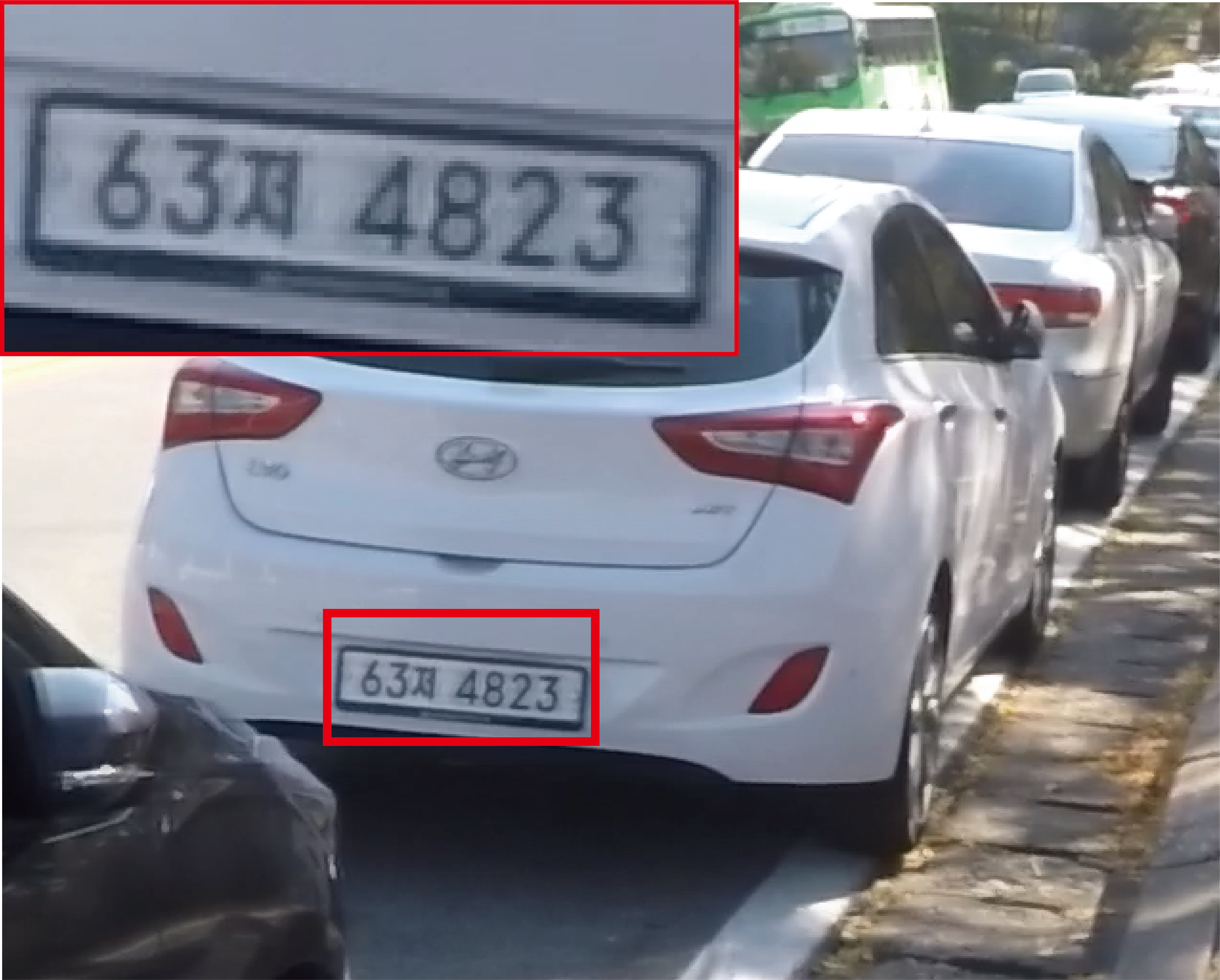}\\
		\includegraphics[width=0.16\linewidth]{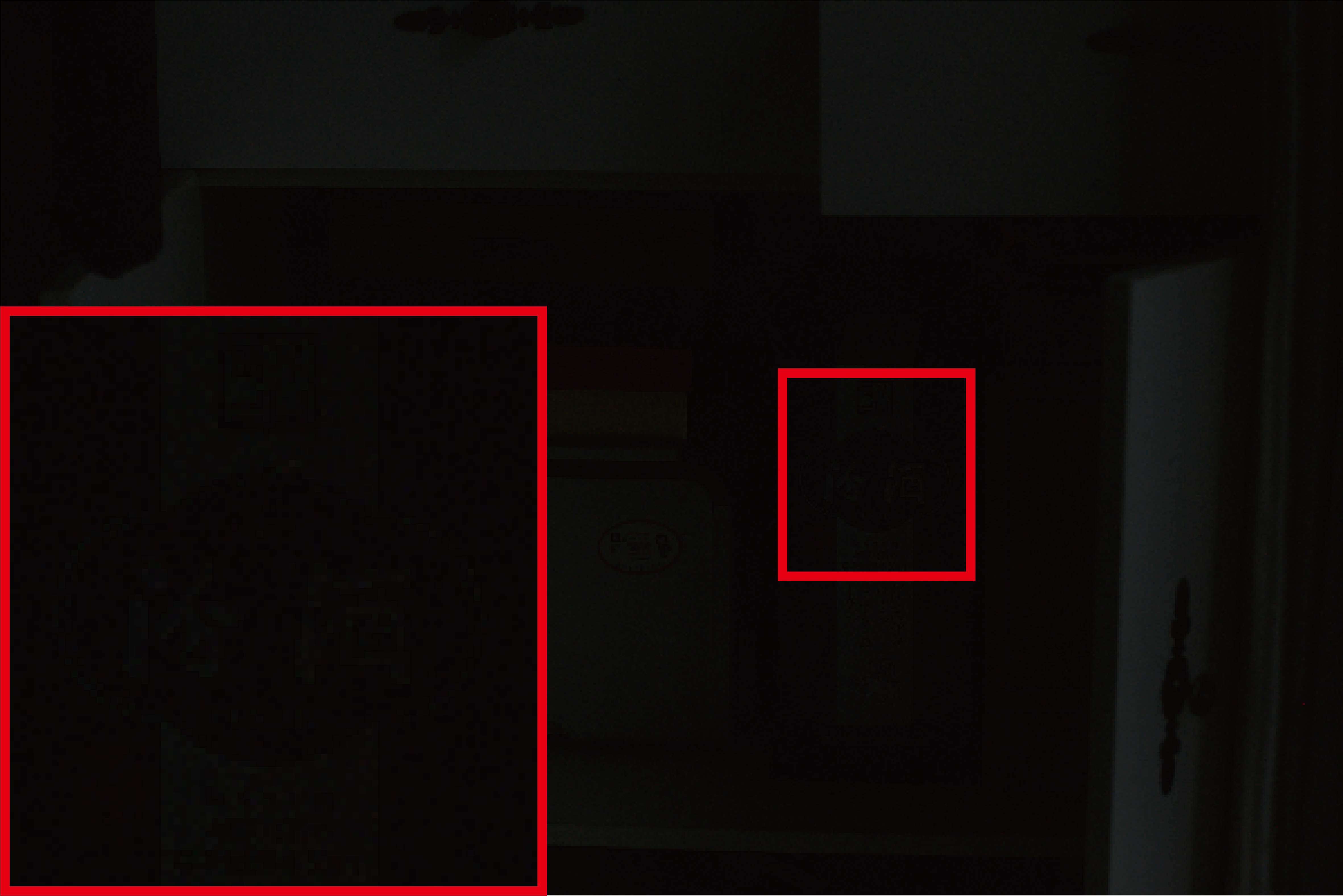}&
		\includegraphics[width=0.16\linewidth]{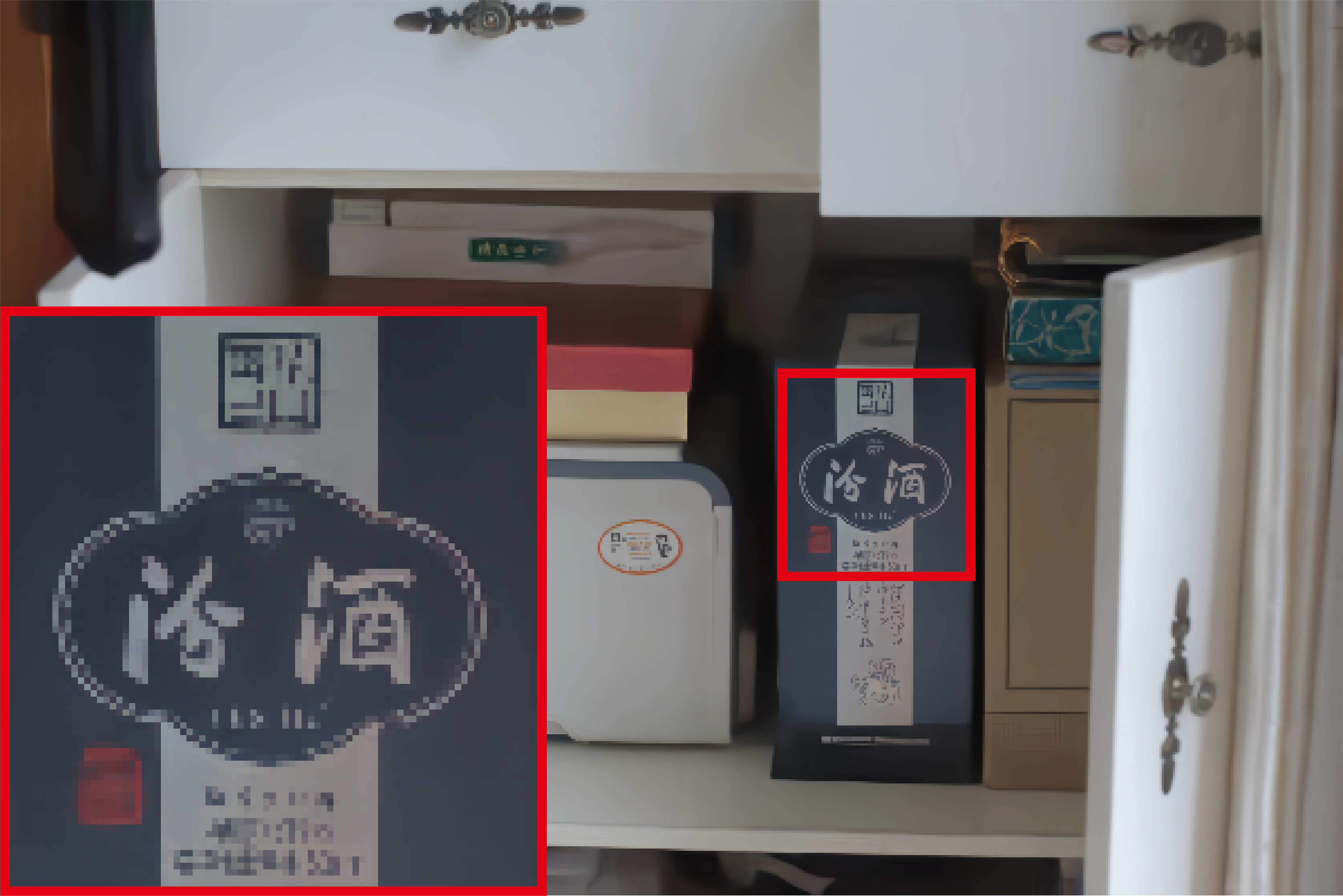}&
		\includegraphics[width=0.16\linewidth]{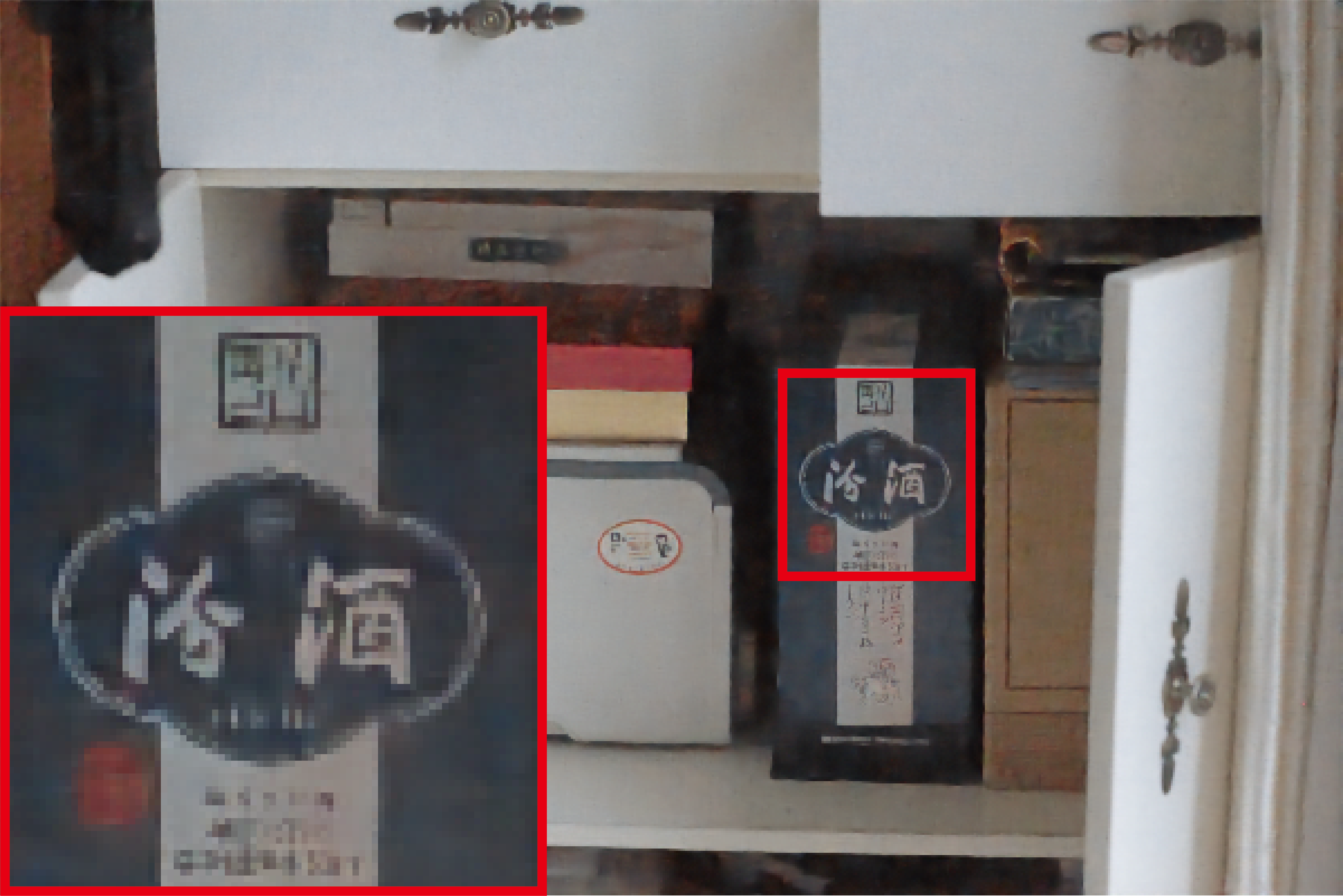}&
		\includegraphics[width=0.16\linewidth]{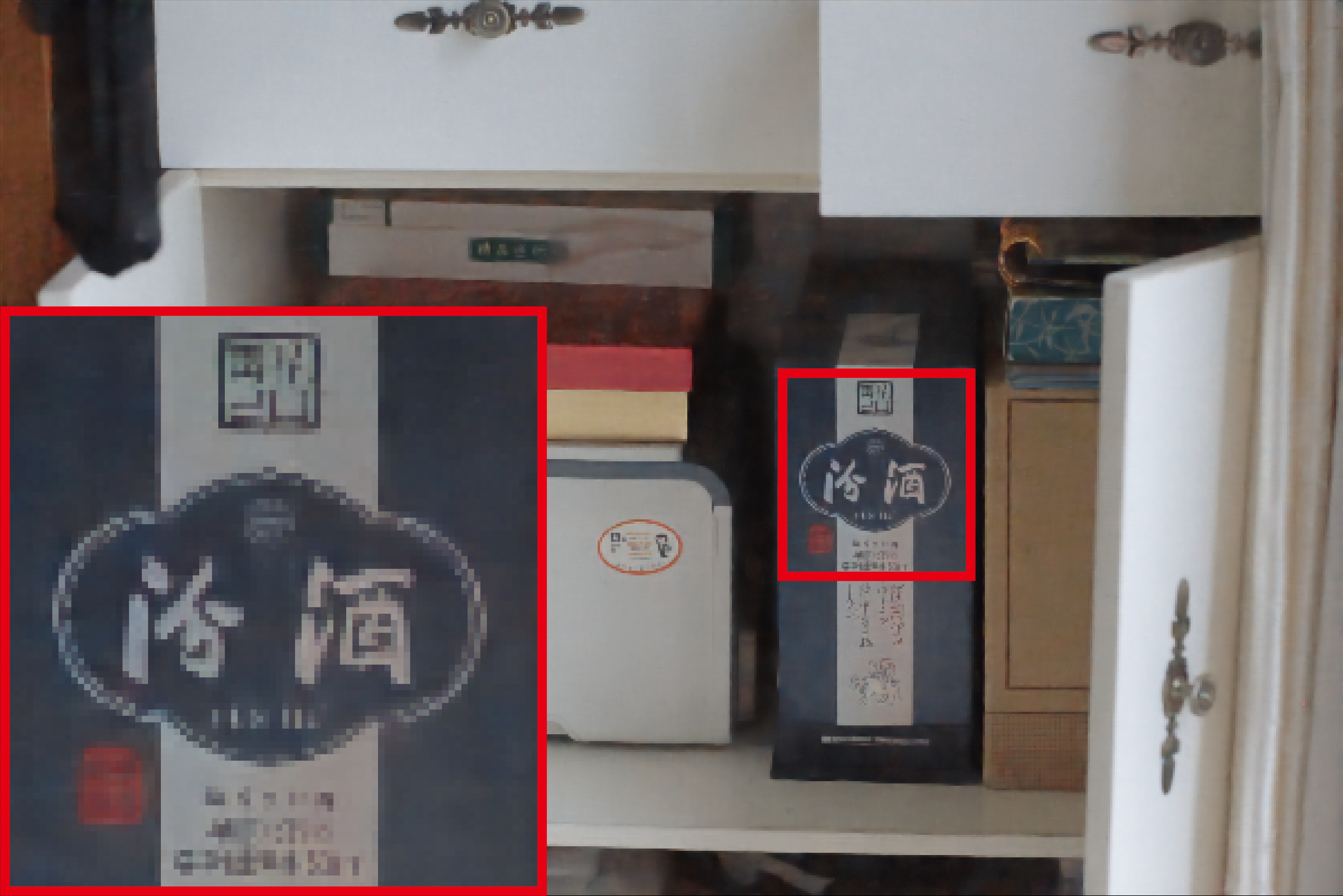}&
		\includegraphics[width=0.16\linewidth]{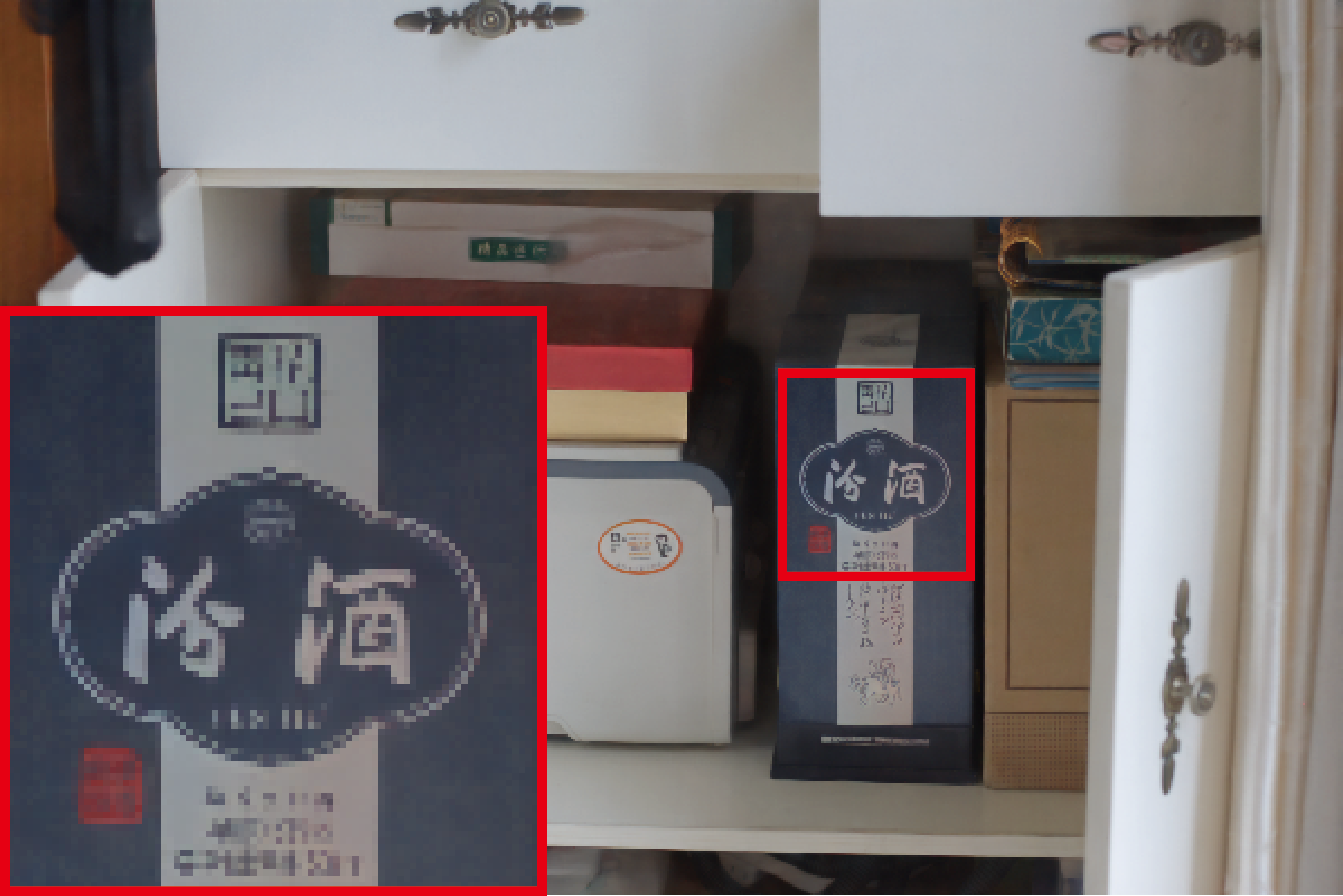}&
		\includegraphics[width=0.16\linewidth]{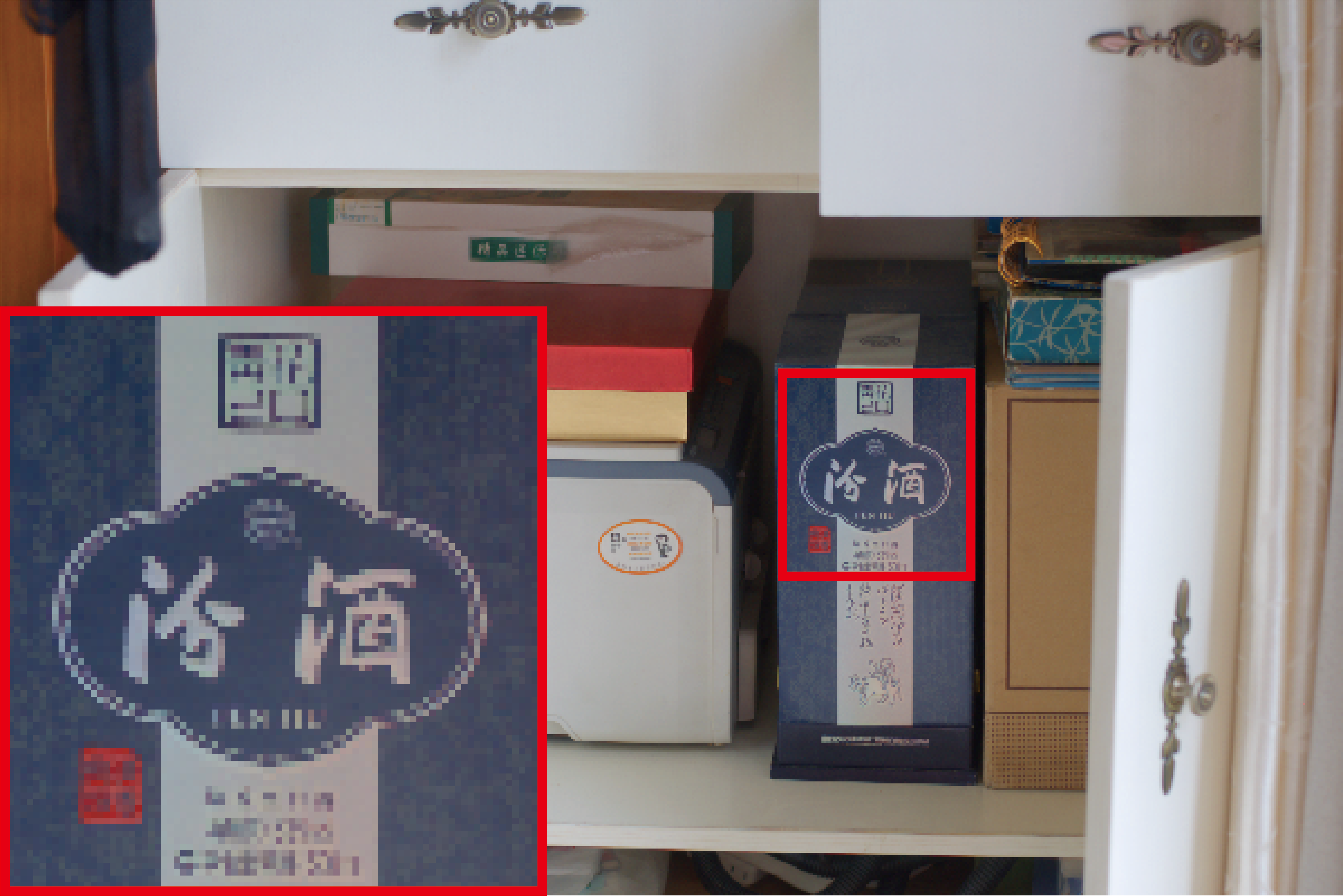}\\
		Degraded&PromptIR&DA-CLIP&InstructIR&DA-RCOT&BaryIR\\
	\end{tabular}
	
	\captionsetup{type=figure}
	\caption{Visual comparison of five-degradation all-in-one restoration results. BaryIR consistently restores multisource degraded images.}
	\label{vair5}
\end{table*}
We evaluate BaryIR for AIR on the three-degradation and five-degradation benchmarks, Following the setting of prior works \cite{zamfir2025complexity, conde2024instructir, potlapalli2023promptir}. We compare with SOTA methods, including the baseline Restormer \cite{Zamir2021Restormer} and AIR models, \textit{i.e.,} PromptIR \cite{potlapalli2023promptir}, DA-CLIP \cite{luocontrolling}, DiffUIR \cite{zheng2024selective}, InstructIR \cite{conde2024instructir}, AdaIR \cite{cui2025adair}, DA-RCOT \cite{tang2025degradation}, and MoCE-IR \cite{zamfir2025complexity}.  

\textbf{Three degradations.} The first comparison is conducted across three restoration tasks: dehazing, deraining, and denoising at noise levels $\sigma\in\{15,25,50\}$. Tab. \ref{air3} reports the quantitative results, showing that BaryIR offers consistent performance gains over other methods. Compared to PromptIR \cite{potlapalli2023promptir} which adopts the same backbone (Restormer \cite{Zamir2021Restormer}), BaryIR obtains an average PSNR gain of 0.81 dB.  BaryIR also surpasses the recent DA-RCOT \cite{tang2025degradation} with an average PSNR gain of 0.26 dB and a gain of 0.66 dB on deraining, achieving more balanced all-in-one performance.

\textbf{Five degradations.} We further verify the effectiveness of BaryIR in a five-degradation scenario:  dehazing, deraining, denoising at level $\sigma=25$, deblurring, and low-light enhancement. As shown in Tab. \ref{air5}, BaryIR excels degradation-agnostic learning-based MoCE-IR \cite{zamfir2025complexity} with an average PSNR gain of 0.52 dB. Notably, BaryIR also proceeds MoCE-IR \cite{zamfir2025complexity} with 0.66 dB PSNR gain on the dehazing task, demonstrating its balanced performance and robustness to multiple degradations.

Fig. \ref{vair5} presents visual results under the five-degradation scenario. Interestingly, BaryIR not only consistently delivers balanced and superior performance in removing degradations (\textit{e.g.,} dense haze in the distant scene as shown in row 1), but also produces results with better fine-grained structural contents (\textit{e.g.,} textures, colors). The underlying reason can be that BaryIR learns barycenters that capture common patterns of natural images, thereby effectively balancing multiple degradations and alleviate from overfitting to dominant training data.

\textbf{PromptIR as backbone.} As BaryIR offers a generic plug-in framework that can be integrated into existing restoration architectures, we explore its performance using PromptIR \cite{potlapalli2023promptir} as the backbone. As reported in Tab. \ref{air3} and \ref{air5}, BaryIR yields notable gains of 1.08 dB and 1.33 dB over the original PromptIR under three- and five-degradation settings, respectively. Notably, our method also consistently outperforms the second-best competitor, DA-RCOT \cite{tang2025degradation} by margins of 0.53 dB and 0.65 dB, respectively. These results further demonstrate its significant superiority over current SOTA methods and its robust adaptability as an efficient plug-in framework.

\subsection{Generalization to Unseen Degradations}	
\begin{figure*}[!h]
	\centering
	\setlength\tabcolsep{1pt}
	\renewcommand{\arraystretch}{0.75} 
	\begin{tabular}{ccccc}
		\includegraphics[width=0.20\linewidth]{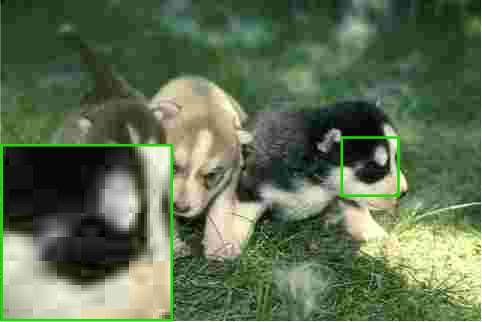}&\includegraphics[width=0.20\linewidth]{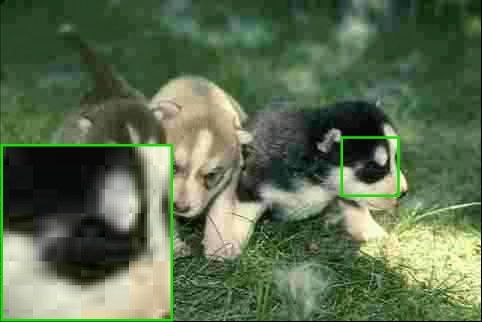}&\includegraphics[width=0.20\linewidth]{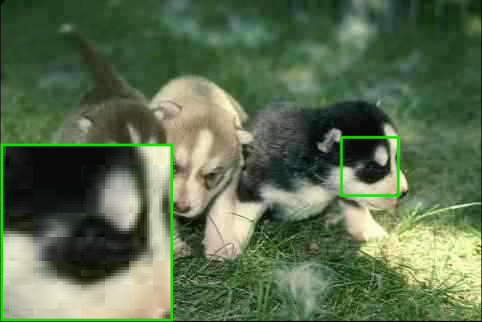}&\includegraphics[width=0.20\linewidth]{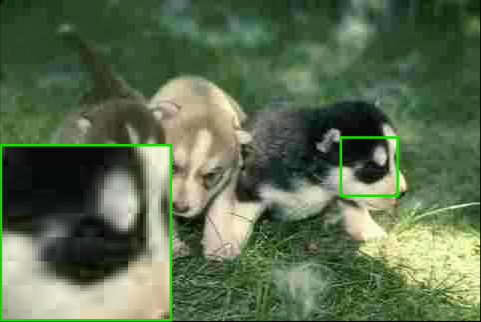}&\includegraphics[width=0.20\linewidth]{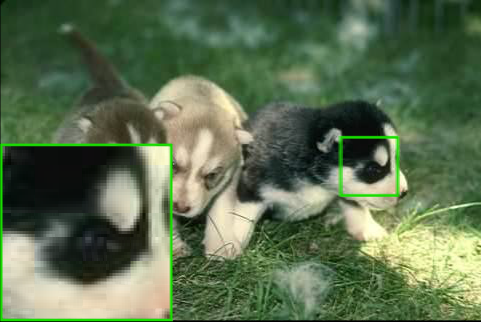}\\
		\includegraphics[width=0.20\linewidth]{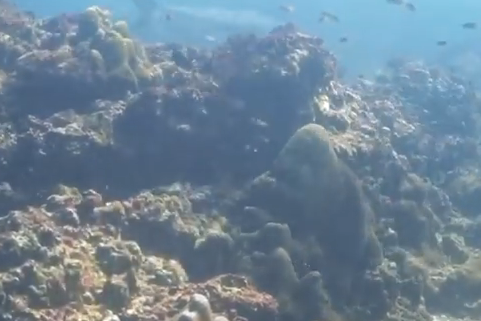}&\includegraphics[width=0.20\linewidth]{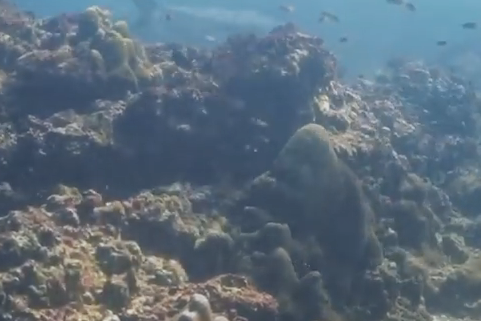}&\includegraphics[width=0.20\linewidth]{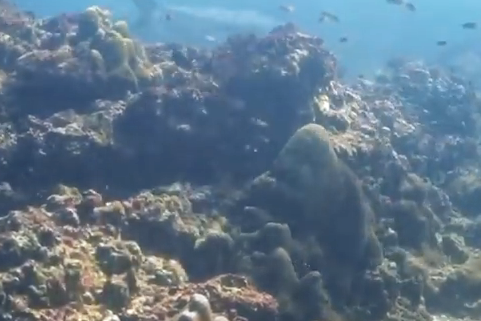}&\includegraphics[width=0.20\linewidth]{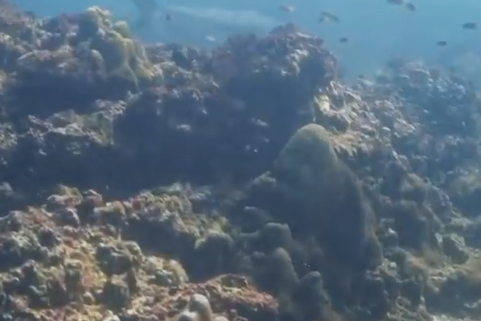}&\includegraphics[width=0.20\linewidth]{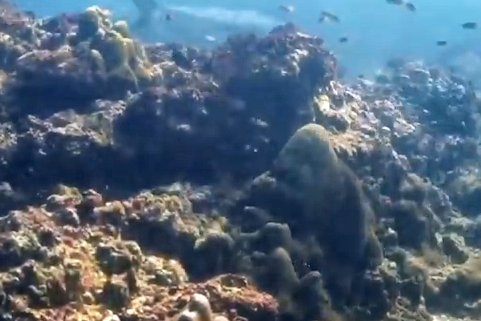}\\
		Degraded&InstructIR&DA-RCOT&MoCE-IR&BaryIR \\
	\end{tabular}
	\caption{Visual comparison of generalization results on \textbf{unseen degradation types},  \textit{i.e.,} JPEG artifact correction (row 1) on BSD500 (QF=10) and underwater image enhancement on UIEB (row 2). BaryIR restores high-quality images with more faithful structural contents such as textures and colors.}
	\label{ood-eva}
\end{figure*}

\begin{figure*}[!h]
	\centering
	\includegraphics[scale=0.55]{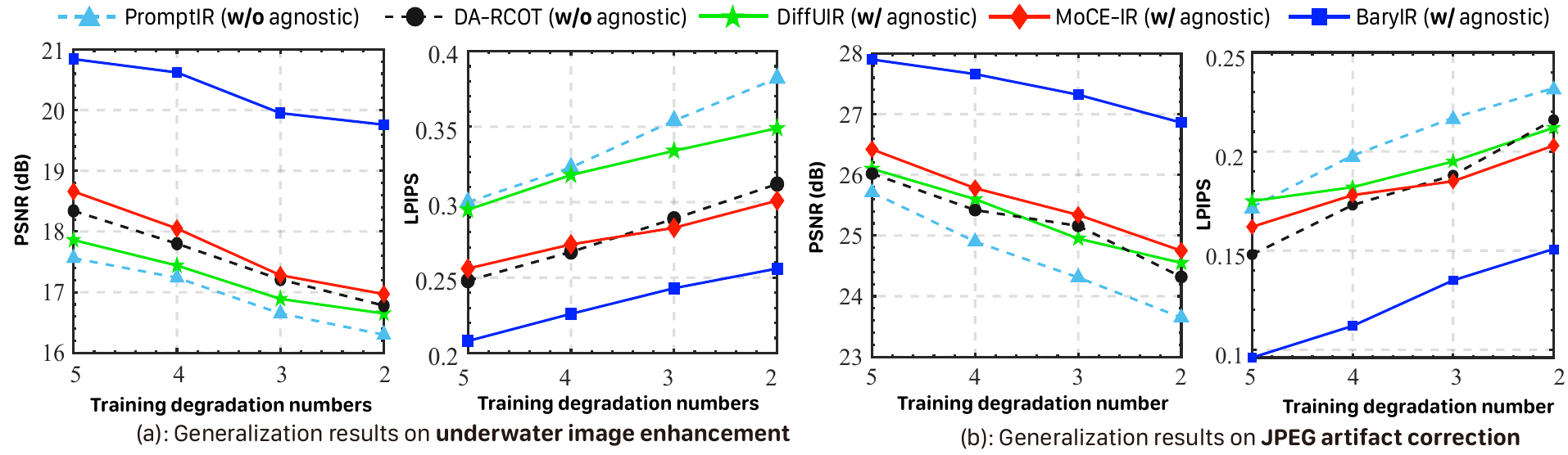}
	\caption{Numerical comparison of robustness of generalization to training degradation numbers. As the degradation number decreases, BaryIR remain superior generalization to the unseen degradation types while achieving the best quantitative performance in comparison with other methods.}
	\label{Rob}
\end{figure*}

To validate the generalization advantages of BaryIR, we evaluate its performance on both out-of-distribution (OOD) degradation types (\textit{i.e.,} JPEG artifact correction and underwater image enhancement) and degradation levels (\textit{i.e.,} unseen rain/noise levels). Particularly, the compared methods DiffUIR \cite{zheng2024selective} and MoCE-IR \cite{zamfir2025complexity} also aim to capture the degradation-agnostic semantics for improved generalization.

\noindent{\textbf{Unseen degradation types.}}  We evaluate the 5-degradation models on OOD degradation types,  \textit{i.e.,} JPEG artifact correction on BSD500 \cite{arbelaez2010contour} with quality factor (QF) = 10 and underwater image enhancement on UIEB \cite{li2019underwater}. 

\begin{table}[!h]
	\centering
	\caption{ Generalization performance on unseen degradation types, \textit{i.e.,} JPEG artifact correction on BSD500 (QF=10) and underwater image enhancement on UIEB. The metrics are reported as PSNR($\uparrow$)/SSIM($\uparrow$)/LPIPS($\downarrow$)/FID($\downarrow$)}
	\label{ood-eva-tab}
	\resizebox{1.0\linewidth}{!}{
		\begin{tabular}{lcc}
			\toprule
			Method & BSD500 & UIEB\\
			\midrule 
			Restormer \cite{Zamir2021Restormer} & 25.60/0.742/0.177/71.24&17.34/0.770/0.300/47.12\\
			PromptIR \cite{potlapalli2023promptir} & 25.71/0.748/0.172/65.33&17.56/0.778/0.285/40.26\\
			DiffUIR \cite{zheng2024selective} &26.10/0.762/0.175/75.22&17.86/0.784/0.295/36.23\\
			InstructIR \cite{conde2024instructir} & 25.54/0.746/0.185/76.68&17.51/0.780/0.288/43.26\\
			DA-RCOT \cite{tang2025degradation}&26.02/0.765/\underline{0.148}/\underline{45.71}&{18.34}/\underline{0.802}/\underline{0.248}/\underline{30.53}\\
			MoCE-IR \cite{zamfir2025complexity}&\underline{26.42}/\underline{0.768}/{0.162}/58.96&\underline{18.66}/{0.800}/0.256/33.25\\\midrule
			BaryIR &\textbf{27.94}/\textbf{0.835}/\textbf{0.096}/\textbf{30.65}&\textbf{20.84}/\textbf{0.825}/\textbf{0.208}/\textbf{20.65}\\
			\toprule
	\end{tabular}}
\end{table}

The quantitative results are reported in Tab.~\ref{ood-eva-tab}. As observed, BaryIR consistently outperforms existing methods across all metrics, achieving substantial gains in all metrics. Compared with recent degradation-agnostic methods such as DiffUIR and MoCE-IR, BaryIR consistently yields superior generalization results. This demonstrates that the learned barycenter and WB embeddings effectively capture degradation-agnostic invariance and degradation-specific information beyond the training domain and mitigate overfitting to specific degradation types. From the visual results in Fig. \ref{ood-eva}, we can observe thar BaryIR yields high-quality images with faithful structural patterns  (\textit{e.g.,} colors and textures), which indicates that BaryIR captures the fine-grained geometric structures of data.

\begin{table}[!h]
	\centering
	\caption{The deraining results on unseen rain levels using the five-degradation models. The metrics are reported as PSNR($\uparrow$)/SSIM($\uparrow$)/LPIPS($\downarrow$)/FID($\downarrow$).}
	\label{unseen1}
	\resizebox{1.0\linewidth}{!}{
		\begin{tabular}{lcc}
			\toprule
			Method &Rain100L&Rain100H\\
			\midrule
			Restormer \cite{Zamir2021Restormer}  &28.76/0.901/0.140/63.21 
			&14.50/0.464/0.484/250.2\\
			IR-SDE \cite{luo2023image} &28.49/0.897/0.123/55.21&13.55/0.422/0.465/234.5\\
			\midrule
			PromptIR \cite{potlapalli2023promptir} &31.82/0.931/0.078/38.41&14.28/0.444/0.472/242.7\\
			DA-CLIP \cite{luocontrolling}&32.87/0.944/0.066/35.12&14.40/0.435/0.438/228.6\\
			DiffUIR \cite{zheng2024selective}&33.20/0.942/0.036/34.64&14.78/0.487/0.442/235.5\\
			InstructIR \cite{conde2024instructir}&{33.65}/{0.951}/{0.030}/{28.24}&14.67/0.468/0.460/238.3 \\
			DA-RCOT \cite{tang2025degradation} &\underline{35.88}/\underline{0.973}/\underline{0.019}/\underline{19.55}&15.88/0.523/\underline{0.378}/\underline{167.2}\\
			MoCE-IR \cite{zamfir2025complexity}
			&34.87/0.966/0.027/28.42&\underline{16.02}/\underline{0.528}/0.402/189.4\\
			\midrule
			BaryIR &\textbf{36.69}/\textbf{0.975}/\textbf{0.018}/\textbf{10.28}&\textbf{17.30}/\textbf{0.551}/\textbf{0.342}/\textbf{128.9}\\
			\bottomrule
	\end{tabular}}
	\vspace{0.2cm}
	\centering
	\caption{The denoising results on unseen noise levels using the three-degradation models. The metrics are reported as PSNR($\uparrow$)/SSIM($\uparrow$)/LPIPS($\downarrow$)/FID($\downarrow$).}
	\label{unseen2}
	\resizebox{1.0\linewidth}{!}{
		\begin{tabular}{lcc}
			\toprule
			Method & $\sigma=60$ & $\sigma=75$\\
			\midrule
			Restormer \cite{Zamir2021Restormer}   &18.30/0.465/0.273/165.2&13.76/0.358/0.476/205.1\\
			IR-SDE \cite{luo2023image} &17.55/0.410/0.245/142.2&13.35/0.332/0.456/185.2                                              \\
			\midrule
			PromptIR \cite{potlapalli2023promptir} & 21.94/0.584/0.227/122.4&18.55/0.402/0.401/167.6\\
			DA-CLIP \cite{luocontrolling}&19.68/0.465/0.221/142.1&16.92/0.382/0.402/166.3\\
			DiffUIR \cite{zheng2024selective}&22.25/0.577/0.198/113.4&18.89/0.405/0.388/160.2\\
			InstructIR \cite{conde2024instructir}& {24.56}/{0.626}/{0.178}/92.33&{19.55}/{0.455}/{0.374}/{155.8}\\
			DA-RCOT \cite{tang2025degradation} &\underline{25.15}/\underline{0.675}/\underline{0.166}/\underline{74.23}&\underline{20.65}/\underline{0.474}/\underline{0.357}/\underline{141.2}\\
			MoCE-IR \cite{zamfir2025complexity}&24.89/0.652/0.172/89.65&20.12/0.465/0.382/164.2\\
			\midrule
			BaryIR &\textbf{26.83}/\textbf{0.749}/\textbf{0.134}/\textbf{74.63}&\textbf{22.85}/\textbf{0.507}/\textbf{0.324}/\textbf{116.6}\\
			\bottomrule
	\end{tabular}}
\end{table}

\noindent\textbf{Unseen degradation levels.} We also evaluate the performance of BaryIR on unseen degradation levels. Specifically, we design two evaluations. In the first setting, BaryIR is trained with five degradations as described in \S\ref{5.1}, where Rain100L (light rain) and Rain100H (heavy rain) are alternately used for training and testing—\textit{i.e.,} the model is trained on one set and evaluated on the other. In the second setting, we investigate its generalization to severe noise levels, $\sigma=60$ and $\sigma=75$, while the model is trained under the three-degradation setup in \S\ref{5.1}, which covers noise levels $\sigma\in\{15,25,50\}$.

From Tab. \ref{unseen1} and Tab. \ref{unseen2}, we can observe that BaryIR consistently outperforms both the baseline Restormer and recent degradation-agnostic approaches such as DiffUIR and MoCE-IR under unseen degradation levels. For instance, on Rain100L, BaryIR achieves a remarkable gain of 1.82 dB in PSNR over MoCE-IR, while for severe noise level $\sigma=75$, it surpasses the best competing method by 2.20 dB. These results justifies the strong generalization capability of BaryIR beyond the training degradation levels, demonstrating the effectiveness of barycenter modeling in capturing invariant contents and mitigating overfitting to specific degradations.

\quad\\
\begin{table}[!h]
	\caption{Generalization to unseen real-world O-HAZE \cite{ancuti2018haze}  and SPANet \cite{Wang_2019_CVPR} datasets with the five-degradation models.  The metrics are reported as PSNR($\uparrow$)/SSIM($\uparrow$)/LPIPS($\downarrow$)/FID($\downarrow$). }
	\label{unseen-real-tab}
	\resizebox{1.0\linewidth}{!}{
		\begin{tabular}{lccc}
			\toprule
			Method &  O-HAZE &  SPANet & LOL-v2-real\\
			\midrule
			Restormer \cite{Zamir2021Restormer}   &18.02/0.724/0.345/275.8&34.38/0.917/0.032/43.29&27.12/0.877/0.112/85.45\\
			IR-SDE \cite{luo2023image} &17.85/0.716/0.338/256.3&35.02/0.922/0.029/38.87&23.12/0.801/0.144/122.7\\
			\midrule
			PromptIR \cite{potlapalli2023promptir} & 18.38/0.730/0.336/260.1&35.34/0.938/0.026/33.12&27.65/0.870/0.107/80.1\\
			DA-CLIP \cite{luocontrolling}&18.22/0.725/0.323/242.5&35.65/0.942/0.026/26.96&26.46/0.856/0.104/76.45\\
			DiffUIR \cite{zheng2024selective}&18.75/0.731/0.315/225.4&35.95/0.940/{0.022}/{22.25}&26.12/0.861/0.098/78.24\\
			InstructIR \cite{conde2024instructir}&18.85/0.738/0.308/236.5&{36.42}/{0.946}/0.028/30.54&28.02/{0.901}/0.110/81.85\\
			DA-RCOT \cite{tang2025degradation} &{20.45}/{0.768}/\underline{0.277}/\underline{189.6}&{37.24}/{0.960}/{0.018}/\underline{20.65}&\underline{28.33}/\underline{0.907}/\underline{0.084}/\underline{64.97}\\
			MoCE-IR \cite{zamfir2025complexity}&\underline{20.89}/\underline{0.772}/0.296/200.4&\underline{37.56}/\underline{0.972}/\underline{0.016}/24.14&\underline{28.20}/{0.904}/0.104/84.21\\
			\midrule
			BaryIR &\textbf{22.98}/\textbf{0.794}/\textbf{0.252}/\textbf{169.6}&\textbf{39.24}/\textbf{0.973}/\textbf{0.012}/\textbf{16.24}&\textbf{29.14}/\textbf{0.927}/\textbf{0.065}/\textbf{40.29}\\
			\bottomrule
	\end{tabular}}
\end{table}
\noindent\textbf{Generalization to real-world scenarios.} We compare BaryIR with state-of-the-art methods on unseen real-world haze O-HAZE \cite{ancuti2018haze}, rain SPANet \cite{Wang_2019_CVPR}, and low light LOL-v2-real \cite{yang2021sparse} using the five-degradation models, in which the training dataset mainly covers synthetic degradations.

\begin{figure*}[!htbp]
	\setlength\tabcolsep{1pt}
	\renewcommand{\arraystretch}{0.5} 
	\centering
	\begin{tabular}{cccccc}
		\includegraphics[width=0.16\linewidth]{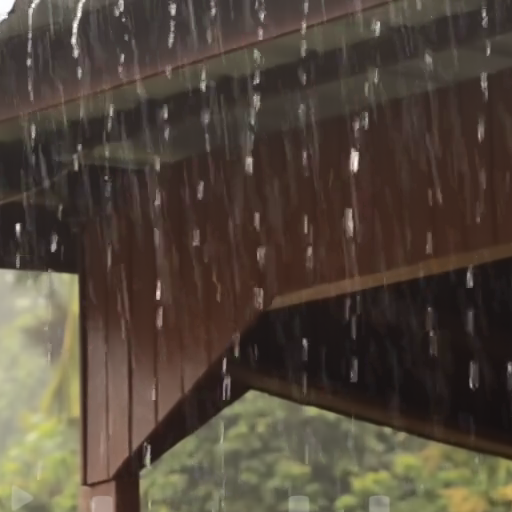}&
		\includegraphics[width=0.16\linewidth]{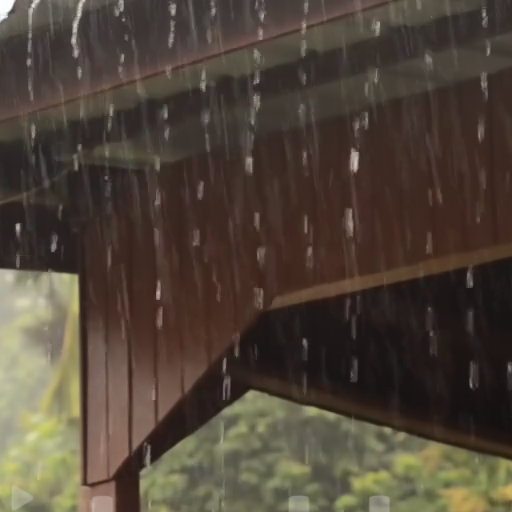}&
		\includegraphics[width=0.16\linewidth]{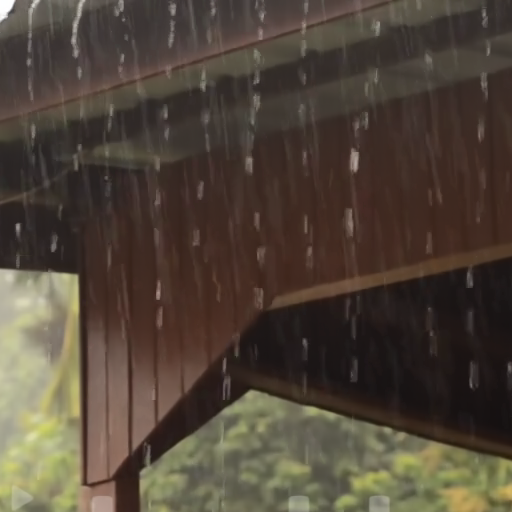}&
		\includegraphics[width=0.16\linewidth]{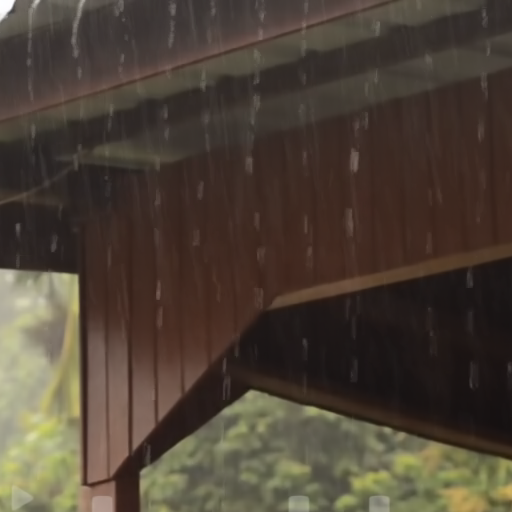}&
		\includegraphics[width=0.16\linewidth]{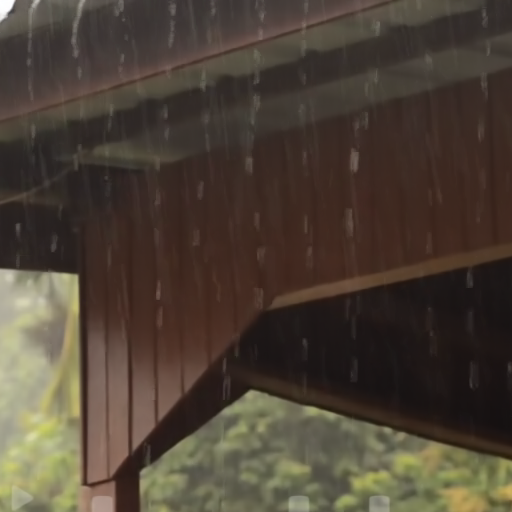}&
		\includegraphics[width=0.16\linewidth]{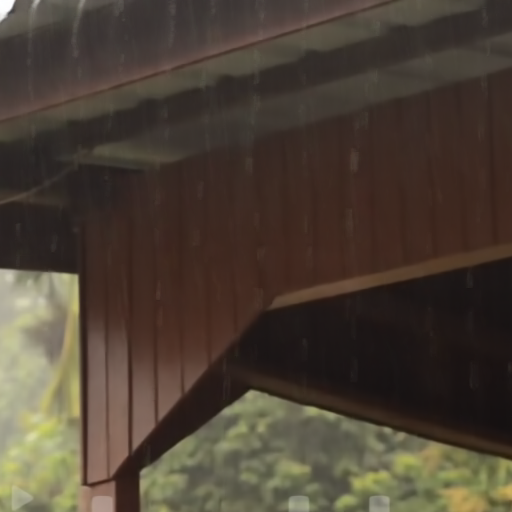}\\
		\includegraphics[width=0.16\linewidth]{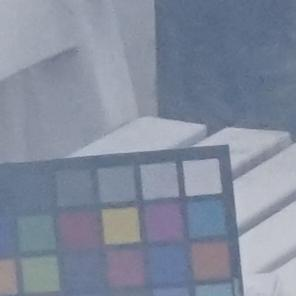}&
		\includegraphics[width=0.16\linewidth]{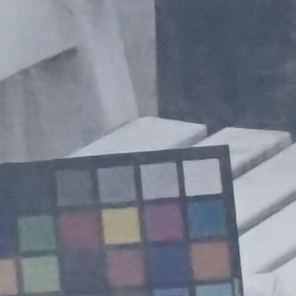}&
		\includegraphics[width=0.16\linewidth]{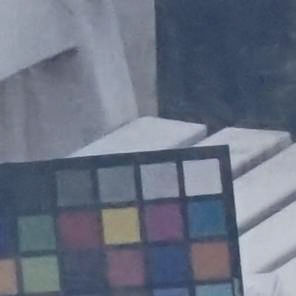}&
		\includegraphics[width=0.16\linewidth]{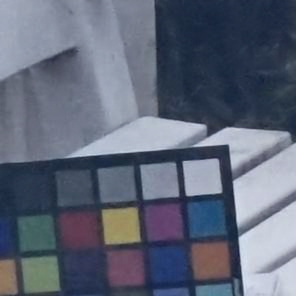}&
		\includegraphics[width=0.16\linewidth]{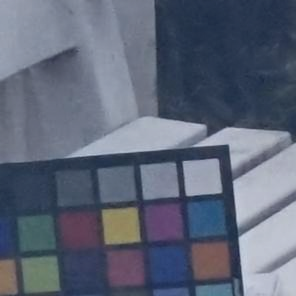}&
		\includegraphics[width=0.16\linewidth]{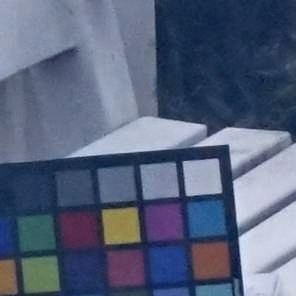}\\
		Degraded&PromptIR&InstructIR&DA-RCOT&MoCE-IR&BaryIR\\
	\end{tabular}
	\caption{Visual examples of generalization evaluation with five-degradation models on \textbf{unseen real-world} O-HAZE \cite{ancuti2018haze} and SPANet \cite{Wang_2019_CVPR}.}
	\label{unseen-real-fig}
	\vspace{0.34cm}
	\setlength\tabcolsep{1pt}
	\renewcommand{\arraystretch}{0.5} 
	\centering
	\begin{tabular}{cccccc}
		\includegraphics[width=0.16\linewidth]{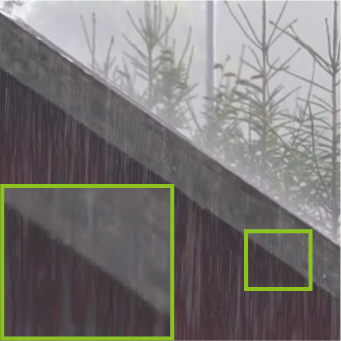}&
		\includegraphics[width=0.16\linewidth]{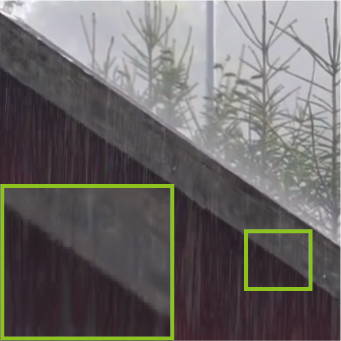}&
		\includegraphics[width=0.16\linewidth]{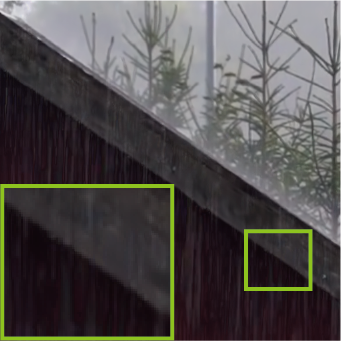}&
		\includegraphics[width=0.16\linewidth]{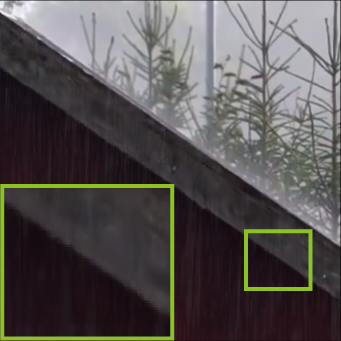}&
		\includegraphics[width=0.16\linewidth]{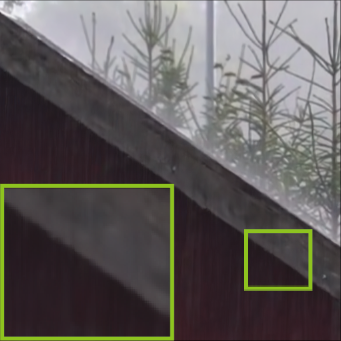}&
		\includegraphics[width=0.16\linewidth]{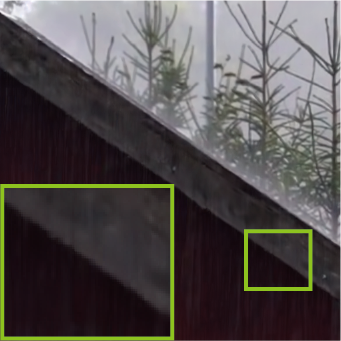}\\
		\includegraphics[width=0.16\linewidth]{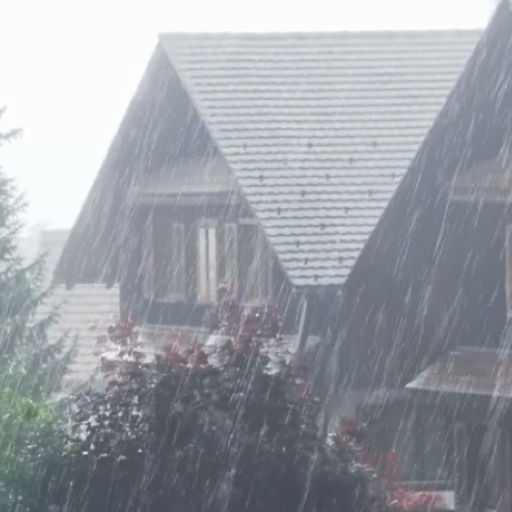}&
		\includegraphics[width=0.16\linewidth]{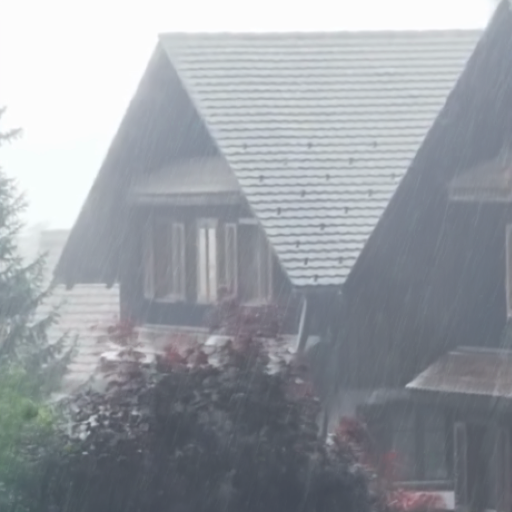}&
		\includegraphics[width=0.16\linewidth]{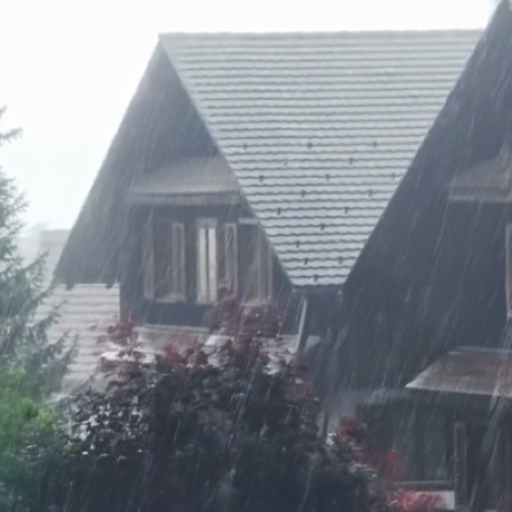}&
		\includegraphics[width=0.16\linewidth]{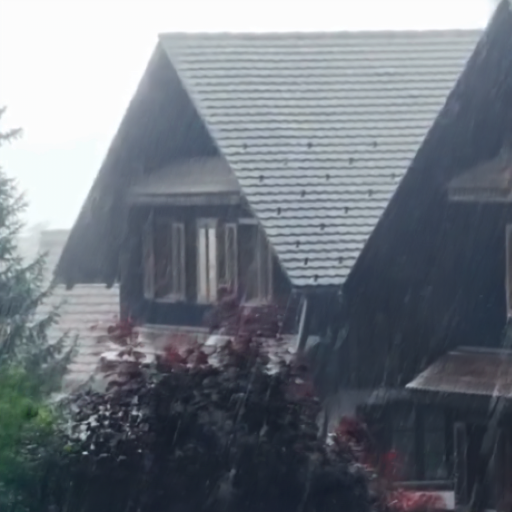}&
		\includegraphics[width=0.16\linewidth]{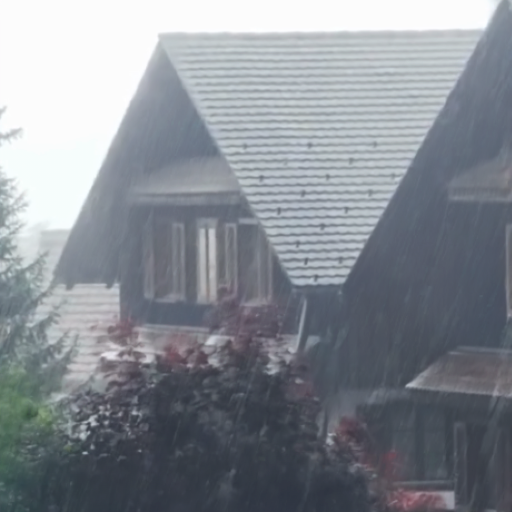}&
		\includegraphics[width=0.16\linewidth]{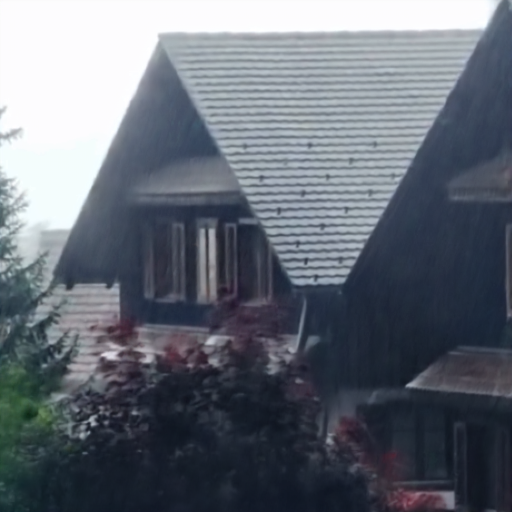}\\
		\includegraphics[width=0.16\linewidth]{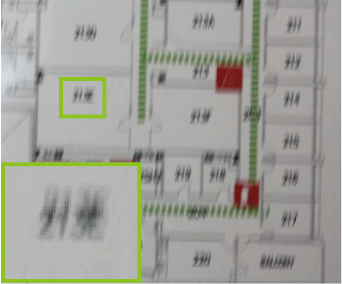}&
		\includegraphics[width=0.16\linewidth]{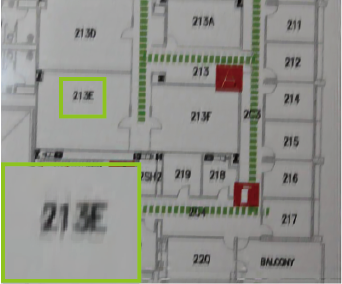}&
		\includegraphics[width=0.16\linewidth]{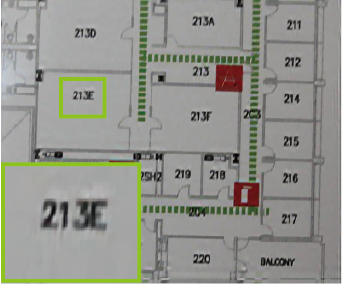}&
		\includegraphics[width=0.16\linewidth]{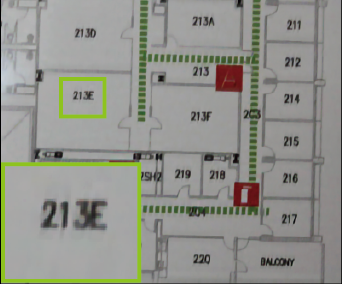}&
		\includegraphics[width=0.16\linewidth]{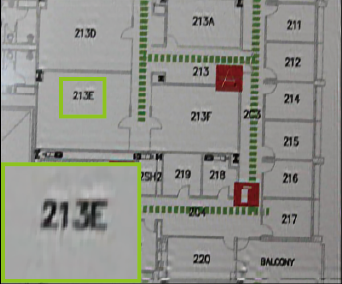}&
		\includegraphics[width=0.16\linewidth]{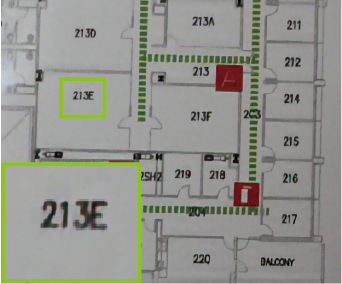}\\
		\includegraphics[width=0.16\linewidth]{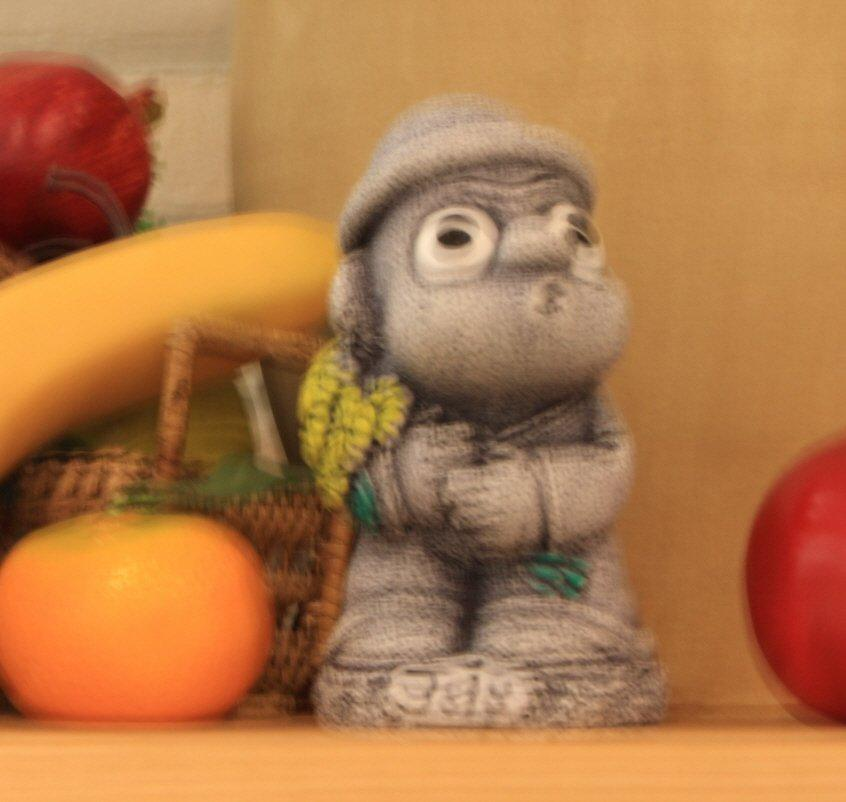}&
		\includegraphics[width=0.16\linewidth]{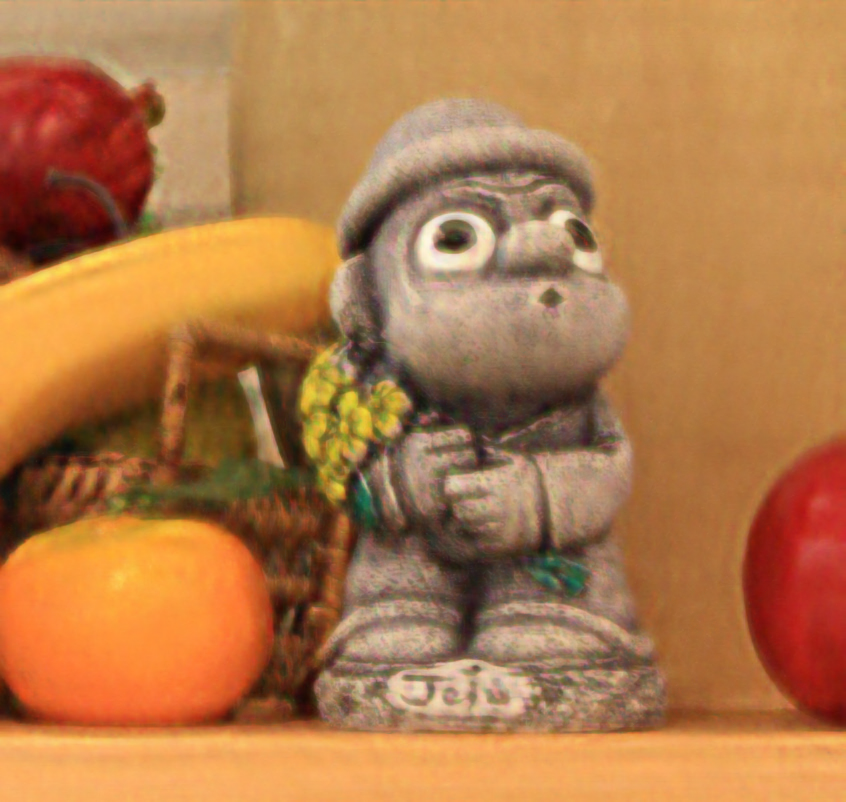}&
		\includegraphics[width=0.16\linewidth]{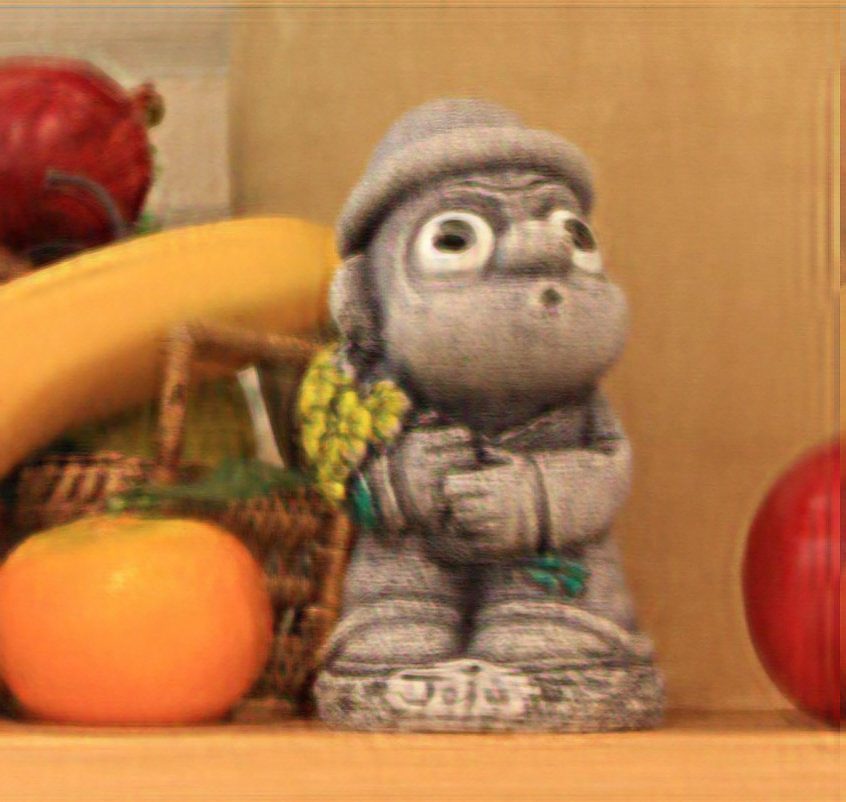}&
		\includegraphics[width=0.16\linewidth]{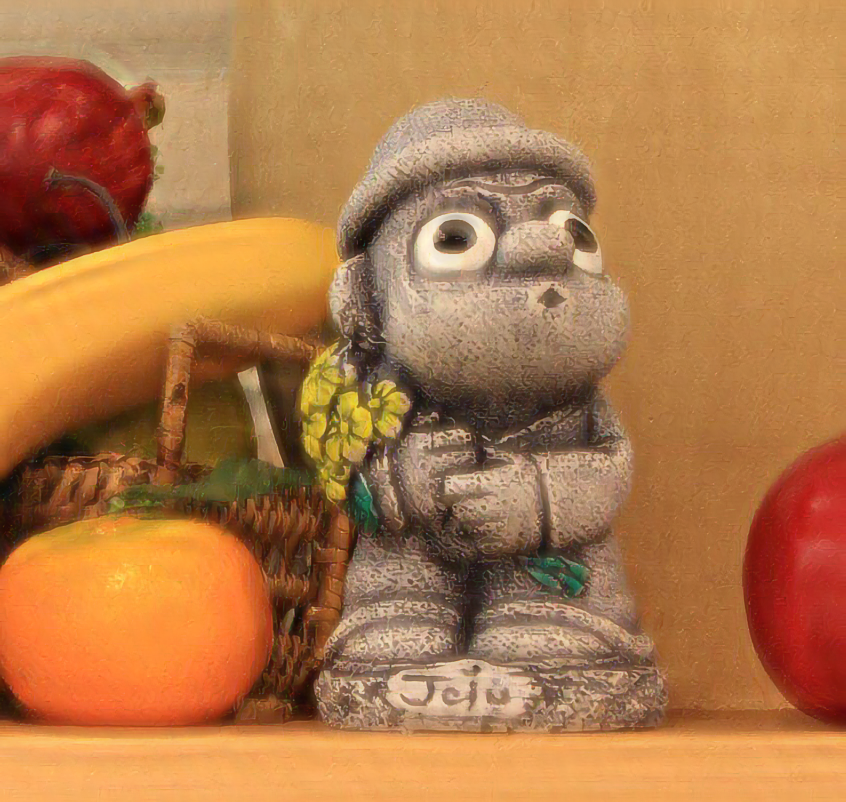}&
		\includegraphics[width=0.16\linewidth]{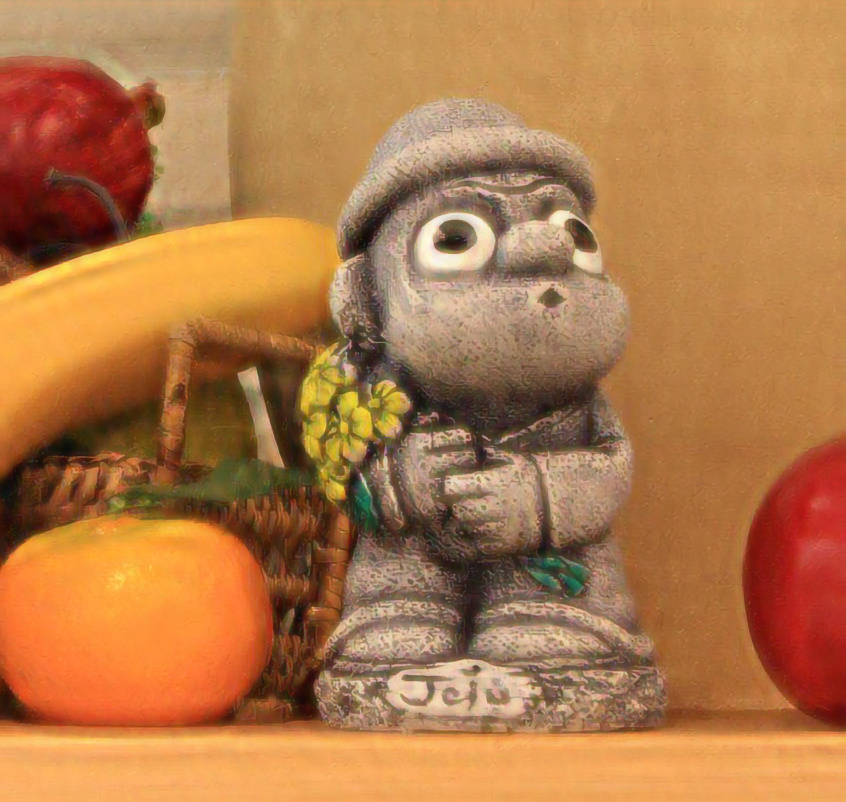}&
		\includegraphics[width=0.16\linewidth]{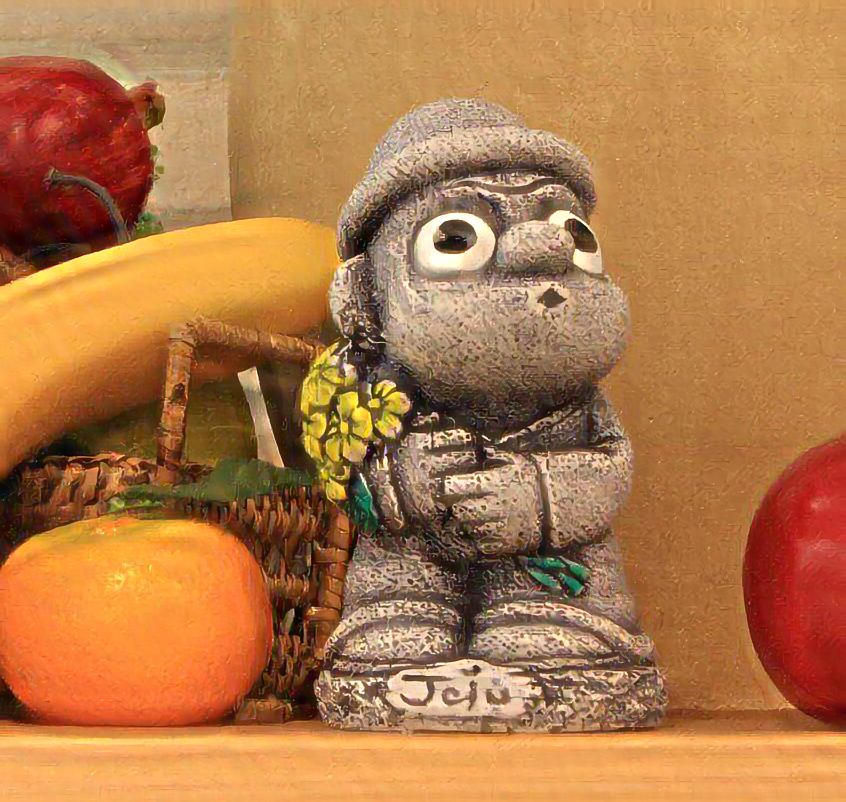}\\
		Degraded&PromptIR&InstructIR&DA-RCOT&MoCE-IR&BaryIR\\
	\end{tabular}
	\caption{Visual examples of the five-degradation models addressing \textbf{real-world mixed degradations}. Row 1-2: haze and rain. Row 3-4: blur and noise.}
	\label{mixed-fig}
\end{figure*}

Tab. \ref{unseen-real-tab} presents the results on unseen real-world data. BaryIR achieves the best performance across all datasets, yielding PSNR gains of 2.09 dB on O-HAZE \cite{ancuti2018haze}, 1.68 dB on SPANet \cite{Wang_2019_CVPR}, and 0.81 dB on LOL-v2-real compared to the second-best methods. Fig. \ref{unseen-real-fig} displays visual examples, where other methods either fail to fully remove degradations or distort structural details, including textures and colors. In contrast, BaryIR restores images with more faithful structures and improved perceptual quality. These results demonstrate that BaryIR generalizes well to real-world unseen degradations, highlighting its applicability for diverse real-world scenarios.

\begin{table}[!h]
	\caption{Quantitative comparison with state-of-the-art methods on mixed-degradation images from CDD-11 \cite{guo2024onerestore}, real-world SPANet \cite{Wang_2019_CVPR} (haze and rain), and Lai \cite{lai2016comparative} (blur and noise). }
	\label{mixed-tab}
	
	\resizebox{1.0\linewidth}{!}{
		\begin{tabular}{lccc}
			\toprule
			\multirow{2}{*}{Method}&CDD-11 \cite{guo2024onerestore} & Real haze and rain& Real blur and noise\\ \cmidrule(lr){2-2}  \cmidrule(lr){3-3} \cmidrule(lr){4-4} 
			& PSNR/SSIM/LPIPS/FID& NIQE ($\downarrow$)/PIQE ($\downarrow$) & NIQE ($\downarrow$)/PIQE ($\downarrow$)\\
			\midrule
			Restormer \cite{Zamir2021Restormer} &26.99/0.864/0.105/23.25& 9.62/115.8&8.56/96.42\\
			IR-SDE \cite{luo2023image} &25.48/0.856/0.112/25.21&9.45/112.1&8.75/100.5\\
			\midrule
			PromptIR \cite{potlapalli2023promptir} &25.90/0.850/0.105/28.54&8.05/102.4&7.22/78.44\\
			DA-CLIP \cite{luocontrolling}&25.88/0.855/0.112/23.66&7.72/95.40&7.45/83.25\\
			DiffUIR \cite{zheng2024selective}&{27.35}/{0.868}/{0.094}/{18.51}&7.78/99.12&7.21/77.35\\
			InstructIR \cite{conde2024instructir} &26.65/0.862/0.108/23.65&7.37/85.93&6.28/{62.18}\\
			DA-RCOT \cite{tang2025degradation} &{28.10}/{0.875}/\underline{0.086}/\underline{13.45}&{6.02}/{64.32}&{5.32}/{60.12}\\
			MoCE-IR \cite{zamfir2025complexity}
			&\underline{29.05}/\underline{0.881}/0.092/17.63&\underline{5.86}/\underline{60.14}&\underline{4.91}/\underline{54.60}\\
			\midrule
			BaryIR &\textbf{29.29}/\textbf{0.887}/\textbf{0.078}/\textbf{11.04}&\textbf{4.62}/\textbf{49.32}&\textbf{3.81}/\textbf{38.32}\\
			\bottomrule
	\end{tabular}}
\end{table}

\begin{table*}[!t]
	\centering
	\caption{Ablation studies on using different embeddings or loss functions \textbf
		{for training}. Metrics are reported as PSNR($\uparrow$)/LPIPS($\downarrow$). }
	\label{rep}
	
	\setlength{\tabcolsep}{5pt}
	\renewcommand{\arraystretch}{1}
	\resizebox{\textwidth}{!}{
		\begin{tabular}{ccccccccccc}
			\toprule
			\multicolumn{3}{c}{Embedding or loss function components} & \multicolumn{6}{c}{\textit{In-distribution}}&\multicolumn{2}{c}{\textit{Out-of-distribution}}\\
			\cmidrule(lr){1-3} \cmidrule(lr){4-9} \cmidrule(lr){10-11}
			\makecell{Original emb.\\ $\bm z_k$}&\makecell{WB emb. \\$\bm b_k$}&\makecell{Residual emb.\\$\bm r_k$}& SOTS &Rain100L &  BSD68\textsubscript{$\sigma$=25} & GoPro& LOL &Average &O-HAZE &SPANet \\
			\midrule
			\cmark & \xmark & \xmark 									
			&24.09/0.065 &34.81/0.045&30.78/0.095&27.22/0.174&20.41/0.109&27.46/0.098&18.02/0.345&34.38/0.032\\
			\xmark & \cmark & \xmark 		&30.27/0.015&37.23/0.025&31.05/0.088&28.05/0.155&22.86/0.096&29.89/0.076&22.04/0.278&38.53/0.022\\
			\cmark & \xmark & \cmark 	
			& 29.40/0.019 &36.23/0.027&30.88/0.093&27.40/0.170&21.78/0.105&29.14/0.083&19.84/0.295&36.22/0.029\\
			\xmark & \cmark & \cmark 	 & \textbf{31.20}/\textbf{0.009}&\textbf{38.10}/\textbf{0.011}&\textbf{31.43}/\textbf{0.086}&\textbf{29.51}/\textbf{0.132}&\textbf{23.37}/\textbf{0.090}&\textbf{30.72}/\textbf{0.066}&\textbf{22.98}/\textbf{0.248}&\textbf{39.24}/\textbf{0.012}\\
			\midrule
			\makecell{MWB loss\\$\mathcal L_{\mathrm{MWB}}$}&\makecell{IRC loss\\$\mathcal  L_{\mathrm{IRC}}$}&\makecell{BRO loss\\$\mathcal L_{\mathrm{BRO}}$}& SOTS &Rain100L &  BSD68\textsubscript{$\sigma$=25} & GoPro& LOL &Average &O-HAZE &SPANet \\\midrule
			\xmark & \xmark & \xmark &24.48/0.057&35.14/0.040&30.82/0.098&27.28/0.170&20.86/0.101&27.72/0.095&19.02/0.312&34.92/0.030\\
			\xmark & \cmark & \cmark 	&24.98/0.049&35.55/0.045&30.88/0.102&27.30/0.170&21.40/0.108&28.02/0.095&19.35/0.316&35.18/0.037\\
			\midrule
			\cmark & \xmark & \xmark 	&28.78/0.023&36.58/0.030&30.98/0.090&27.80/0.154&22.70/0.094&29.37/0.078&21.48/0.287&36.89/0.027\\
			\cmark & \cmark & \xmark 	&30.45/0.013&37.12/0.026&31.02/0.094&28.45/0.148&22.93/0.097&29.99/0.076&21.98/0.285&37.45/0.026\\
			\cmark & \xmark & \cmark 	&29.32/0.021&36.76/0.028&31.07/0.090&27.88/0.156&22.78/0.096&29.56/0.078&21.87/0.289&37.32/0.025\\
			\cmark & \cmark & \cmark 	& \textbf{31.20}/\textbf{0.009}&\textbf{38.10}/\textbf{0.011}&\textbf{31.43}/\textbf{0.086}&\textbf{29.51}/\textbf{0.132}&\textbf{23.37}/\textbf{0.090}&\textbf{30.72}/\textbf{0.066}&\textbf{22.98}/\textbf{0.248}&\textbf{39.24}/\textbf{0.012}\\
			\bottomrule
	\end{tabular}}
\end{table*}
\begin{table*}[t]
	\caption{The \textbf{All-in-One three-degradation} results with different $\lambda_{1:K}$. The metrics are reported as PSNR($\uparrow$)/SSIM($\uparrow$)/LPIPS($\downarrow$)/FID($\downarrow$). }
	\label{weights}
	\setlength{\tabcolsep}{3pt}
	\renewcommand{\arraystretch}{1.2}
	\resizebox{\textwidth}{!}{
		\begin{tabular}{lcccccc}
			\toprule
			\multirow{2}{*}{Method} 
			&\textit{Dehazing} & \textit{Deraining} & \multicolumn{3}{c}{\textit{Denoising}} 
			& \multirow{2}{*}{Average}  \\  \cmidrule(lr){2-2} \cmidrule(lr){3-3} \cmidrule(lr){4-6} 
			& SOTS &Rain100L & BSD68\textsubscript{$\sigma$=15} & BSD68\textsubscript{$\sigma$=25} & BSD68\textsubscript{$\sigma$=50} &  \\\midrule
			uniform $\lambda_{1:K}$&31.27/0.980/0.007/4.655&38.55/0.982/0.012/5.854&34.02/0.932/0.040/24.21&31.42/0.888/0.082/52.56&28.10/0.800/0.166/89.24&32.67/0.916/0.062/35.31\\
			portion-based $\lambda_{1:K}$
			& \textbf{31.40}/\textbf{0.980}/\textbf{0.007}/\textbf{4.523}&\textbf{39.02}/\textbf{0.984}/\textbf{0.008}/\textbf{5.739}& \textbf{34.16}/\textbf{0.935}/\textbf{0.038}/\textbf{22.69}&
			\textbf{31.54}/\textbf{0.892}/\textbf{0.075}/\textbf{40.11} & \textbf{28.25}/\textbf{0.802}/\textbf{0.158}/\textbf{82.63} &\textbf{32.86}/\textbf{0.919}/\textbf{0.057}/\textbf{31.14}\\
			\bottomrule
	\end{tabular}}
\end{table*}

\subsection{Robustness of Generalization Capability to the Number of Training Degradations}
A key question is how effectively the model can capture degradation-agnostic invariance without relying on large-scale training data, which is critical for generalization. To this end, we investigate the robustness of BaryIR in learning generalized degradation-agnostic features when trained on only a limited set of degradation types. Specifically, we consider four settings: 1) all five degradations including dehazing, deraining, denoising ($\sigma=25$), deblurring, and low-light enhancement; 
2) four degradations after removing low-light enhancement; 
3) three degradations after further removing deblurring; 
4) two degradations with only dehazing and deraining. 

Fig. \ref{Rob} presents the quantitative results of image restoration on out-of-distribution (OOD) degradation types. As shown, compared with All-in-One models without agnostic modeling (PromptIR and DA-RCOT), methods that explicitly capture degradation-agnostic representations (DiffUIR, MoCE-IR, and BaryIR) consistently achieve higher performance. Notably, BaryIR achieves the best PSNR and LPIPS scores and exhibits the smallest performance drop as the number of training degradation types decreases, demonstrating superior generalization robustness to unseen degradations. This advantage should arise from BaryIR’s ability to capture intrinsic degradation-agnostic invariance via the WB embeddings, as well as the residual embeddings that retain adaptive degradation-specific knowledge, enables adaptive restoration towards OOD generalization.

\subsection{Handling Images with Mixed Degradations}

The images captured in real scenarios often suffer from mixed degradations caused by a combination of adverse weather and imaging device limitations. To evaluate All-in-One performance under such conditions, we assess BaryIR on 1) the synthetic CDD-11 dataset \cite{guo2024onerestore} and 2) 49 real-world mixed-degradation images collected from Lai \cite{lai2016comparative} (blur and noise) and SPANet \cite{Wang_2019_CVPR} (rain and haze), using the no-reference metrics NIQE \cite{mittal2012making} and PIQE \cite{venkatanath2015blind} for evaluation. For CDD-11, we adopt the same training setup as OneRestore \cite{guo2024onerestore}, while for real-world images, the pre-trained five-degradation BaryIR model from \S\ref{5.1} is used to assess generalization.

Tab. \ref{mixed-tab} and Fig. \ref{mixed-fig} present the results, showing that BaryIR consistently outperforms other methods with significant quantitative and qualitative improvements for handling real-world mixed-degradation images. From the visual results we observe that BaryIR effectively restores high-quality images compared to competing approaches: it removes rain streaks and haze while preserving scene details, and eliminates blur without introducing artifacts. This advantage can be attributed to BaryIR’s ability to learn degradation-agnostic commonalities, which enhances its capacity to recover the underlying structures distorted by composite degradations.

\begin{figure}[!h]
	\setlength\tabcolsep{1pt}
	\renewcommand{\arraystretch}{0.75} 
	\centering
	\begin{tabular}{ccc}
		\includegraphics[width=0.32\linewidth]{Results/850p}&
		\includegraphics[width=0.32\linewidth]{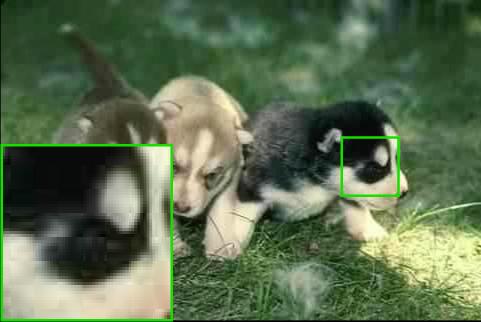}&
		\includegraphics[width=0.32\linewidth]{Results/850baryirpp}\\
		\includegraphics[width=0.32\linewidth]{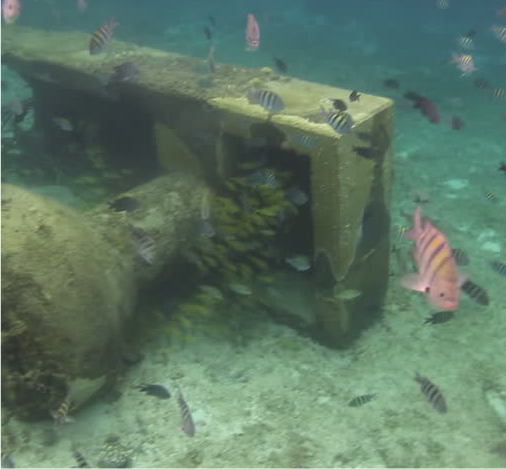}&
		\includegraphics[width=0.32\linewidth]{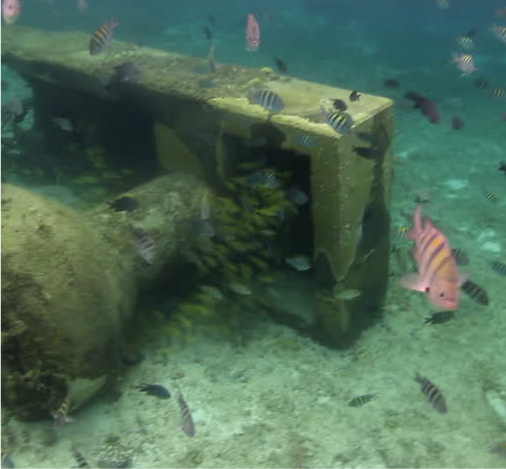}&
		\includegraphics[width=0.32\linewidth]{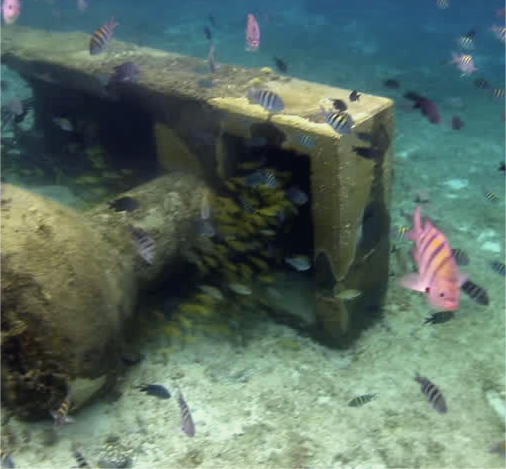}\\
		\includegraphics[width=0.32\linewidth]{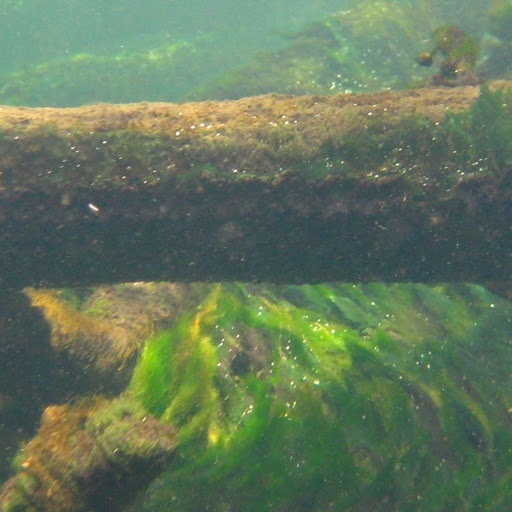}&
		\includegraphics[width=0.32\linewidth]{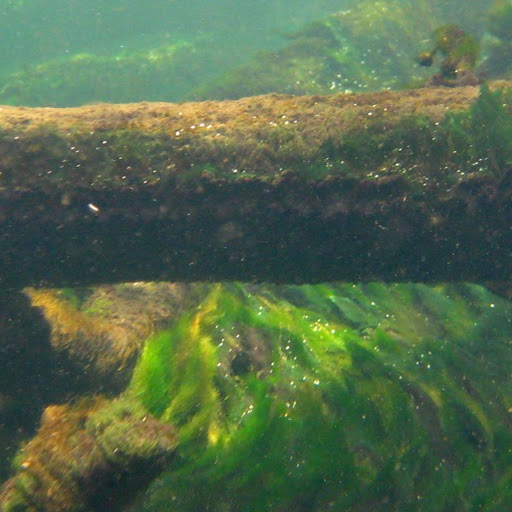}&
		\includegraphics[width=0.32\linewidth]{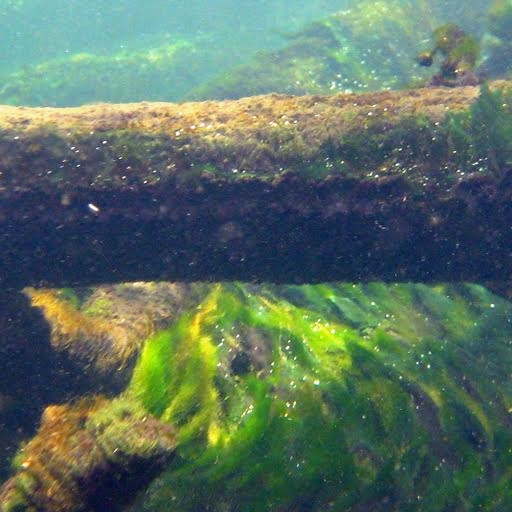}\\
		Degraded&w/o WB&w/ WB\\
		& (Orignal + Res.)& (WB + Res.)
	\end{tabular}
	\caption{\textbf{Visual results to show the effect of embeddings during training}. The model with WB embeddings generalize better to the OOD degradations, restoring sharper images with more faithful structural contents.}
	\label{F-REC}
\end{figure}

\subsection{Ablation Studies}
\noindent\textbf{Effect of different embedding components.}
To examine the role of different embedding components during training, we compare models using: 1) the original embeddings $\bm z_k$; 2) the WB embeddings $\bm b_k$; 3) the residual embeddings $\bm r_k$ combined with the original embeddings $\bm z_k$; 4) both the WB embeddings $\bm b_k$ and residual embeddings $\bm r_k$. 

As shown in Tab. \ref{rep} and Fig. \ref{F-REC}, a key highlight of our contribution is that the WB embeddings alone already \textbf{improve generalization} to unseen degradation type and provide strong restoration performance. This demonstrates that the WB embedding enhances the degradation-agnostic semantics, which is essential for generalization across diverse degradations. The integration of WB and residual embeddings yields optimal performance, indicating that the residuals act as degradation-specific  cues to promote WB space optimization and refine the WB-based common structures for accurate restoration. This synergistic relationship is further reinforced by the barycenter-residual orthogonality constraint.

\begin{figure}[!h]
	\centering
	\includegraphics[width=1\linewidth]{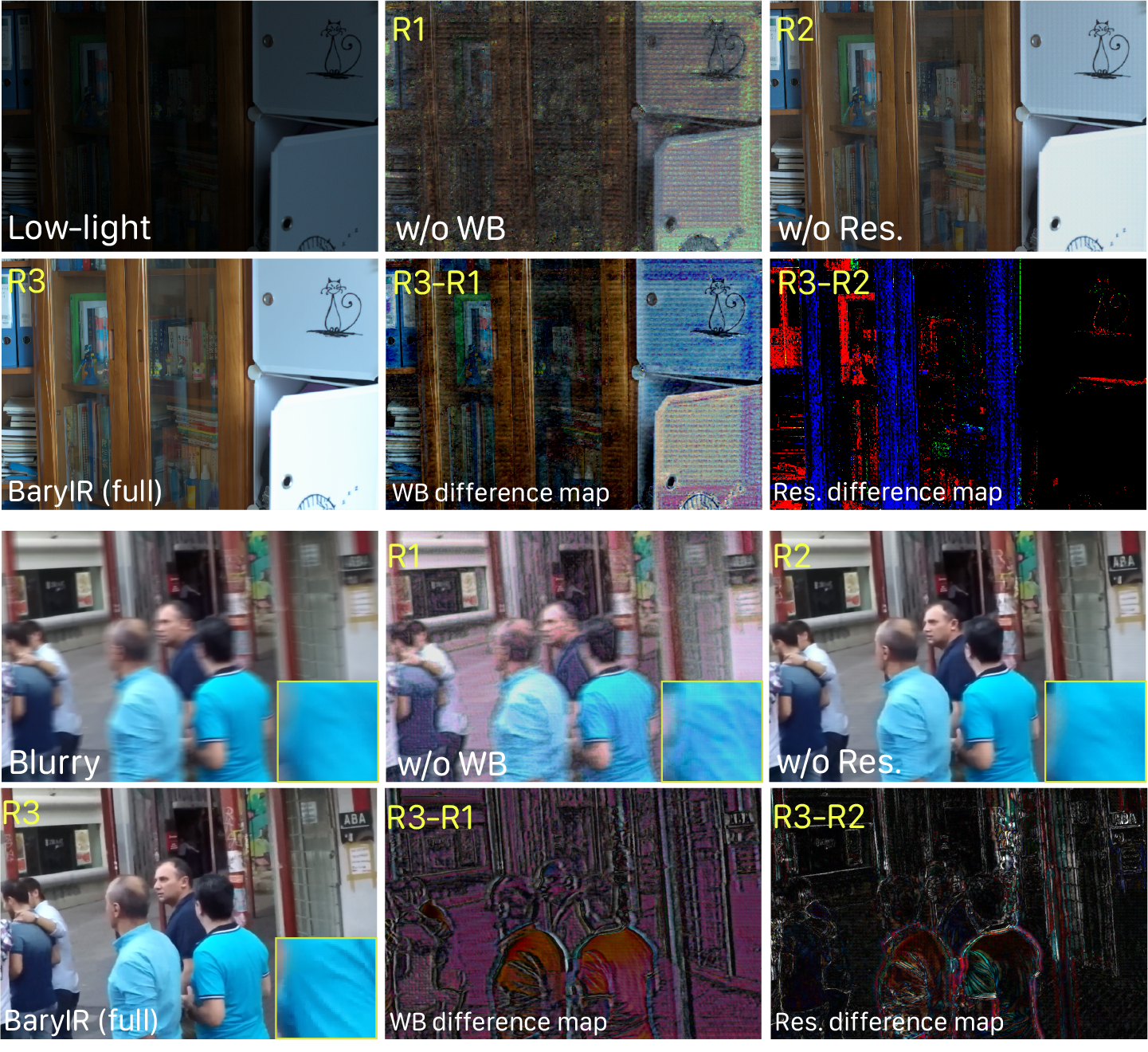}
	\caption{\textbf{Illustration of the contributions of WB and residual embeddings at the image level during inference.} The WB difference maps (R3-R1) primarily capture common image structures and contents. Conversely, the Res. difference maps (R3-R2) are spatially-adaptive,  concentrating on the regions/edges which are heavily affected by the degradations, thereby complementing the WB features with degradation-specific detail refinements, such as high-frequency textures and local contrast adjustments. }
	\label{testvis}
\end{figure}

\noindent\textbf{Test-time visualization of the contributions of WB and residual embeddings at the image level.}
To explore how the learned WB and residual features contribute to the restoration, we perform test-time feature ablation by zeroing out the WB and residual features in turn, yielding two partially restored results, denoted as R1 (w/o WB) and R2 (w/o Res.). The full BaryIR output is R3, and  difference maps R3-R1 and R3-R2 are used to visualize their respective contributions. Fig. \ref{testvis} shows that the WB difference maps primarily capture common image structures and contents, exhibiting degradation-invariant characteristics across different degradations. In contrast, the residual difference maps are spatially-adaptive, concentrating on regions and edges heavily affected by the degradations, thereby capturing degradation-specific patterns that complement the common structures. 

\quad\\
\noindent\textbf{Impact of loss functions.}
We analyze the effect of different loss components used to optimize the barycenter map, including the multisource Wasserstein barycenter loss $\mathcal L_{\mathrm{MWB}}$, the inter-residual contrastive loss $\mathcal L_{\mathrm{IRC}}$, and the barycenter-residual orthogonal loss $\mathcal L_{\mathrm{BRO}}$. As shown in Tab. \ref{rep}, the MWB loss $\mathcal L_{\mathrm{MWB}}$ alone already yields notable performance boosts and substantially improves OOD generalization performance. In contrast, by discarding $\mathcal L_{\mathrm{MWB}}$ while retaining $\mathcal L_{\mathrm{IRC}}$ and $\mathcal L_{\mathrm{BRO}}$, the performance gain is marginal, with only limited improvement in OOD generalization. Notably, the combination of all three terms yields the best performance. The results confirm that the WB is the fundamental contributor of generalization in our framework. The other two loss terms act as synergistic components that facilitate the optimization of WB embeddings to capture degradation-agnostic common structures, while simultaneously retaining degradation-specific patterns within the residual embeddings.

\quad\\
\textbf{Impact of the transport map architectures.} To evaluate the effectiveness of the transport map network and the adaptability of the barycenter optimization framework to different architectures, we conduct an ablation study on the network architectures for parameterizing the transport map $T_\theta$, which is used to approximate the WB in the latent space. Specifically, we adopt four variants: 1) a multilayer perceptron (MLP); 2) a UNet-style mapping implemented using two convolution–ReLU–normalization blocks in the feature space; 3) ViT-style standard transformer blocks composed of self-attention and feed-forward networks; and 4) Our transformer-based blocks with MDTA and GDFN. The architectures are adjusted to maintain the same output dimensionality, ensuring alignment with the restoration backbone.

As shown in Tab. \ref{map}, all candidate architectures for $T_\theta$ yield notable improvements over baseline, which indicates the superior adaptability of BaryIR to different architectures. Particularly, our transformer-based blocks with MDTA and GDFN achieve the best performance, demonstrating the effectiveness of this architectural design.
\begin{table}[!h]
	\centering
	\caption{Ablation studies on architectures for parameterizing the transport map $T_\theta$. Metrics are reported as PSNR($\uparrow$)/LPIPS($\downarrow$). }
	\label{map}
	\resizebox{1.0\linewidth}{!}{
		\begin{tabular}{cccc}
			\toprule
			\multirow{2}{*}{\makecell{Architecture for \\ transport map $T_\theta$}}& {\textit{In-distribution}}&\multicolumn{2}{c}{\textit{Out-of-distribution}}\\
			\cmidrule(lr){2-2} \cmidrule(lr){3-4}
			&Average&O-HAZE&SPANet\\\midrule
			w/o $T_\theta$ (Restormer) &27.46/0.098&18.02/0.345&34.38/0.032\\
			MLP-based mapping&30.04/0.081 &22.14/0.268&38.76/0.022\\
			UNet-style mapping&30.40/0.072&22.55/0.264&39.09/0.019\\
			ViT-style mapping&30.48/0.075&22.71/0.255&39.03/0.017\\
			Ours&30.72/0.066&22.98/0.248&39.24/0.012\\
			\bottomrule
	\end{tabular}}
\end{table}

\quad\\
\textbf{Impact of the barycenter weights setting.} We evaluate two weighting strategies for the barycenter weights $\lambda_k$: uniform weights for all sources, and proportion-based weights according to the number of training samples per source. As shown in Tab.~\ref{weights}, proportion-based weights consistently outperform uniform weights across all tasks, achieving higher PSNR/SSIM and lower LPIPS/FID. This demonstrates that assigning barycenter weights according to the data distribution helps BaryIR better aggregate multisource information for more efficient all-in-one image restoration. 

\begin{figure}[!htbp]
	\centering
	\includegraphics[width=1\linewidth]{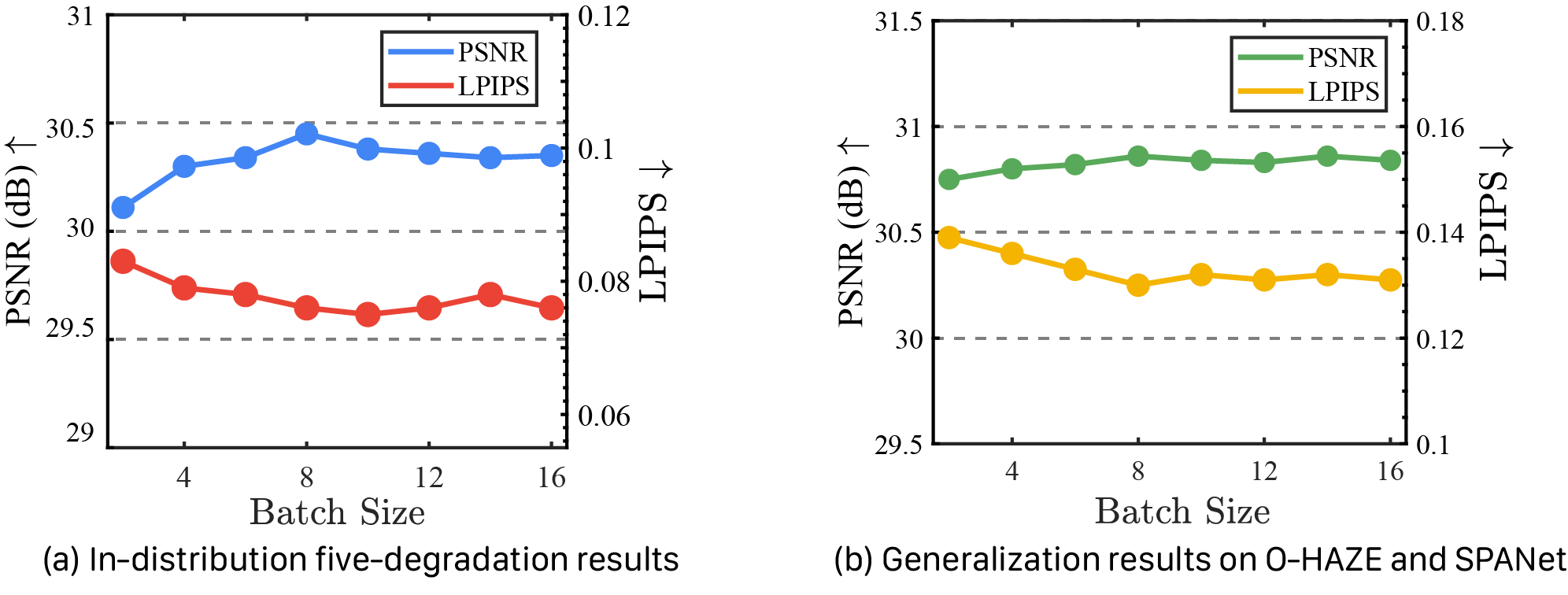}
	\caption{Impact of the training batch size. We report the average results on five-degradation test sets and on OOD O-HAZE+SPANet data. }
	\label{batch}
\end{figure}

\noindent\textbf{Impact of the training batch size.}  Since the barycenter is optimized over mini-batches and may vary significantly with batch size, we evaluate its impact on the five-degradation and OOD scenarios, in Fig. \ref{batch}. The training images are cropped as $64\times64$ patches. Across both in-distribution (a) and OOD (b) scenarios, PSNR initially improves and stabilizes beyond a batch size of 8, while LPIPS consistently decreases and levels off. This indicates that a moderate batch size is sufficient for robust barycenter optimization and efficient restoration.

To ensure a fair comparison in Tabs. \ref{air3} and \ref{air5}, we maintain a batch size of 4, aligning with baseline methods while using images cropped as $128\times128$ patches.

\subsection{Discussion and Model Analysis}
\subsubsection{Parameter quantity and computational complexity}
As shown in Tab. \ref{para}, BaryIR introduces only a moderate computational overhead compared to Restormer \cite{Zamir2021Restormer}. Specifically, it incorporates an additional 8.3M parameters and 64G FLOPs brought by the barycenter mapping module, while keeping the inference time at 0.16s (0.13s + 0.03s), which is still competitive among recent models. Notably, BaryIR remains significantly more efficient than DA-CLIP \cite{luocontrolling} with its large parameter size (174.1M) and long inference latency (4.59s). Compared to MoCE-IR \cite{zamfir2025complexity}, BaryIR achieves a favorable balance between accuracy and efficiency, requiring fewer parameters and FLOPs while maintaining fast inference. These results demonstrate that BaryIR achieves state-of-the-art generalization without sacrificing computational efficiency.
\begin{table}[!h]
	\centering
	\vspace{-0.2cm}
	\caption{Comparison of the number of parameters, model computational efficiency, and inference time. The flops and inference time are computed on rainy image of size 256$\times$256.}
	\vspace{-0.2cm}
	\label{para}
	\resizebox{0.48\textwidth}{!}{
		\begin{tabular}{l|cccccc}
			\toprule
			Method  & Restormer & PromptIR & DA-CLIP&DA-RCOT&MoCE-IR&BaryIR \\
			\midrule
			\#Param&26.1M&36.3M&174.1M&40.9M&25.4M&26.1M + 8.3M\\
			Flops&118G&158G&118.5G&262G&142G&118G + 64G\\
			Time&0.13s&0.15s&4.59s&0.21s&0.16s
			&0.13s + 0.03s\\
			\bottomrule
	\end{tabular}}
\end{table}

\noindent\textbf{Cost-performance trade-off analysis.} As shown in Table~\ref{cp}, incorporating the proposed OT-based framework introduces a moderate increase in training cost (approximately +9.8M parameters), which mainly comes from the parameterization of the transport map $T_\theta$ and the potential networks $f_{\omega_{1:K}}$. Despite this additional overhead, both training and inference time only increase marginally, while the model achieves consistent and substantial performance improvements, especially under unseen and mixed degradation settings. These results demonstrate a favorable cost-performance trade-off, where a limited increase in computational cost leads to enhanced robustness and generalization.

\begin{table}[!h]
	\centering
	\caption{Cost-performance trade-off analysis.  Metrics are reported as PSNR($\uparrow$)/LPIPS($\downarrow$).}
	\label{cp}
	\resizebox{1.0\linewidth}{!}{
		\begin{tabular}{lccccccc}
			\toprule
			\multirow{2}{*}{Method}&	\multicolumn{2}{c}{Training cost}&	\multicolumn{2}{c}{Inference cost}& {\textit{In-distribution}}&\multicolumn{2}{c}{\textit{Out-of-distribution}}\\
			\cmidrule(lr){2-3} \cmidrule(lr){4-5} \cmidrule(lr){6-6} \cmidrule(lr){7-8}
			&\#Param&Time (1 epoch)&\#Param&Time&Average&O-HAZE&SPANet\\\midrule
			Restormer&26.1M&3.2 hours&26.1M&0.13s&27.46/0.098 &18.02/0.345&34.38/0.032\\
			BaryIR$_{\text{Restormer}}$&35.9M&3.8 hours&34.4M&0.16s&30.72/0.066&22.98/0.248&39.24/0.012\\
			\midrule
			PromptIR&36.3M&3.5 hours&36.3M&0.15s&29.72/0.084&18.38/0.338&35.34/0.026\\
			BaryIR$_{\text{PromptIR}}$&46.1M&4.0 hours&44.6M&0.18s&31.05/0.070&23.21/0.245&39.38/0.013\\
			\bottomrule
	\end{tabular}}
\end{table}

\subsubsection{Training cost curves}
\begin{figure}[!htbp]
	\centering
	\includegraphics[width=0.9\linewidth]{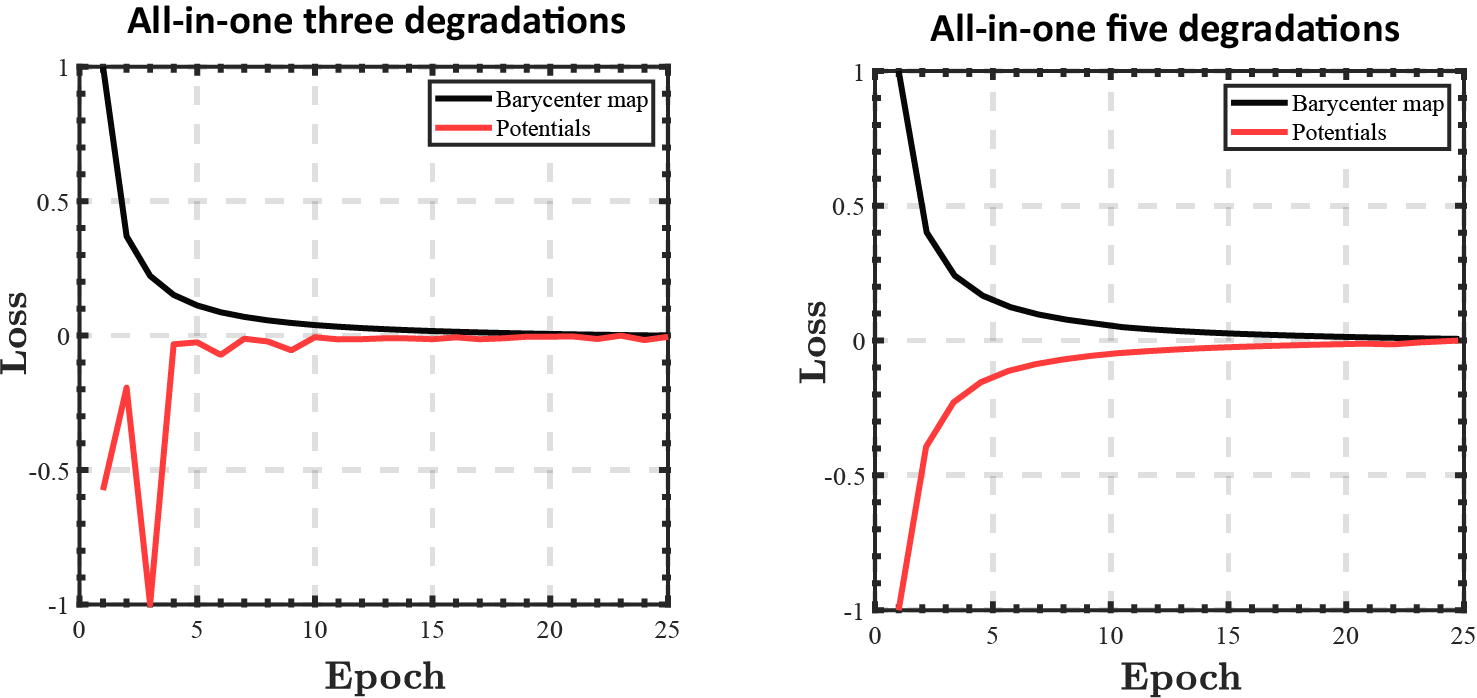}
	\caption{The loss curves of the barycenter map and potentials during training. The loss of $T_\theta$ is scaled to $[0,1]$. The loss of $f_{\omega_{1:K}}$ is scaled to $[0,1]$ and then takes the negative.}
	\label{curve}
\end{figure}
In Fig. \ref{curve}, we present the loss curves of the barycenter map $T_\theta$ and the potentials $f_{\omega_{1:K}}$ during training on All-in-One settings with three and five degradations. For clarity, the loss of $T_\theta$ is normalized to $[0,1]$, while the loss of $f_{\omega_{1:K}}$ is scaled to $[0,1]$ and then negated. We observe that both curves converge stably in an adversarial manner, which validates the effectiveness of our optimization scheme for solving the barycenter map.

\begin{figure}[!htbp]
	\centering
	\includegraphics[width=1\linewidth]{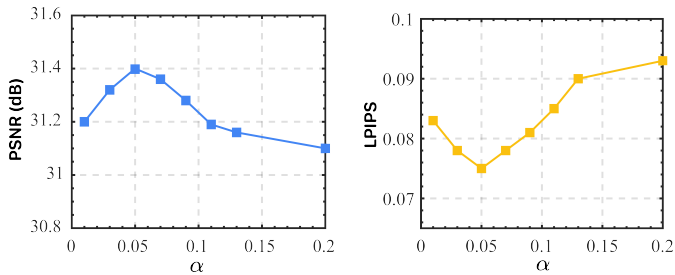}
	\caption{Effect of the trade-off hyperparameter $\alpha$ on validation performance.}
	\label{sens}
\end{figure}

\subsubsection{Hyperparameter selection}
To determine the trade-off hyperparameter $\alpha$ that constrains the contrastive and orthogonal terms in (\ref{baryloss}), we conduct a sensitivity analysis by training the five-degradation models on 90\% of the data and using the remaining 10\% as validation sets. The results with respect to $\alpha$ are shown in Fig.~\ref{sens}, indicating that $\alpha = 0.05$ yields the best performance. Accordingly, we set $\alpha = 0.05$ in all settings.

\subsection{Analysis of Representative Failure Cases}
\begin{figure}[!htbp]
	\centering
	\includegraphics[width=1\linewidth]{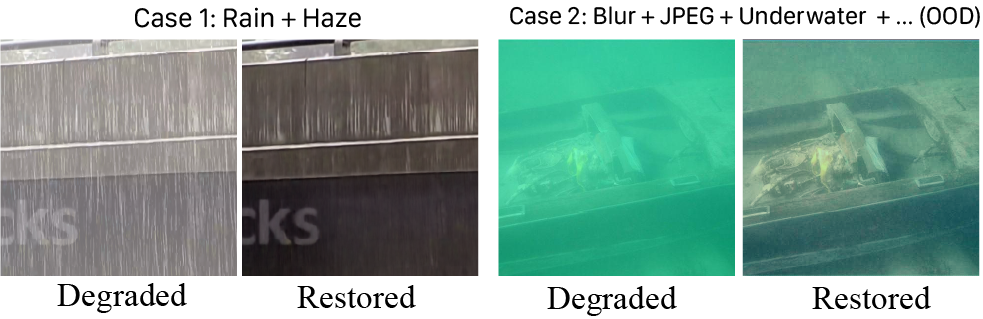}
	\caption{Some failure cases on severe real-world mixed degradations.}
	\label{fail}
\end{figure}

Despite its robustness, our model encounters limitations in severe real-world mixed-degradation scenarios (Fig. \ref{fail}). 

In \textbf{Case 1 (rain + haze)}, the model fails to eliminate rain streaks that exhibit significantly higher intensity relative to surrounding regions. This failure reveals that the residual space, though designed to capture degradation-specific patterns, lacks sufficient sensitivity to localized intensity outliers. Due to this insensitivity, the residual decoupling mechanism fails to identify these streaks as degradations, leading the model to incorrectly treat them as high-salience image structures. Consequently, the residual space remains "blind" to these outliers, thereby resulting in their unintended preservation. 

In \textbf{Case 2 (OOD complex degradations)} involving complex degradations such as underwater scattering and JPEG artifacts, the model achieves color correction but fails to restore sharp textures or suppress artifacts. The potential reason is that reliable scene details are heavily obscured in such complex mixed degradations. In these extreme cases, the barycenter representation tends to emphasize shared global structures across degradations, which facilitates color consistency but may trade off fine-grained detail recovery.  Furthermore, this representation makes systematic JPEG compression hardly distinguishable from common structures. As a result, the model recovers common image structures but preserves the artifacts while smoothing genuine textures. This reflects a trade-off where global consistency is maintained at the expense of texture clarity and artifact suppression.

\section{Conclusion, Limitation, and Future Work}

In this work, we presented BaryIR, a novel framework for all-in-one image restoration that explicitly disentangles degradation-agnostic and degradation-specific representations. By learning a continuous Wasserstein barycenter (WB) space to capture invariant contents and constructing orthogonal residual subspaces for dynamic degradation-specific knowledge, BaryIR effectively alleviates overfitting and adapts to unseen degradations. Our adversarial max-min optimization ensures a smooth, geometrically consistent barycenter mapping, while theoretical error bounds provide guarantees for the recovered barycenter distribution. Extensive experiments on both synthetic and real-world datasets demonstrate that BaryIR not only achieves state-of-the-art restoration performance but also exhibits superior generalization robustness to out-of-distribution degradations and maintains stable performance under limited training degradation types. These results validate the effectiveness of barycenter-based feature disentanglement as a principled approach for generalized image restoration.

We acknowledge that there are potential minor limitations. For example, based on our ablation study, the weights $\lambda_k$ are currently determined by the number of training samples for each source, which is still empirical. In future work, we aim to provide theoretical justifications for weight selection and develop more adaptive strategies. Moreover, we are interested in extending the barycenter-driven framework to general multimodal data (\textit{e.g.,} text, image, audio) and apply it to broader tasks such as multimodal understanding and generation.

\bibliographystyle{ieeetr}
\bibliography{ref}

%
%
%




\end{document}